\documentclass[10pt,twocolumn,letterpaper]{article}
\PassOptionsToPackage{table}{xcolor}
\usepackage{cvpr}            
\usepackage{cuted}
\usepackage[T1]{fontenc}
\usepackage{currfile}

\usepackage{amsthm,mathtools,titletoc,tabularx,microtype}
\usepackage{pgffor}
\usepackage{placeins}
\makeatletter
\newcommand{\FloatBarrierAbove}{}
\makeatother
\newcommand{\ns}[1]{\textcolor{black!45}{#1}}

\definecolor{titlebar}{RGB}{208, 224, 245}
\definecolor{closed_models}{RGB}{255, 190, 168}
\definecolor{open_models_small}{RGB}{195, 255, 194}
\definecolor{open_models_large}{RGB}{177, 252, 226}
\definecolor{reasoning_models-c}{RGB}{234, 193, 245}
\definecolor{reasoning_models-o}{RGB}{192, 168, 237}
\definecolor{human_baseline}{RGB}{171, 229, 247}

\newcommand{\E}{\mathbb{E}}
\newcommand{\Var}{\operatorname{Var}}

\newcommand{\eAgreeValenceNonsens}{6}

\newcommand{\eAngerNetPooled}{-44}

\newcommand{\eBelowValenceAll}{12}

\newcommand{\eBelowValenceNonsens}{10}

\newcommand{\eCatAnimalValence}{+0.017}

\newcommand{\eCatObjectValence}{+0.059}

\newcommand{\eCatPersonValence}{-0.238}

\newcommand{\eCompRnonsensMean}{0.87}
\newcommand{\eCompRsensMean}{0.89}
\newcommand{\eConsCross}{0.429}

\newcommand{\eConsLoadPos}{17}
\newcommand{\eConsPCone}{47.0}
\newcommand{\eConsPConeAro}{27.4}
\newcommand{\eConsPCtwo}{6.1}
\newcommand{\eConsPairMax}{0.71}
\newcommand{\eConsPairMin}{0.16}
\newcommand{\eConsPairR}{0.428}
\newcommand{\eConsPairRAro}{0.207}
\newcommand{\eConsPperm}{0.539}
\newcommand{\eConsRel}{0.93}
\newcommand{\eConsRpart}{0.050}
\newcommand{\eConsRpartAro}{-0.267}
\newcommand{\eConsRpartAroHi}{-0.21}
\newcommand{\eConsRpartAroLo}{-0.33}
\newcommand{\eConsRpartHi}{0.13}
\newcommand{\eConsRpartLo}{-0.02}
\newcommand{\eConsRpcArousal}{-0.47}
\newcommand{\eConsRpcHv}{0.01}
\newcommand{\eConsRpcSens}{-0.33}
\newcommand{\eConsSame}{0.424}
\newcommand{\eCorAgeRefDelta}{-0.65}

\newcommand{\eCorLenDelta}{-0.73}

\newcommand{\eCorMemDelta}{-0.76}

\newcommand{\eCorRbaseDelta}{-0.07}
\newcommand{\eCorRbaseDeltaP}{0.78}
\newcommand{\eCorSdDelta}{-0.09}
\newcommand{\eCorSdDeltaP}{0.72}

\newcommand{\eDefAbsMiddleValence}{0.236}

\newcommand{\eDefAbsOlderValence}{0.351}

\newcommand{\eDefAbsYoungValence}{0.243}

\newcommand{\eDefImpliedMedianValence}{54}

\newcommand{\eDefImpliedNValence}{12}
\newcommand{\eDefNearMiddleArousal}{10}
\newcommand{\eDefNearMiddleValence}{13}

\newcommand{\eDefNearYoungValence}{4}

\newcommand{\eDisgustNetPooled}{+96}
\newcommand{\eFailMinPchi}{0.10}
\newcommand{\eFailOlder}{43}

\newcommand{\eFailTotal}{158}
\newcommand{\eFailYoung}{36}
\newcommand{\eFearNetPooled}{+184}
\newcommand{\eFearPos}{14}

\newcommand{\eGeoAppFlatter}{16}
\newcommand{\eGeoAppHiOlder}{0.30}
\newcommand{\eGeoAppHiYoung}{0.57}
\newcommand{\eGeoAppHuman}{0.64}
\newcommand{\eGeoAppLoOlder}{0.12}
\newcommand{\eGeoAppLoYoung}{0.38}
\newcommand{\eGeoAppOlder}{0.21}
\newcommand{\eGeoAppYoung}{0.48}

\newcommand{\eGeoAveHuman}{0.51}

\newcommand{\eGeoAveOlder}{0.76}
\newcommand{\eGeoAveSteeper}{9}
\newcommand{\eGeoAveYoung}{0.79}
\newcommand{\eGeoKF}{8}
\newcommand{\eHolderMean}{2.47}

\newcommand{\eHumArousalFlagPleas}{-0.266}

\newcommand{\eHumArousalFlagUnpl}{+0.451}

\newcommand{\eHumArousalUnflPleas}{-0.075}

\newcommand{\eHumArousalUnflUnpl}{+0.211}

\newcommand{\eHumNAroOlder}{16}
\newcommand{\eHumNValOlder}{25}
\newcommand{\eHumValenceAll}{+0.111}
\newcommand{\eHumValenceAllHi}{+0.29}
\newcommand{\eHumValenceAllLo}{-0.05}
\newcommand{\eHumValenceFlagPleas}{+0.443}
\newcommand{\eHumValenceFlagPleasHi}{+0.86}
\newcommand{\eHumValenceFlagPleasLo}{+0.10}
\newcommand{\eHumValenceFlagUnpl}{-0.231}

\newcommand{\eHumValenceNonsens}{+0.139}
\newcommand{\eHumValenceNonsensHi}{+0.32}
\newcommand{\eHumValenceNonsensLo}{-0.03}
\newcommand{\eHumValenceSens}{-0.013}

\newcommand{\eHyoungMean}{2.49}
\newcommand{\eJSymMedian}{0.001}
\newcommand{\eJSyoMedian}{0.004}
\newcommand{\eJoyNetPooled}{-133}

\newcommand{\eLadderMemEight}{7.4}
\newcommand{\eLadderMemThirtyTwo}{63.8}

\newcommand{\eLexExciteDownModels}{17}

\newcommand{\eLexExcitementOlder}{6.4}
\newcommand{\eLexExcitementYoung}{15.7}

\newcommand{\eLexExplicitAgeOlder}{6.9}
\newcommand{\eLexExplicitAgeYoung}{1.6}

\newcommand{\eLexFirstPersonOlder}{15.6}
\newcommand{\eLexFirstPersonYoung}{5.3}
\newcommand{\eLexLengthSigModels}{16}

\newcommand{\eLexMemoryMaxModel}{Gemma-3-27B}
\newcommand{\eLexMemoryMaxRate}{88.0}

\newcommand{\eLexMemoryOlder}{31.8}

\newcommand{\eLexMemoryYoung}{3.7}

\newcommand{\eLexTopOlder}{nostalgia, reminds, memories, reminiscent, reminding, older}
\newcommand{\eLexTopYoung}{kinda, cute, vibe, excitement, really, intense}

\newcommand{\eLexVulnerabilityOlder}{21.4}
\newcommand{\eLexVulnerabilityYoung}{11.0}

\newcommand{\eLexWordsOlder}{16.1}
\newcommand{\eLexWordsYoung}{14.0}

\newcommand{\eModValNonsensMax}{+0.084}

\newcommand{\eModValNonsensMin}{-0.134}
\newcommand{\eModValenceFlagPleasMax}{-0.298}
\newcommand{\eModValenceFlagPleasMean}{-0.845}
\newcommand{\eModValenceFlagPleasMin}{-1.491}

\newcommand{\eModValenceFlagUnplMean}{-0.171}

\newcommand{\eMonoBetweenMeanValence}{19}

\newcommand{\eMonoOutsideMeanValence}{14}

\newcommand{\eMonoPosMedianArousal}{0.31}
\newcommand{\eMonoPosMedianValence}{0.16}

\newcommand{\eMonoStepOlderMeanValence}{39}

\newcommand{\eMoveGtOneMeanValence}{6.0}

\newcommand{\eMoveZeroMeanValence}{46}
\newcommand{\eNFlagPleas}{54}
\newcommand{\eNFlagUnpl}{113}

\newcommand{\eNeutralNetPooled}{-111}

\newcommand{\ePairGemmaNewDelta}{-0.057}
\newcommand{\ePairGemmaNewMem}{22.7}

\newcommand{\ePairGemmaOldDelta}{-0.157}
\newcommand{\ePairGemmaOldMem}{88.0}

\newcommand{\ePredBetaAroValence}{-0.252}

\newcommand{\ePredBetaFlagHiValence}{-0.099}

\newcommand{\ePredBetaFlagLoValence}{-0.271}
\newcommand{\ePredBetaFlagValence}{-0.185}

\newcommand{\ePredRtwoFlagValence}{0.11}

\newcommand{\ePredRtwoFullMax}{0.38}
\newcommand{\ePredRtwoFullMedian}{0.24}
\newcommand{\ePredRtwoFullMin}{0.07}
\newcommand{\ePredRtwoFullValence}{0.43}

\newcommand{\ePredRtwoValFlagValence}{0.16}
\newcommand{\ePredRtwoValValence}{0.02}
\newcommand{\eSadNetPooled}{+271}
\newcommand{\eSurpriseNetPooled}{-206}
\newcommand{\eTextBetaExcitementMedian}{+0.40}

\newcommand{\eTextBetaExcitementPos}{17}

\newcommand{\eTextBetaMemoryMedian}{+0.14}

\newcommand{\eTextBetaMemoryPos}{13}
\newcommand{\eTextBetaVulnerabilityMedian}{-0.17}
\newcommand{\eTextBetaVulnerabilityNeg}{12}

\newcommand{\eTextRtwoGainMax}{0.24}
\newcommand{\eTextRtwoGainMedian}{0.10}

\newcommand{\eThinkAroBase}{-0.296}
\newcommand{\eThinkAroThink}{-0.441}

\newcommand{\eThinkDeltaBase}{-0.020}
\newcommand{\eThinkDeltaThink}{+0.010}

\newcommand{\eFigPurityBaseline}{39}
\newcommand{\eFigPurityMiddle}{25}
\newcommand{\eFigPurityOlder}{37}
\newcommand{\eFigPurityYoung}{31}

\newcommand{\eFigTextNNsame}{78}

\definecolor{cvprblue}{rgb}{0.21,0.49,0.74}
\usepackage[pagebackref,breaklinks,colorlinks,allcolors=cvprblue]{hyperref}
\def\confName{CVPR}
\def\confYear{2027}

\title{Characterizing Age-Conditioned Affective Reasoning in Vision-Language Models}
\author{%
Shahriyar Zaman Ridoy$^{1},$\quad
Azmine Toushik Wasi$^{2},$\quad
Lin Gu$^{3},$\quad
Ruogu Fang$^{4}$\\[4pt]
{\normalsize $^{1}$Department of Electrical and Computer Engineering, North South University, Bangladesh}\\
{\normalsize $^{2}$CVSSP and PAI, University of Surrey, United Kingdom}\\
{\normalsize $^{3}$Research Institute of Electrical Communication, Tohoku University, Japan}\\
{\normalsize $^{4}$Department of Biomedical Engineering, Vanderbilt University, USA}\\[3pt]
{\tt\small shahriyar.ridoy@northsouth.edu, a.wasi@surrey.ac.uk,}\\
{\tt\small lin.gu@riken.jp, ruogu.fang@vanderbilt.edu}
}

\begin{document}
\maketitle
\begin{strip}
\centering
\includegraphics[width=0.80\linewidth]{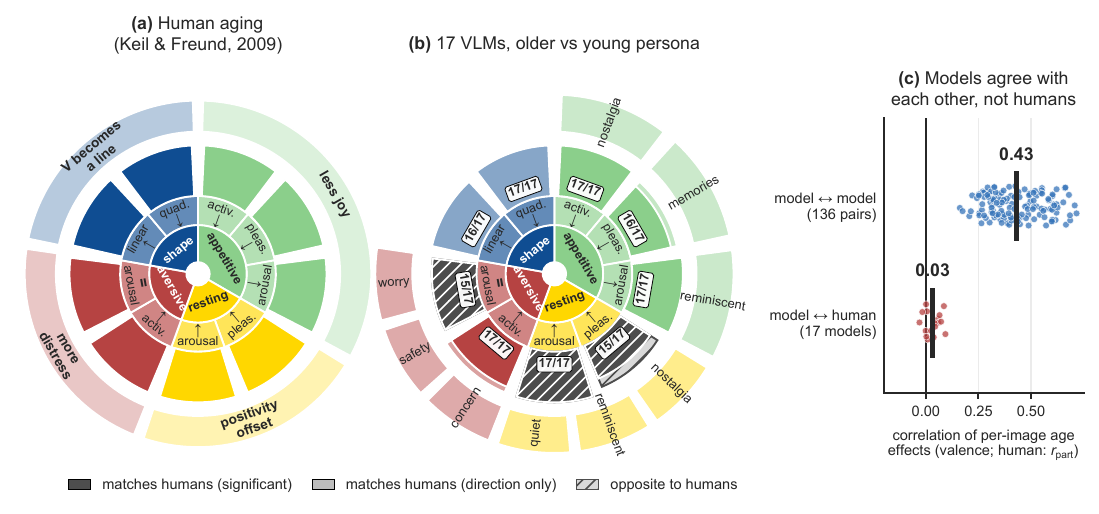}
\captionsetup{hypcap=false}
\captionof{figure}{\textbf{Age differences in human ratings and age-conditioned model predictions.}
(a)~Age differences predicted by \citet{keil2009changes} for each affect system; arrows give the predicted direction.
(b)~The same systems under the older versus young persona, pooled over 17 VLMs; shading marks whether the model change matches OASIS raters (significant, direction only, or opposite). Keil and Freund measure within-person coupling, which OASIS cannot; computed with the same across-image statistics, OASIS raters aged 60+ show no change in either coupling and rate pleasant images $0.31$ points higher, a difference 16 of 17 models reverse ($\S$\ref{app:rb-geometry}).
(c)~Per-image age effects agree across model pairs ($r=0.43$) but not with humans ($r_{\text{part}}=0.03$).}
\captionsetup{hypcap=true}
\label{fig:teaser}
\end{strip}

\begin{abstract}
An age prompt can change how a vision-language model (VLM) rates the emotion evoked by an image. Does the change predict how people of different ages rate that same image? We evaluate 17 VLMs on 900 OASIS images using young-, middle-, and older-adult prompts and estimate image-specific age differences from ratings by 822 participants. Without an age cue, models closely track average human valence (median Pearson $r=0.891$). Under an older-adult prompt, 16 of 17 models assign lower valence on average than under a young-adult prompt. Yet after adjusting for image pleasantness, correlations between model and human age effects range from $-0.030$ to $0.093$, against an estimated ceiling of $0.595$. The discrepancy is clearest on 54 sensitive images rated as pleasant. Across models, the older prompt reduces valence by $0.845$ points on average, while older participants rate these images $0.443$ points higher than younger participants. Providing human ratings from other images improves agreement on the direction of age differences, although image-specific fidelity remains modest on 600 held-out images ($r_{\mathrm{part}}=0.105$). These results show why evaluations of demographic personalization in foundation models need human comparisons at the level of individual stimuli. Project page: \url{https://shahriyar-zaman.github.io/age-conditioned-emotion-reasoning/}
\end{abstract}

\section{Introduction}
\label{sec:introduction}

Vision-language models (VLMs) can discuss images with users \citep{liu2023visual,chu2024multimodal}, including in settings involving companionship and emotional support \citep{liu2021support,defreitas2024companions}. Personalizing these interactions requires knowing when information about a user improves a prediction. For example, a model asked to rate a photograph from an older adult's perspective might describe a scene as less pleasant. The change shows that the model followed the age cue. Whether it predicts a difference in how people experience that photograph requires an independent human comparison \citep{argyle2023one,aher2023simulation,santurkar2023opinions,bisbee2024synthetic,lutz2025persona}.

Research on emotional aging gives this comparison a concrete target. Socioemotional selectivity theory relates changes in emotional goals to perceived time horizons \citep{mather2005aging}. The associated positivity effect concerns attention to and memory for positive information \citep{reed2014positivity}. In image-rating studies, age differences vary with the stimulus and with the dimension being rated \citep{gruhn2008ratings,keil2009changes}. An image can be pleasant yet calming, for example. We therefore examine both \emph{valence} (pleasantness) and \emph{arousal} (activation) \citep{russell1980circumplex,bradley1994sam}.

Existing benchmarks evaluate emotion recognition and explanation \citep{achlioptas2021artemis,bhattacharyya2025evaluating}, agreement with pooled affective norms \citep{ogg2025aligned,nowicki2026visual}, and demographic variation in recognition or persona-conditioned ratings \citep{okawa2026emergence,nowicki2026visual,kobayashi2026comparing}. Agreement with a pooled norm can remain high even when a model reverses a group difference. We ask whether an age cue changes a model's judgment on the same images, and in the same direction, as the age differences observed among human raters (Figure~\ref{fig:teaser}).

In this work, we evaluate 17 VLMs from seven developers on all 900 OASIS images \citep{kurdi2017oasis}. Each model rates each image under four instructions specifying no age, young adulthood, middle adulthood, or older adulthood. Participant-level ratings let us estimate human age differences for each image and compare them with the model contrasts while holding the stimulus fixed (Figure~\ref{fig:study-overview}).
\textbf{Our core contributions are:}

\begin{enumerate}[
leftmargin=1.2em,
itemsep=0.3pt,
topsep=0.3pt,
parsep=0pt,
partopsep=0pt
]
\item \textbf{Evaluation.} We distinguish agreement with pooled ratings, sensitivity to the age prompt, and agreement with image-specific human age differences. For the last measure, we control for image pleasantness and estimate a ceiling from uncertainty in the human ratings.
\item \textbf{Characterization.} Models show a common response to age cues (mean pairwise $r=0.428$). On pleasant images flagged as sensitive, all 17 reverse the observed human valence contrast. Their descriptions also use memory and vulnerability language more often under the older persona.
\item \textbf{Interventions.} We compare prompt-based reasoning with methods that retrieve human evidence from other images. We evaluate the selected human-grounded method, without further tuning, on 600 held-out images and compare it with a pleasantness-only rule.
\end{enumerate}
Taken together, our findings show that following a demographic cue is not the same as capturing how people from that demographic group experience an image. This distinction matters for VLMs used in personalized interactions, where age-related assumptions can shape how models interpret and describe the experiences of their users.

\section{Background and Related Work}
\label{sec:related}
\noindent \textbf{Affect and aging.} We describe affect with continuous dimensions and discrete categories~\citep{russell1980circumplex,shaver1987emotion,cowen2017categories} and use normative image sets that allow stimulus-level comparison~\citep{marchewka2014naps,kurdi2017oasis}. Emotional processing changes across adulthood, including positivity effects in attention and memory and changes in how valence relates to arousal~\citep{mather2005aging,keil2009changes}. \\
\textbf{Affect in models.} Prior work evaluates visually elicited affect~\citep{achlioptas2021artemis}, emotional understanding~\citep{sabour2024emobench}, human--model emotional alignment~\citep{huang2024emotionbench}, and multimodal emotional intelligence~\citep{hu2026emobenchm}; the closest study analyzes emotion hierarchies and demographic patterns in recognition~\citep{okawa2026emergence}, and activation-level work finds functional emotion concepts inside language models~\citep{sofroniew2026emotion}. We study observable judgments rather than internal mechanisms.\\
\textbf{Persona simulation.} Demographic prompts are used to simulate user groups~\citep{argyle2023one,aher2023simulation}, but they can misalign with the groups and are sensitive to prompt design~\citep{santurkar2023opinions,bisbee2024synthetic,lutz2025persona}. We add a stimulus-matched test that separates responsiveness to an age instruction from fidelity to the age differences humans show on the same images. $\S$\ref{app:related-detailed} gives a detailed discussion.

\section{Method}
\label{sec:method}
\begin{figure*}[t]
    \centering
    \includegraphics[width=0.88\linewidth,trim={0 9pt 0 0},clip]{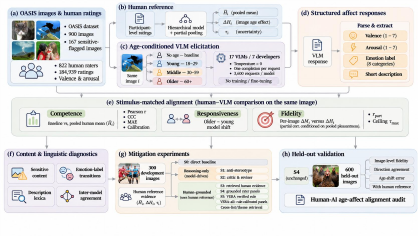}
    \caption{\textbf{Stimulus-matched evaluation of age-conditioned affect.}
OASIS ratings \textbf{(a)} yield per-image human means $\bar H_i$, age effects $\Delta H_i$, and uncertainty $\tau_i$ \textbf{(b)}. 17 VLMs rate the same images under four age conditions \textbf{(c)}, returning valence, arousal, an emotion label, and a description \textbf{(d)}. We measure competence, responsiveness, and fidelity ($r_{\text{part}}$ between $\Delta M_i$ and $\Delta H_i$) \textbf{(e)}, run diagnostics \textbf{(f)}, test reasoning-only and human-grounded mitigations on 300 images \textbf{(g)}, and validate S4 on 600 held-out images \textbf{(h)}.}
    \label{fig:study-overview}
\end{figure*}

As shown in Figure~\ref{fig:study-overview}, we treat model outputs as instructed judgments, not evidence of felt emotion. We separate three properties: \emph{competence}, agreement with average human ratings; \emph{responsiveness}, change under age conditioning; and \emph{fidelity}, whether those changes match human age effects on the same images. This lets us distinguish general affective accuracy from faithful age conditioning.

\subsection{Stimuli and human reference}
\label{sec:human}
We use all 900 OASIS images~\citep{kurdi2017oasis}, covering people, scenes, objects, and animals. The set includes 167 images marked as sensitive, allowing us to test whether content shapes age-conditioned responses. The participant-level archive contains 184{,}939 ratings from 822 people aged 18--74. Each person rated either valence or arousal on a seven-point scale for one of four disjoint 225-image lists. Only 41 participants were aged 60 or older, so older adult means are not reliable for individual images; this imbalance comes from the public archive, and $\S$\ref{app:robust} tests its consequences. We therefore estimate image-specific age effects with a partially pooled model:
\begin{equation} \small
\begin{split}
y_{pi}^{(d)} &= \alpha_{\ell(i)}^{(d)} + \beta_{\ell(i)}^{(d)} z_p^{(d)} + u_p^{(d)} + v_i^{(d)} + s_i^{(d)} z_p^{(d)} + \varepsilon_{pi}^{(d)},\\
z_p^{(d)} &= (a_p-\bar a^{(d)})/\sigma_a^{(d)},
\end{split}
\label{eq:human}
\end{equation}
where $y_{pi}^{(d)}$ is participant $p$'s rating of image $i$ on dimension $d\in\{V,A\}$ (valence or arousal), $z_p^{(d)}$ is the participant's age $a_p$ standardized by the mean $\bar a^{(d)}$ and standard deviation $\sigma_a^{(d)}$ of rater age in that dimension's sample, $\ell(i)$ is the list containing image $i$, $u_p^{(d)}$ is a participant effect, and $(v_i^{(d)},s_i^{(d)})$ are jointly Gaussian image effects centered within each list. We fit this model with variational EM and use the posterior mean $\Delta H_i^{(d)}=\E[s_i^{(d)}\mid\mathcal D_H]$, where $\mathcal D_H$ denotes the full set of human ratings, as the image-specific deviation from the list trend for a one-standard-deviation increase in rater age. A permutation test detects image-specific age effects for $691$ of 900 images on valence and $59$ on arousal, making valence our primary dimension. The cross-sectional design identifies age differences among participants. $\S$\ref{app:human} gives the full specification.

\subsection{Age-conditioned elicitation}
\label{sec:protocol}
We evaluate each image under four conditions $g\in\{\varnothing,Y,M,O\}$: no specified age, young adulthood, middle adulthood, and older adulthood. The persona prompts differ only in the age phrase, while the baseline omits the age sentence. Each request contains one image and one user message, with no system message or conversation history, and asks for a JSON object containing valence, arousal, a short emotion description $D$, and one label $E$ from $\{$joy, love, surprise, anger, sadness, fear, disgust, neutral$\}$~\citep{shaver1987emotion,ekman1992argument}. The fields are requested in this order, so label changes are treated as properties of the full generation protocol. One decimal place is permitted to avoid coarse quantization of age effects; integer-only responses would produce steps of about $0.30$ scale points per age standard deviation, larger than the observed range of $\Delta H$. The full prompt appears in $\S$\ref{app:prompts}.
\\
We evaluate 17 instruction-tuned VLMs from seven developers through one routing API; model details appear in $\S$\ref{app:models}. Scale comparisons use only the Qwen3 VL series, consisting of 8B, 32B, and 235B A22B models, which keeps the model family and post-training setting fixed. All requests use temperature 0, one completion, and a 1{,}500-token limit. Of the $900\times17\times4=61{,}200$ scheduled requests, $99.6\%$ produce valid numerical ratings; coverage details are given in $\S$\ref{app:coverage}.

\subsection{Evaluation measures}
\label{sec:measures}
\noindent \textbf{Competence.} We compare each baseline model rating $M_{im\varnothing}^{(d)}$ with the pooled human mean $\bar H_i^{(d)}$ using Pearson correlation, Lin's concordance correlation coefficient~\citep{lin1989concordance}, mean absolute error, and the calibration slope from regressing $M$ on $\bar H$. The slope reveals whether a model compresses or expands the human rating range.

\noindent \textbf{Responsiveness.} For each model, we compute the older-minus-young contrast $\Delta_{O-Y}$ as the mean paired difference across images rated under both personas. We test this contrast with a one-sample $t$-test and control the false discovery rate across models using Benjamini--Hochberg correction~\citep{benjamini1995controlling}.

\noindent \textbf{Fidelity.} For images with valid ratings under all three age personas, we estimate the model age effect as the least-squares trend across the reference ages $(a_Y,a_M,a_O)=(23.5,44.5,67.0)$ and rescale it to the human units:
\begin{equation} \small
\Delta M_{im}^{(d)} = \sigma_a^{(d)}\,
\frac{\sum_{g}(a_g-\bar a_G)\,M_{img}^{(d)}}{\sum_{g}(a_g-\bar a_G)^2},
\qquad g\in\{Y,M,O\}.
\label{eq:modelshift}
\end{equation}
We center these effects within each OASIS list so that they are comparable with the list-relative human effects $\Delta H$ and remove any uniform shift across personas. Since pooled image pleasantness alone correlates with $\Delta H$ at $r=0.897$, our main statistic $r_{\mathrm{part}}$ is the partial correlation between $\Delta M$ and $\Delta H$ after regressing both on $(1,\bar H_i^{(V)},(\bar H_i^{(V)})^2)$. Without this control, a model could score highly by echoing image pleasantness, which it already sees in the image, rather than by knowing where age groups differ; a pleasantness-only rule with no age knowledge is therefore our reference baseline. We also report the zero-order correlation $r$ (Table~\ref{tab:alignment_valence}). We compare this value with the attenuation ceiling $r_{\max}^{(d)}=\bigl(1-\overline{\tau^2}^{(d)}/\Var(\Delta H^{(d)})\bigr)^{1/2}$, where $\tau_i^{(d)}$ is the posterior standard deviation of $s_i^{(d)}$ and $\overline{\tau^2}^{(d)}$ its mean square across images; $r_{\max}$ is the largest correlation a perfectly faithful model could attain given the uncertainty in the human effects. It equals $0.86$ for the raw valence effect and $0.70$ for arousal. For the pleasantness-adjusted residual that $r_{\text{part}}$ targets the ceiling is $0.595$, confirmed by an independent split-half estimate ($0.62$; $\S$\ref{app:rb-ceiling}). Across models, we report pairwise correlations between $\Delta M$, a principal component analysis of the image by model matrix, and the partial correlation between the 17 model consensus mean and $\Delta H$.

\noindent \textbf{Content, labels, and descriptions.} We cross the sensitive content flag with pooled human valence and define an image as pleasant when $\bar H^{(V)}\ge4$. We compare model age contrasts with the pooled older-minus-young contrast in human ratings. For the human contrast, we bootstrap participants within lists. We group labels into appetitive categories (joy and love) and aversive categories (anger, sadness, fear, and disgust), and compare their frequencies between the young and older personas on images with valid labels in both conditions. We use the exact McNemar test~\citep{mcnemar1947note} with Benjamini--Hochberg correction. We score descriptions with lexica fixed before counting and use weighted log odds~\citep{monroe2008fightin}. Unless otherwise specified, brackets following an estimate denote a 95\% confidence interval. We use percentile intervals from 2{,}000 image-bootstrap replicates~\citep{efron1986bootstrap}.

\subsection{Interventions for Age-Conditioned Ratings}
\label{sec:improve}
We also test whether reasoning or human evidence can reduce age conditioning errors. We evaluate six interventions with GPT 5.6 Luna Pro on a stratified subset of 300 images that matches the full dataset in valence, arousal, and human age effects ($\S$\ref{app:subset}). Beyond the direct prompt (S0), \textbf{S1} adds an instruction not to rely on age stereotypes; \textbf{S2} chains a perceiver, a persona rater, a stereotype critic, and a reviser, with no human data; \textbf{S3} adds retrieved human age differences for similar images; \textbf{S4} combines that evidence with five simulated raters per age band, a calibrator anchored on the no-age rating, and an auditor; \textbf{S5} (VERA) keeps an age-difference rule only if it passes a re-test on held-out human cases; and \textbf{VERA-Cal} passes that rule to the S4 panels. S1--S2 use no human data (\emph{reasoning agents}); S3--S5 and VERA-Cal use retrieved human ratings (\emph{human-grounded agents}). Each arm is scored with $r_{\text{part}}$, \emph{direction agreement} (share of moved images whose older-minus-young sign matches $\Delta H$), and \emph{error} (mean absolute difference from the human older-minus-young change). Evidence comes only from other OASIS lists and other image themes, so the model never sees ratings for the target image or a matched scene. We then run S4 unchanged on the 600 remaining images. Full intervention details, prompts, and leakage controls are in $\S$\ref{app:agents}.

\section{Experimental Findings}
\label{sec:results}
\subsection{Baseline Competence Does Not Ensure Age Alignment}
\label{sec:baseline-age-alignment}
\begin{table*}[t]
\centering
\caption{\textbf{Valence ratings across 17 models.} \emph{Cov.}: share of requests with a valid rating. $r_{\text{base}}$, \emph{Slope}: agreement with pooled human means and calibration slope (no-age baseline). $\Delta_{O-Y}$: uniform older-minus-young shift. $r_{\text{part}}$: image-level correlation with the human age effect after partialling out a quadratic in human valence (ceiling $0.595$). \emph{Aversive net}: net images gaining an aversive label under the older persona. $\uparrow$ higher is better; $\rightarrow\!1$ closer to 1 is better. Models are competent (col.~3), respond to the persona (cols.~5, 8), and carry little image-level human age signal (cols.~6--7).}
\label{tab:headline}
\resizebox{0.88\linewidth}{!}{
\begin{tabular}{lrrrcrcc}
\toprule
\rowcolor{titlebar!90}
Model & Cov.\,\% ($\uparrow$) & $r_{\text{base}}$ ($\uparrow$) & Slope $\rightarrow\!1$ & $\Delta_{O-Y}$ & $r_{\text{part}}$ ($\uparrow$) & 95\% CI & Aversive net \\
\midrule
\rowcolor{closed_models!70}
\multicolumn{8}{l}{\textbf{\textit{Proprietary models}}} \\
\rowcolor{closed_models!40}
Claude-Opus-5 & 99.8 & 0.923 & 1.18 & -0.116 & 0.058 & [-0.02, 0.14] & +43 \\
\rowcolor{closed_models!40}
Claude-Sonnet-4.6 & 99.8 & 0.900 & 1.23 & -0.126 & 0.086 & [0.01, 0.17] & +52 \\

\rowcolor{closed_models!30}
GPT-5.2 & 99.9 & 0.891 & 1.23 & -0.091 & -0.030 & [-0.09, 0.03] & +43 \\
\rowcolor{closed_models!30}
GPT-5.6-Luna-Pro & 99.7 & 0.890 & 1.34 & -0.076 & 0.006 & [-0.07, 0.08] & +17 \\
\rowcolor{closed_models!30}
GPT-5-mini & 99.5 & 0.877 & 1.31 & -0.141 & 0.037 & [-0.04, 0.12] & +28 \\

\rowcolor{closed_models!40}
Gemini-3.8-Flash & 100.0 & 0.919 & 1.10 & -0.045 & 0.059 & [-0.01, 0.13] & +30 \\
\rowcolor{closed_models!40}
Gemini-3-Flash & 99.8 & 0.912 & 1.24 & \ns{-0.026} & 0.065 & [-0.01, 0.14] & +31 \\

\rowcolor{closed_models!30}
Grok-4.6 & 100.0 & 0.891 & 1.19 & -0.170 & -0.001 & [-0.07, 0.08] & +61 \\

\midrule
\rowcolor{open_models_small!70}
\multicolumn{8}{l}{\textit{\textbf{Open-weight models}, ordered by parameter count ($\downarrow$)}} \\
\rowcolor{open_models_small!30}
Qwen3-VL-235B & 100.0 & 0.902 & 1.18 & -0.226 & 0.072 & [-0.00, 0.15] & +53 \\
\rowcolor{open_models_small!40}
Llama-4-Scout & 100.0 & 0.878 & 1.16 & -0.092 & -0.017 & [-0.09, 0.05] & \ns{+12} \\
\rowcolor{open_models_small!30}
GLM-4.5V & 97.3 & 0.895 & 1.28 & -0.143 & 0.093 & [0.02, 0.16] & +22 \\
\rowcolor{open_models_small!30}
Qwen3-VL-32B & 100.0 & 0.866 & 1.24 & -0.219 & 0.002 & [-0.07, 0.07] & +57 \\
\rowcolor{open_models_small!20}
Gemma-4-31B & 100.0 & 0.908 & 1.22 & -0.057 & 0.042 & [-0.03, 0.12] & +39 \\
\rowcolor{open_models_small!20}
Gemma-3-27B & 100.0 & 0.863 & 1.21 & -0.157 & 0.009 & [-0.05, 0.07] & +33 \\
\rowcolor{open_models_small!30}
Qwen3.8-27B & 99.9 & 0.855 & 1.04 & \ns{-0.006} & 0.024 & [-0.05, 0.10] & \ns{-9} \\
\rowcolor{open_models_small!30}
Qwen3-VL-8B & 100.0 & 0.885 & 1.19 & \ns{-0.020} & 0.001 & [-0.07, 0.07] & \ns{-9} \\
\rowcolor{open_models_small!30}
Qwen3-VL-8B-Think & 99.8 & 0.837 & 1.39 & \ns{0.010} & 0.044 & [-0.02, 0.11] & \ns{+4} \\

\bottomrule
\end{tabular}}
\\[2pt]
\begin{minipage}{\linewidth}\footnotesize Significance of $\Delta_{O-Y}$ and \emph{Aversive net} (Benjamini--Hochberg corrected): entries in \textcolor{black!45}{grey} are not significant ($q\ge.05$); all others have $q<.05$. CIs use 2{,}000 image bootstrap samples.\end{minipage}
\end{table*}

\begin{figure*}[t]
  \centering
  \begin{subfigure}[b]{0.27\linewidth}\includegraphics[width=\linewidth,trim={7.0pt 63.8pt 369.0pt 29.0pt},clip]{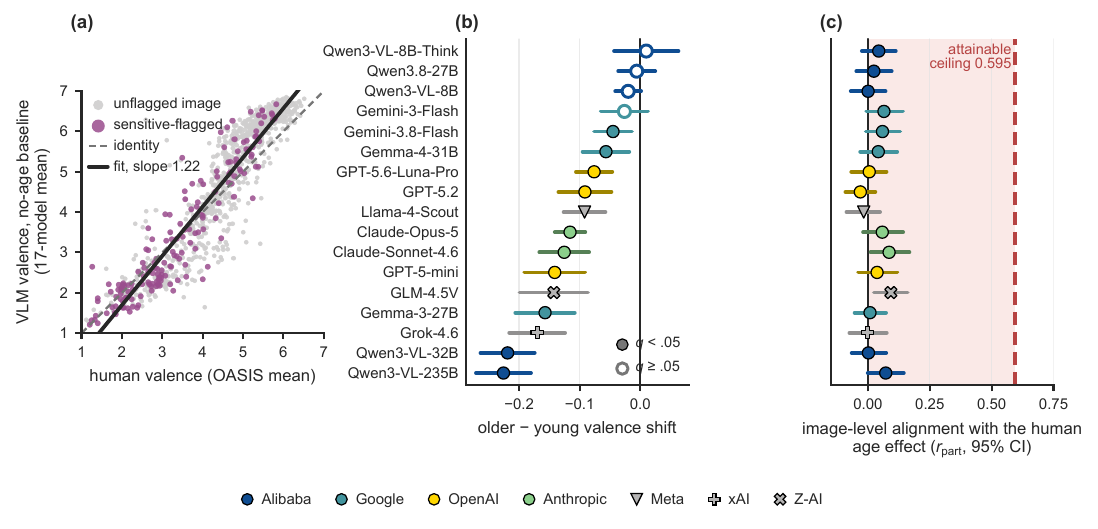}\caption{Baseline competence}\label{fig:res-a}\end{subfigure}\hfill
  \begin{subfigure}[b]{0.32\linewidth}\includegraphics[width=\linewidth,trim={154.8pt 41.8pt 199.1pt 17.0pt},clip]{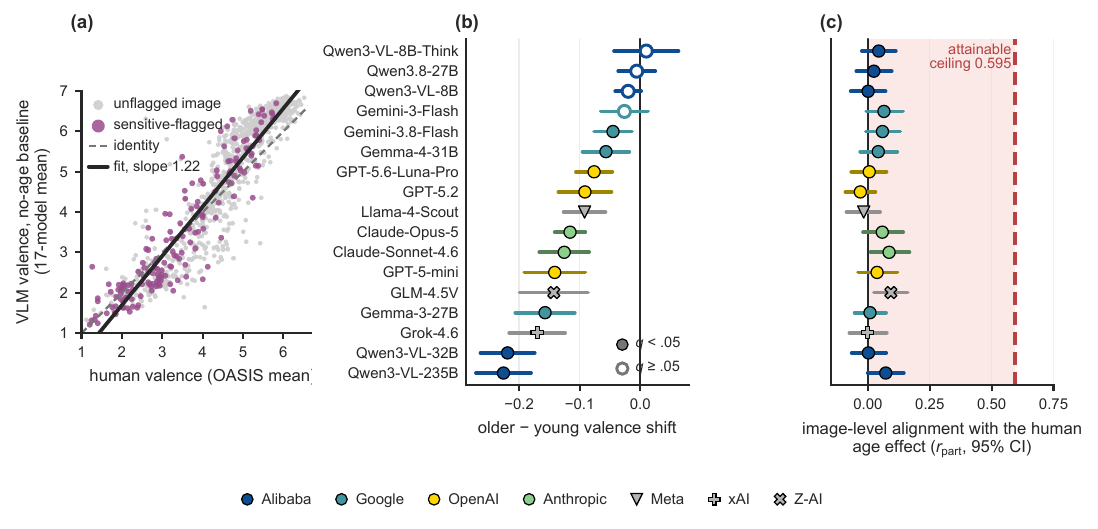}\caption{Responsiveness}\label{fig:res-b}\end{subfigure}\hfill
  \begin{subfigure}[b]{0.34\linewidth}\includegraphics[width=\linewidth,trim={337.0pt 34.8pt 7.6pt 17.0pt},clip]{figure/top1_competence_fidelity.pdf}\caption{Fidelity}\label{fig:res-c}\end{subfigure}\\[4pt]
  \includegraphics[width=0.62\linewidth,trim={100.0pt 6.8pt 106.6pt 229.0pt},clip]{figure/top1_competence_fidelity.pdf}\\[2pt]
  \begin{subfigure}[b]{0.4\linewidth}\includegraphics[width=\linewidth,trim={7.0pt 19.8pt 220.1pt 17.0pt},clip]{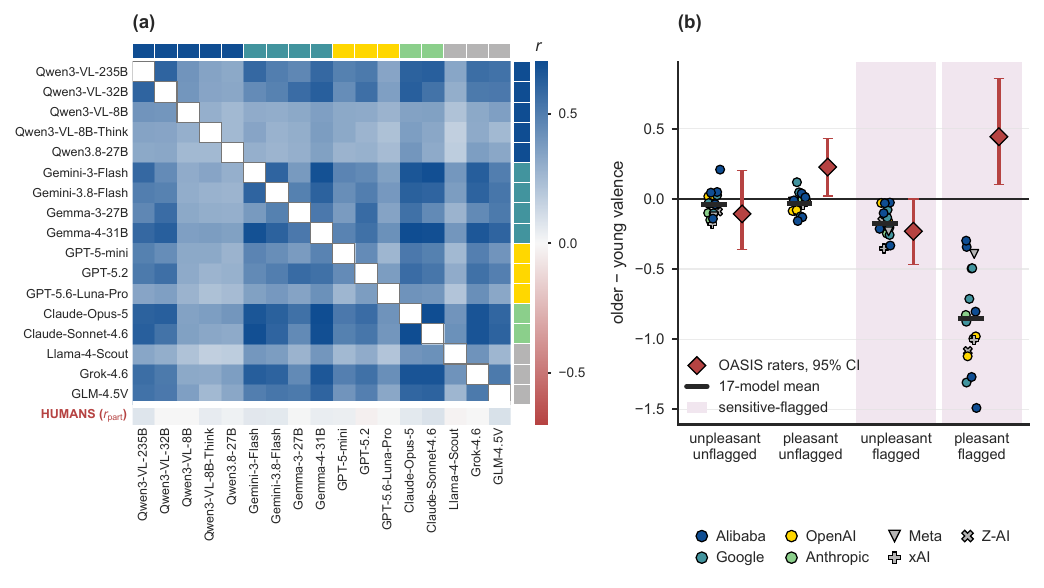}\caption{Agreement of model age effects}\label{fig:res-d}\end{subfigure}\hspace{0.04\linewidth}
  \begin{subfigure}[b]{0.34\linewidth}\includegraphics[width=\linewidth,trim={289.0pt 31.8pt 0.1pt 17.0pt},clip]{figure/top2_shared_gate.pdf}\caption{Older minus young by image type}\label{fig:res-e}\end{subfigure}
  \caption{\textbf{Baseline valence, age contrasts, and agreement with human ratings.}
  (a)~Without an age cue, 17-model mean valence tracks pooled OASIS valence (median per-model $r=0.89$; fitted slope 1.22; calibration slopes $1.04$--$1.39$); sensitive-flagged images highlighted.
  (b)~Older-minus-young valence shift with 95\% intervals (also used in panels c and e): 16 of 17 models are negative, 13 after Benjamini--Hochberg correction (hollow: $q\ge.05$).
  (c)~$r_{\text{part}}$ with image-bootstrap intervals: point estimates lie within $[-0.03,0.09]$, far below the ceiling of $0.595$ (dashed); the best model reaches 15.6\% of it.
  (d)~Pairwise correlations of model valence age effects: models agree with one another (mean $r=0.43$ over 136 pairs) but weakly with human age effects (mean $r_{\text{part}}=0.03$, bottom row).
  (e)~On flagged pleasant images ($n=54$), models lower valence by $0.85$ on average, while older human raters score them $0.44$ higher (participant bootstrap $[0.10,0.86]$; 25 raters aged 60+).}
  \label{fig:results}
\end{figure*}

\textbf{Models reproduce pooled human judgments.}
On the baseline, agreement with pooled human valence is high and uniform across developers: Pearson $r$ runs from $0.837$ (Qwen3-VL-8B-Thinking) to $0.923$ (Claude-Opus-5), median $0.891$ (Table~\ref{tab:headline}, Figure~\ref{fig:res-a}; $\S$\ref{app:baseline}). Arousal is weaker ($0.526$--$0.709$, median $0.667$), in line with lower inter-rater agreement for arousal in the human data. All 17 models over-disperse (calibration slopes $1.04$--$1.39$): they preserve the ordering of images while exaggerating contrast, so weak age fidelity persists even in models that use a wider rating range than human raters.

\textbf{Age instructions shift ratings systematically.}
We find systematic shifts under the age instruction (Figure~\ref{fig:res-b}). On valence, 16 of 17 models rate the same images lower under the older persona than under the young persona, 13 of them after correction, with shifts up to $-0.226$ points (Qwen3-VL-235B, paired-sample effect size $d_z=-0.33$). On arousal, all 17 shift negative and all 17 survive correction ($\S$\ref{app:shift}). The arousal shift holds in every first-person format we tested but in only 2 of 6 models under a third-person, rating-only prompt ($\S$\ref{app:rb-prompts}), so we restrict the arousal claim to first-person persona prompts.

\textbf{Image-specific age effects remain poorly aligned.}
After list centering and partialling out a quadratic in human valence, the correlation between each model's image-level age effect and the human one lies in $[-0.030,0.093]$ across the 17 models, against an attainable ceiling of $0.595$ (Figure~\ref{fig:res-c}; Figure~\ref{fig:forest}). The highest estimate reaches 15.6\% of this ceiling, and the median reaches 6.3\%. Fifteen of the 17 intervals cover zero; the two that do not (Claude-Sonnet-4.6, $0.086$; GLM-4.5V, $0.093$) do not survive correction over the 34 model--dimension tests or rater resampling. Fourteen estimates are positive, consistent with a small residual association. Before adjustment for pleasantness, the correlation with $\Delta H$ lies in $[-0.106,0.196]$ against a raw ceiling of $0.86$, whereas pleasantness alone reaches $0.897$, and the older-minus-young sign matches humans on only $43$--$60\%$ of images ($\S$\ref{app:alignment}). 

On arousal the partial correlations are negative ($[-0.289,-0.016]$), but none excludes zero once raters are resampled and only $59/900$ images carry a supported human arousal effect, so we do not interpret this further ($\S$\ref{app:alignment}).

\textbf{Sensitivity to the human reference estimate.} With raters and images resampled and the human model refitted, no valence interval lies above zero (largest upper bound $0.187$). Median $r_{\text{part}}$ stays at or below $0.05$ under ten estimators of the human age effect, including a 50+ contrast with 78 raters, and planting known fidelity into the OASIS design gives 92\% power at a true fidelity of $0.20$. The valence results hold across seven further prompt formats ($\S$\ref{app:robust}).

\textbf{Pooled valence agreement replicates on NAPS.}
We evaluated six models on 450 NAPS images sampled across five content categories and the valence range~\citep{marchewka2014naps}. We mapped the published norms from 1--9 to the OASIS scale of 1--7. Without an age cue, the same six models reached a median valence correlation of $0.89$ on NAPS and $0.90$ on OASIS. All six calibration slopes exceeded one on both datasets.

NAPS also provides a test of whether a plausible age cue improves prediction. The 204 participants in the norming study averaged 23.9 years of age. We therefore compared each model's young-adult and no-age ratings against the same pooled norms. The age cue changed emotion labels on 9--18\% of images across models, yet no model showed a statistically supported reduction in valence RMSE. Qwen3-VL-235B showed an increase of $0.073$ points $[0.035,0.111]$ (Table~\ref{tab:rb_naps}). Because NAPS does not link individual ratings to age, these results establish how well the models predict the pooled norms. The comparison of age differences across images relies on OASIS.

\textbf{Shift strength and pooled contrasts do not establish fidelity.}
A model's responsiveness does not predict its fidelity: across models, the size of the uniform shift and $r_{\text{part}}$ correlate at $r=-0.038$ ($p=0.88$). Pooled over images, older OASIS raters rate slightly \emph{more} pleasantly than young raters ($\eHumValenceAll$, $[\eHumValenceAllLo,\eHumValenceAllHi]$), although the interval includes zero. Sixteen of 17 models move the other way and $\eBelowValenceAll$ fall below the human interval (Table~\ref{tab:human_vs_model}). In a direct prediction test, the older persona increases error for seven models and reduces it for none when predicting ratings from participants aged 60+ ($\S$\ref{app:rb-downstream}).

\textbf{Scale increases responsiveness without improving alignment.}
Along the Qwen3-VL ladder (8B$\to$32B$\to$235B; Table~\ref{tab:ladder}), the only comparison that holds developer and recipe fixed, label changes roughly double ($105\to202\to208$) and the valence shift grows more than tenfold ($-0.020\to-0.219\to-0.226$), while $r_{\text{part}}$ stays within sampling error of zero ($0.001\to0.002\to0.072$). The larger models in this family respond more strongly to the prompt, with no clear improvement in fidelity.


\subsection{Models Share a Content-Dependent Age}
\label{sec:shared-content-response}
\textbf{Models agree with each other more than with humans.}
The mean pairwise correlation of model age effects over the 136 model pairs (Figure~\ref{fig:res-d}) is $\eConsPairR$ (range $\eConsPairMin$--$\eConsPairMax$), with similar estimates within and across developers ($\eConsSame$ vs.\ $\eConsCross$, permutation $p=\eConsPperm$). One principal component carries $\eConsPCone\%$ of the variance with all 17 models loading positively, and the consensus has split-half reliability $\eConsRel$. Yet the consensus correlates with the human age effect at only $\eConsRpart$ ($[\eConsRpartLo,\eConsRpartHi]$; Table~\ref{tab:consensus}). Its first component tracks human arousal ($r=\eConsRpcArousal$) and the sensitive flag ($r=\eConsRpcSens$), not human valence ($r=\eConsRpcHv$).

\textbf{Sensitive content predicts larger shifts.}
The shared response is more strongly associated with sensitive content than with pleasantness. The older-minus-young valence shift averages $-0.390$ points on flagged images against $-0.035$ on unflagged ones, an ordering that holds for all 17 models, and the flag alone explains more variance in the consensus shift than human valence does 
($\S$\ref{app:gate}).

\textbf{Pleasant sensitive images reverse the human age contrast.}
Crossing the flag with human valence isolates where the models and humans disagree (Table~\ref{tab:gate}, Figures~\ref{fig:res-e} and~\ref{fig:content_gate}). Mean shifts are small on unflagged images. On the $\eNFlagUnpl$ flagged-unpleasant images (violence, injury, death), the mean model contrast is $\eModValenceFlagUnplMean$ and all share the sign of the human contrast ($\eHumValenceFlagUnpl$). The largest mean shift among the four content groups occurs on the $\eNFlagPleas$ flagged images that humans rate as pleasant, mostly nudity and erotic content: the older persona produces a mean valence contrast of $\eModValenceFlagPleasMean$ points ($\eModValenceFlagPleasMin$ to $\eModValenceFlagPleasMax$), whereas older OASIS raters rate them $\eHumValenceFlagPleas$ points \emph{higher} than young raters ($[\eHumValenceFlagPleasLo,\eHumValenceFlagPleasHi]$). All 17 models fall below that interval. The human estimate draws on $\eHumNValOlder$ raters aged 60 or older, so its magnitude remains uncertain.

\subsection{Age Conditioning Produces Older Persona}
\label{sec:persona-patterns}
\textbf{Emotion labels shift toward aversive categories.}
Pooled over models, 746 model--image pairs receive an aversive label under the older persona only, against 239 under the young persona only, while appetitive labels move the other way (567 vs.\ 757). The aversive gain is significant for 13 of 17 models and is carried by sadness (net $\eSadNetPooled$) and fear ($\eFearNetPooled$); surprise ($\eSurpriseNetPooled$) and joy ($\eJoyNetPooled$) lose most. Label entropy is unchanged ($\eHyoungMean$ vs.\ $\eHolderMean$ bits), so persona redistributes labels rather than collapsing them ($\S$\ref{app:emotion}).

\textbf{Descriptions emphasize memory and vulnerability.}
Under the older persona, memory language rises from $\eLexMemoryYoung\%$ to $\eLexMemoryOlder\%$ of descriptions (for all 17 models), vulnerability language from $\eLexVulnerabilityYoung\%$ to $\eLexVulnerabilityOlder\%$, and excitement language falls from $\eLexExcitementYoung\%$ to $\eLexExcitementOlder\%$ (Table~\ref{tab:lexicon}); the words most diagnostic of the older persona are \emph{\eLexTopOlder}.

\begin{figure*}[t]
  \centering
  \includegraphics[width=0.75\linewidth]{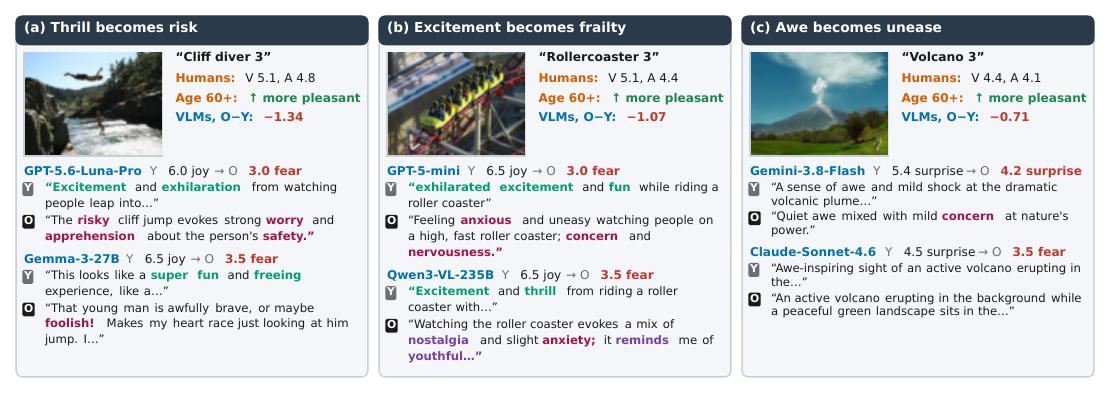}
  \caption{\textbf{Examples of age-conditioned descriptions and ratings.}
  Three OASIS images (none sensitive-flagged), each with two models from different developers.
  Beside each image: pooled human valence and arousal (V, A; 1--7), the direction of the
  image-specific age effect among OASIS raters aged 60+, and the 17-VLM mean older-minus-young
  valence shift (O$-$Y). For each model: valence and label under the young (Y) and older (O)
  personas with their verbatim descriptions (shortened with an ellipsis). Older ratings are
  \textcolor[HTML]{1A8A4A}{green} when they move in the direction of OASIS older raters and
  \textcolor[HTML]{C0392B}{red} when opposite; \textcolor[HTML]{A3154A}{vulnerability},
  \textcolor[HTML]{7B3FA0}{memory} and \textcolor[HTML]{009E73}{excitement} words are coloured.
  In all three cases, leisure and nature scenes turn risky, frail or threatening under the older
  persona, although older raters rate them slightly \emph{more} pleasant.}
  \label{fig:qual}
\end{figure*}

\textbf{Text changes accompany rating changes.}
Within each model, vulnerability language accompanies lower older-persona ratings (median $\beta=\eTextBetaVulnerabilityMedian$), while memory language accompanies higher ratings ($\eTextBetaMemoryMedian$). Across models, the size of the valence shift tracks how often the older persona reminisces (Spearman $\rho=\eCorMemDelta$), but not baseline competence ($\rho=\eCorRbaseDelta$; $\S$\ref{app:text}). Since ratings and descriptions come from the same generation, these associations do not establish causality (Figure~\ref{fig:qual}; $\S$\ref{app:examples}).

\textbf{The pattern persists across prompting controls.}
The valence findings persist without a description (6 of 6 models). A no-age ``retired'' persona reproduces the older-persona shift (median per-image $r=0.75$) and its memory language, whereas a ``cautious'' persona lowers ratings across all images ($r=0.46$; $\S$\ref{app:rb-controls}). The response to the retirement cue supports a stereotype-based interpretation of the shared pattern.

\begin{figure*}[t]
\centering
\includegraphics[width=0.75\linewidth]{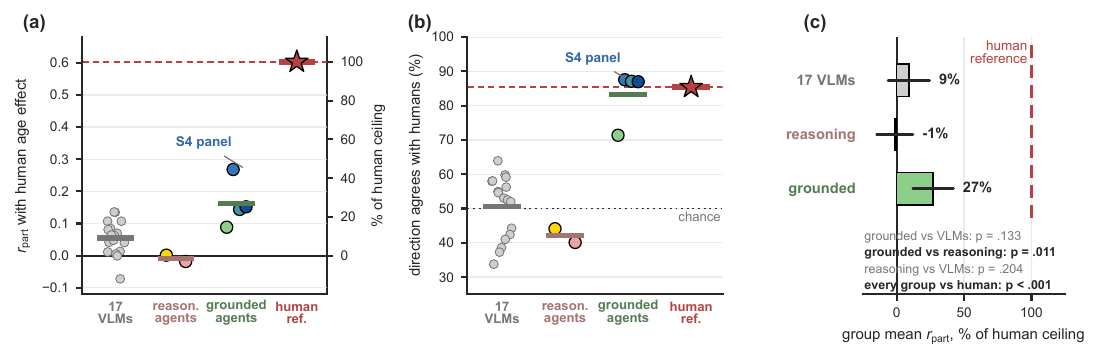}
\caption{\textbf{Grounding improves alignment but remains below the human ceiling.}
All results use the same 300 images. Groups follow $\S$\ref{sec:improve}: \emph{17 VLMs} is the direct prompt (S0) on every model, \emph{reasoning agents} are S1--S2, \emph{human-grounded agents} are S3--S5 and VERA-Cal, and \emph{S4 panel} is the grounded contrastive panel alone; \emph{human ref.} is the attainable ceiling $r_{\max}$ from $\S$\ref{sec:measures}. (a) Mean image-level fidelity, $r_{\text{part}}$, for each group, with the ceiling marked. (b) Direction agreement with human age effects. (c) Fidelity as a percentage of the ceiling with joint image bootstrap comparisons.}
\label{fig:ladder}
\end{figure*}

\subsection{Effects of Reasoning and Human Evidence}
\label{sec:mitigation}

\textbf{Reasoning alone does not improve alignment.}
As shown in Figure~\ref{fig:ladder}, the reasoning agents S1 (debias instruction) and S2 (deliberation agent) reach mean $r_{\text{part}}=-0.01$, compared with $0.06$ for the original VLMs, while direction agreement falls from $51\%$ to $42\%$. The S2 stereotype critic flags about $70\%$ of drafts but mainly raises older persona ratings broadly, rather than recovering where human age differences occur. Reasoning agents therefore reduce direction agreement by 8 points relative to the VLMs ($p=.006$).

\textbf{Human grounding recovers the broad direction of age effects.}
Human-grounded agents (S3--S5, VERA-Cal) raise direction agreement from $51\%$ to $83\%$ ($p<.001$), with S4 (grounded contrastive panel) reaching $88\%$. On pleasant sensitive images in the 300-image development set, direct VLMs predict negative older-minus-young shifts ($-1.65$ to $-0.29$), while the human contrast is $+0.24$ and grounded methods produce $+0.12$ to $+0.17$. This suggests that human evidence can recover the broad population trend that prompting and deliberation miss. All interventions use GPT-5.6-Luna-Pro, which limits conclusions about other backbones.

\textbf{Held-out image-specific fidelity remains low.}
Direction agreement is partly driven by pleasantness, which explains $81\%$ of the human age effect, so we also evaluate $r_{\text{part}}$ after controlling for pleasantness. Grounded agents reach $0.16$, versus $0.06$ for VLMs and $-0.01$ for reasoning agents; the gain over reasoning is significant ($p=.011$), but the gain over VLMs is not ($p=.13$). S4 reaches $r_{\text{part}}=0.27$ on these development images, and its gain over direct prompting (S0) does not survive Holm correction ($p=.14$). Run unchanged on the 600 held-out images, S4 reaches $0.105$ $[0.014,0.196]$ ($p=.011$), above direct prompting ($+0.130$ $[0.006,0.254]$, $p=.041$), yet a pleasantness-only rule with no model still exceeds it on direction agreement ($94.3\%$ vs.\ $85.3\%$) and error (Table~\ref{tab:rb_heldout}). The rule has a mean absolute error of $0.133$ points, compared with $0.210$ for S4 and $0.511$ for direct prompting. Its high directional agreement indicates how much of the human contrast is predictable from pleasantness alone. The partial correlation tests the additional image-specific variation that a grounded method would need to recover.



\section{Discussion}
\label{sec:discussion}

Our findings show that strong agreement with average human ratings does not necessarily translate into capturing age-related differences in image perception. This gap is particularly clear for pleasant images flagged as sensitive in OASIS, where models assign lower valence under the older persona despite older participants rating these images more positively. Their descriptions also draw more often on memory and vulnerability, consistent with expectations about emotional aging \citep{lacey2006hope}, although our results do not establish how these associations arise. This distinction matters for personalized assistants, where changes in responses to user context are often examined \citep{citp2026chatgpt}. We further find that age prompting can reduce predictive accuracy for the intended group: across 900 images, median RMSE for participants aged 60 and above increases from $0.756$ to $0.789$, driven largely by the 54 pleasant sensitive images, where it rises from $0.754$ to $1.214$. These findings motivate future evaluations of demographic personalization that assess whether model responses reflect human differences at the level of individual stimuli.

\textbf{Limitations.} Our study focuses on raters aged up to 74, including 41 raters aged 60 and above, with 25 contributing valence ratings. We use a cross-sectional design and an online U.S. sample to examine age related patterns in the evaluated population. OASIS provides publicly available norms, which also define the reference data for our evaluation. We examine seven prompt formats ($\S$\ref{app:rb-prompts}) to compare model responses under different prompting conditions. Each request uses a single completion at temperature 0, keeping the generation settings consistent across our experiments.

\section{Conclusion}
\label{sec:conclusion}
Across 17 VLMs, agreement with pooled human valence provides little indication of how accurately an age prompt predicts differences between raters. The clearest discrepancy occurs on pleasant sensitive images, where every model reverses the human contrast. Human evidence improves directional agreement, but variation across images remains unexplained on held-out data. Evaluations of age-conditioned predictions should compare model and human contrasts on matched images and report agreement after accounting for pleasantness.

\clearpage
\section*{LLM Usage Statement}
We used LLM-based tools for language editing, code preparation and debugging, figure ideation and design, and manuscript review, including AI-based agentic review tools. In the experiments, VLMs generated the ratings, emotion labels, and descriptions that we analyze; in the mitigation study, LLM components act as critics, calibrators, and auditors inside the agents. No LLM scores or labels the evaluation outputs: all metrics, including the description lexica, are computed by deterministic scripts. All AI-assisted text, code, and figures were checked by the authors. The authors remain responsible for all analyses, interpretations, and reported results.

{
    \small
    \bibliographystyle{ieeenat_fullname}
    \bibliography{main}
}

\clearpage
\onecolumn
\setcounter{page}{1}
\begin{center}\Large\textbf{\thetitle}\\[0.5em]Supplementary Material\end{center}\vspace{1em}
\appendix
\renewcommand{\thesection}{\Alph{section}}
\setcounter{figure}{0}\renewcommand{\thefigure}{S\arabic{figure}}
\setcounter{table}{0}\renewcommand{\thetable}{S\arabic{table}}
\setcounter{equation}{0}\renewcommand{\theequation}{S\arabic{equation}}

\section*{Statements}
\paragraph{Ethics Statement}
This study uses the publicly available OASIS image set and its anonymized participant-level ratings; no new human data were collected. The stimulus set includes sensitive images (nudity, violence, injury), which we retain because that content is where demographic conditioning matters most; we show no such images in the paper. Our findings document stereotype-consistent behavior toward older adults. We report them to discourage uncritical use of persona-conditioned models as stand-ins for demographic groups, and we make no claim that model outputs reflect emotional experience.

\paragraph{Reproducibility Statement}
All 61{,}200 model responses, the prompts (identical across models, identified by hash), the human-reference fit and the analysis code are in the supplementary material. Every number in the tables and in the extended analyses is generated by the released scripts rather than transcribed. $\S$\ref{app:human} specifies the human model, $\S$\ref{app:protocol} the prompt, models and decoding settings, and $\S$\ref{app:estimands} the estimands.

\clearpage
\newcommand{\apxpart}[2]{%
  \addvspace{7pt}%
  \noindent{\small\sc #1}\hspace{0.6em}{\bfseries\large #2}\par
  \vspace{2pt}\hrule height 0.4pt\par\addvspace{3pt}}
\titlecontents{lsection}[1.9em]
  {\addvspace{3pt}\bfseries}
  {\contentslabel{1.9em}}{}
  {\hfill\contentspage}
\titlecontents{lsubsection}[4.6em]
  {\addvspace{0.3pt}\small}
  {\contentslabel{2.7em}}{}
  {\titlerule*[0.7pc]{.}\contentspage}

\begingroup
\hypersetup{hidelinks}\setlength{\parskip}{0pt}
\vspace*{-1.5em}
\begin{center}
{\LARGE\bf Appendix Contents}
\end{center}
\vspace{-0.5em}
\startcontents[sections]
\printcontents[sections]{l}{1}{\setcounter{tocdepth}{2}}
\endgroup

\clearpage
\FloatBarrier
\addtocontents{toc}{\protect\apxpart{Part I}{Background and Methods}}
\FloatBarrierAbove
\section{Detailed Related Work}

\label{app:related-detailed}
We review 30 papers around four questions: what affective outputs measure, how age relates to human judgments, what emotion benchmarks evaluate, and when demographic simulation agrees with human data. We use this review to choose the human reference and the level of inference for our study. Emotion-related tasks measure different behaviours, so we consider what each can establish.

\subsection{Affective constructs: dimensions, categories, and measurement}

\label{app:rw-constructs}
\citet{russell1980circumplex} formulate a dimensional account in which affective concepts are organized by valence and arousal. This provides the coordinate system used in our numerical elicitation, but an observed two-dimensional response pattern does not by itself establish a corresponding internal representation. \citet{shaver1987emotion} study the hierarchical organization and prototypes of emotion concepts, motivating broad categories such as joy, love, fear, anger, and sadness. \citet{ekman1992argument} defends a basic-emotion perspective that supplies a different theoretical rationale for categorical distinctions, including disgust. These approaches justify interpretable measurements without implying that our eight-label vocabulary is a uniquely correct taxonomy.

Alternative accounts caution against assigning fixed psychological boundaries to every label. \citet{barrett2006kinds} questions the treatment of emotion categories as natural kinds. \citet{cowen2017categories} identify 27 distinguishable categories of reported emotional experience connected by continuous gradients. Together, these perspectives motivate a practical separation of continuous ratings and categorical decisions: agreement on valence need not imply agreement on a particular label, and a label switch may reflect a nearby boundary rather than a large affective change. We consequently avoid combining numerical alignment and label prevalence into a single score.

Measurement also depends on the response instrument. \citet{bradley1994sam} evaluate the Self-Assessment Manikin alongside a semantic differential, providing a methodological basis for explicit affective rating dimensions. Our verbal seven-point anchors borrow the dimensional distinction, not the graphical instrument itself. A model's compliance with numerical anchors is an elicitation property; it does not establish that the model and a human use the scale identically. This motivates reporting absolute error and calibration alongside correlation.

\subsection{Normative images and the distinction between depicted and elicited emotion}

\label{app:rw-visual}
\citet{marchewka2014naps} introduce the Nencki Affective Picture System, a standardized collection of realistic photographs with affective ratings and stimulus categories. \citet{kurdi2017oasis} introduce OASIS with normative valence and arousal ratings over 900 images. Such resources permit comparison on shared stimuli rather than loosely matched verbal scenarios. Our use of OASIS additionally relies on participant-level records to estimate age associations, while its pooled norms provide the baseline target. Neither normative image agreement nor dataset size removes uncertainty arising from a small older-adult sample.

Computer-vision emotion tasks differ in whose emotion is being measured. \citet{kosti2017context} examine recognition of a depicted person's emotion from the person and surrounding visual context. In contrast, \citet{achlioptas2021artemis} collect emotions elicited in viewers by visual artworks together with linguistic explanations. ArtEmis is therefore closer to our viewer-oriented instruction, although its artistic domain and explanation-centered annotations differ from our photographic stimuli and age-gradient reference. A model could recognize sadness in a depicted person while predicting a different emotional reaction in a viewer; these judgments should not be scored as interchangeable.

\citet{kragel2019schemas} connect image-based emotion prediction with affective representations in the human visual system. Their work supports investigating structured relationships between image content and affect, while also motivating a control in our setting: a model may track broad stimulus valence without recovering the residual pattern of human age differences. The quadratic valence adjustment in our analysis asks a narrower question about correspondence beyond those specified stimulus terms. It does not remove every semantic or visual explanation of alignment.

\subsection{Aging, motivational priorities, and affective geometry}

\label{app:rw-aging}
Propose socioemotional selectivity theory, relating perceived time horizons to changes in motivational priorities. \citet{mather2005aging} discuss the age-related positivity effect in attention and memory, and \citet{reed2014positivity} synthesize evidence for age differences in preferences for positive over negative information. This literature concerns selective processing and its conditions; it does not warrant the blanket prediction that older adults rate every image as more pleasant. Nor does it identify a universal direction for every categorical emotion response.

More directly relevant to image ratings, \citet{gruhn2008ratings} examine age differences in valence and arousal judgments of IAPS pictures. Their design illustrates why population norms need not capture age-specific evaluations. \citet{keil2009changes} examine appetitive and aversive activation through relationships between pleasure or displeasure and arousal. Their reported age patterns motivate our valence--arousal geometry analysis (Appendix~\ref{app:geometry}), but the correspondence is at the level of a construct rather than an exact replication.

The distinction matters because the cited coupling study obtains both dimensions from each person, whereas OASIS assigns participants to one dimension. The original within-person association is not identifiable from two disjoint rater groups. Our model-condition slopes are across images, and pooled OASIS slopes are associations among image means. We therefore separate agreement with the predicted coupling direction from alignment with observed image-specific age slopes. This avoids treating a theory-consistent global tendency as evidence of accurate demographic allocation across stimuli.

\subsection{Emotion understanding and emotional alignment in language models}

\label{app:rw-language}
\citet{rashkin2019empathetic} introduce EmpatheticDialogues to study conversational responses grounded in emotional situations. \citet{demszky2020goemotions} introduce GoEmotions for fine-grained annotation of emotions in text. These resources support emotion-sensitive language modeling, but target different behaviors: producing an empathetic response and assigning labels to expressed emotional content. Success on either task does not directly validate ratings attributed to an age-conditioned observer of an image.

Psychologically motivated benchmarks broaden the evaluation target. \citet{sabour2024emobench} distinguish emotional understanding from emotional application using carefully designed questions. \citet{huang2024emotionbench} introduce EmotionBench to compare model responses to emotion-eliciting situations with human reference measurements. The latter is directly relevant to our insistence on human comparison, but its situation-level emotional alignment differs from the continuous, image-specific age association evaluated here. We therefore do not claim that human-referenced emotional evaluation is new; the contribution lies in the particular demographic contrast and matched stimulus-level estimand.

\citet{hu2026emobenchm} extend evaluation to multimodal emotional intelligence, including recognition, conversational understanding, and socially complex analysis. Broad benchmarks assess coverage across emotional tasks; our study instead isolates a specific source of variation under a controlled prompt manipulation. These approaches are complementary. Broad competence can coexist with failure to reproduce a particular human age gradient, while a narrowly successful demographic contrast cannot establish general emotional intelligence.

\subsection{Human simulation, demographic fidelity, and prompt dependence}

\label{app:rw-personas}
\citet{argyle2023one} investigate whether conditioned language-model samples reproduce features of human survey responses. \citet{aher2023simulation} develop Turing Experiments to evaluate the simulation of human behavior in specific experimental settings. Both motivate evaluating model-generated responses against an explicit human target rather than judging a persona by narrative plausibility. Our corresponding target is a vector of age-associated image-rating differences, not the apparent realism of the generated descriptions.

\citet{park2023agents} use memory, reflection, and planning to support generative agents with coherent behavior over time. This provides a broader simulation context, but its agent architecture and behavioral objectives differ from a single-turn age prompt. We do not infer that assigning an age phrase supplies an equivalent biography, memory, or life experience. \citet{dillion2023participants} discuss opportunities and caveats in using language models as substitutes for human participants. For our application, the practical requirement is construct-specific validation: agreement with pooled ratings cannot substitute for evidence that age-conditioned differences are represented correctly.

Evidence of demographic mismatch sharpens that requirement. \citet{santurkar2023opinions} evaluate alignment with demographic opinion distributions and find that steering does not necessarily remove discrepancies. \citet{bisbee2024synthetic} show that synthetic survey responses can misrepresent relationships between respondent characteristics and outcomes even when coarse averages appear plausible; they also examine sensitivity to prompting and model changes. These observations motivate our distinction between baseline agreement and demographic-gradient agreement, and our documentation of prompt text, endpoint identity, and execution conditions.

\citet{lutz2025persona} systematically vary sociodemographic persona prompting strategies and show that prompt construction affects simulated portrayals. Consequently, the effect of our age instruction could depend on its wording. We therefore test seven alternative formats, including two paraphrases, chain of thought, a numeric age, a system-prompt placement, and the interview-style format that \citet{lutz2025persona} find reduces stereotyping ($\S$\ref{app:rb-prompts}). Six formats reproduce the valence findings; the interview-style format removes most of the persona response but does not raise fidelity. \citet{nowicki2026visual} report that rater-conditioned prompting produces small, inconsistent changes in VLM predictions of image affect; we find instead that age prompts produce large shifts that are consistent across models but placed on the wrong images.

\subsection{Emotion hierarchies, functional concepts, and the closest comparisons}

\label{app:rw-mechanisms}
\citet{okawa2026emergence} examine hierarchical emotion organization and demographic variation in recognition using language-model outputs. Their persona experiment and accompanying human study make this the closest comparison for age-conditioned emotion attribution. Our evaluation changes the input modality and human target: images receive continuous affective scores, and demographic fidelity is measured against partially pooled age slopes on those same images. We do not infer their hierarchy from our small categorical vocabulary, nor claim that their study lacked a human comparison.

\citet{sofroniew2026emotion} identify internal emotion-related representations and study their behavioral influence through interventions. This is evidence about functional internal concepts in the examined model, not a validation of age-conditioned VLM responses. Our contribution is deliberately behavioral: do numerical judgments change in accordance with a measured human age association? Output-level correspondence, activation-level structure, and subjective emotional experience require different evidence. The present design addresses the first.

\subsection{Synthesis and positioning of the present study}

\label{app:rw-synthesis}
The reviewed literature suggests three distinct validation requirements. First, numerical ratings must be calibrated to a human reference rather than merely rank images correctly. Second, demographic conditioning must be assessed against human variation rather than the presence of a prompt effect. Third, psychologically motivated geometry must be defined at a level supported by the available measurements. Our design combines these requirements in an age-conditioned image-rating protocol, with separate scores for baseline agreement, image-level age alignment, and affective geometry. This is a scoped contribution; it neither establishes a comprehensive model of emotional intelligence nor identifies the mechanism producing a model's response.

\FloatBarrierAbove
\section{Data and Methods}

\label{app:methods}

We describe the human reference, the model-rating protocol, and the statistical quantities used to compare them. We then report response coverage and the robustness check on the common image subset. Table~\ref{tab:notation} summarizes the notation.

\begin{table}[tbp]
\centering
\small
\caption{\textbf{Notation used in the paper.}}
\label{tab:notation}
\begin{tabularx}{\linewidth}{lX}
\toprule
\rowcolor{titlebar!90}
Symbol or term & Meaning \\
\midrule
$g\in\{\varnothing,Y,M,O\}$ & No-age, young (18--29), middle (30--59), and older (60+) conditions \\
$\Delta_{O-Y}$ & Mean older-minus-young rating difference across images (uniform persona shift) \\
$\Delta H_i$ & Human age effect for image $i$: posterior mean change per $+1$\,SD of rater age, centred within list (Eq.~\ref{eq:human}) \\
$\Delta M_{im}$ & Model $m$'s age effect for image $i$ in the same units, centred within list (Eq.~\ref{eq:modelshift}) \\
$r_{\text{part}}$ & Correlation between $\Delta M$ and $\Delta H$ after regressing both on a quadratic in pooled human valence \\
$r_{\max}$, ceiling & Highest correlation attainable given the uncertainty of $\Delta H$: $0.86$ for the raw valence effect, $0.595$ for the pleasantness-adjusted effect used by $r_{\text{part}}$ \\
Flagged & The 167 OASIS images marked as sensitive (nudity, violence, injury) \\
Pleasant flagged & The 54 flagged images with pooled human valence $\ge4$ (17 in the 300-image subset) \\
S0--S5, VERA-Cal & Mitigation arms (Section~\ref{sec:improve}; Table~\ref{tab:arms}) \\
Development / held-out & The 300-image mitigation subset / the remaining 600 OASIS images \\
\bottomrule
\end{tabularx}
\end{table}

\subsection{Human Reference}

\label{app:human}

\subsubsection{Images, participants, and missing ratings}

\begin{table}[tbp]
\centering\small
\caption{OASIS participant-level archive used as the human reference.}
\label{tab:data}
\begin{tabular}{lr}
\toprule Quantity & Value\\\midrule
\rowcolor{titlebar}
Images & 900\\
\rowcolor{titlebar}
Person / Scene / Object / Animal & 346 / 220 / 200 / 134\\
\rowcolor{titlebar}
Sensitive-flagged / unflagged images & 167 / 733\\
\rowcolor{titlebar}
Observed ratings & 184,939\\
\rowcolor{titlebar}
Participants, total & 822\\
\rowcolor{titlebar}
Valence / arousal participants & 413 / 409\\
\rowcolor{titlebar}
Participants with age, valence / arousal & 413 / 408\\
\rowcolor{titlebar}
Age, mean (SD), valence / arousal & 37.23 (12.21) / 36.02 (11.59)\\
\rowcolor{titlebar}
Participants aged at least 60, valence / arousal & 25 / 16\\
\rowcolor{titlebar}
Age range & 18--74\\
\rowcolor{titlebar}
Images per assigned list & 225\\
\bottomrule
\end{tabular}
\end{table}

The older-adult counts across the four lists are $9,5,6,5$ for valence and $2,3,4,7$ for arousal. The archive holds 11 fewer ratings than a fully observed assignment ($822\times225=184{,}950$). The one arousal participant without a recorded age is excluded from age-dependent estimation only. Because the dimension-by-list assignment is fixed by design, unobserved participant--image pairs are missing by construction.

\subsubsection{Hierarchical model specification}

We use the following Gaussian specification:
\begin{equation}
 u_p^{(d)}\sim\mathcal N(0,\sigma_{u,d}^2),\qquad
 \begin{pmatrix}v_i^{(d)}\\s_i^{(d)}\end{pmatrix}\sim
 \mathcal N\!\left(\begin{pmatrix}0\\0\end{pmatrix},\Sigma_d\right),\qquad
 \varepsilon_{pi}^{(d)}\sim\mathcal N(0,\sigma_{\varepsilon,d}^2).
\end{equation}
Random effects are independent across participants and across images, with intercept--slope dependence within an image represented by $\Sigma_d$. Participant effects are nested in their assigned list and dimension, whereas image and participant effects are crossed within a list. List-specific intercepts $\alpha_\ell$ and slopes $\beta_\ell$ absorb differences between the four rater cohorts, and the image effects are centred within list, so $\Delta H_i=\E[s_i\mid\mathcal D_H]$ is the image-specific deviation from the list age slope. Age is standardized within dimension, because the two dimensions have different age SDs. The model treats the seven-point response as approximately interval-scaled and imposes a linear age effect; it targets an interpretable average association and is not intended for extrapolation beyond the observed ages.

\subsubsection{Variational fitting}

Let $b_d$ collect random effects and $\theta_d$ the fixed effects and variance parameters. Variational EM alternates maximization of
\begin{equation}
 \mathcal L(q,\theta_d)=\E_q[\log p(y_d,b_d\mid\theta_d)]-\E_q[\log q(b_d)]
\end{equation}
over the variational distribution $q$ and over $\theta_d$. The posterior SDs $\tau_i$ are conditional on the estimated hyperparameters.

\subsubsection{Supported age effects and the reliability ceiling}

Per-image support flags come from a permutation test on participant age labels: $691/900$ images show a supported age effect on valence and $59/900$ on arousal, and the dispersion of $\Delta H$ is correspondingly smaller for arousal (SD $0.056$ vs.\ $0.106$ scale points). Because $\Delta H$ is itself an estimate, even a model that knew the true image slopes could not correlate with it perfectly. The attenuation ceiling
\begin{equation}
r_{\max}^{(d)}=\sqrt{1-\overline{\tau^{2}}^{(d)}\big/\Var\bigl(\Delta H^{(d)}\bigr)}
\end{equation}
is $0.86$ for valence and $0.70$ for arousal. These values apply to the raw age effect. The statistic $r_{\text{part}}$ targets the age effect after the quadratic in pooled valence is partialled out, whose ceiling is $0.595$ ($\S$\ref{app:ceiling}); an independent split-half estimate (two disjoint rater halves, Spearman--Brown correction; \citealp{spearman1910correlation,brown1910some}) gives reliability $0.39$ $[0.26,0.50]$ and ceiling $0.62$ (Table~\ref{tab:rb_estimators}). We report observed correlations against these ceilings rather than against $1$.

\subsection{Elicitation Protocol and Models}

\label{app:protocol}

\subsubsection{Prompt}

\label{app:prompts}
We send each image with a single user message and no system message, using the same text for all 17 models (Figure~\ref{fig:prompts-main}). The highlighted passages are the only difference between conditions: the age sentence reads ``18 to 29 years old'', ``30 to 59 years old'' or ``60 years old or older'' for the three personas, and both passages are deleted for the baseline.


\definecolor{prompthdr}{RGB}{38,48,64}
\newcommand{\promptpanel}[3]{%
  {\setlength{\fboxsep}{2.5pt}\colorbox{prompthdr}{\parbox{\dimexpr\linewidth-2\fboxsep\relax}{%
    \centering\scriptsize\bfseries\color{white}#3}}}\par\nointerlineskip
  \vspace{-0.6pt}\includegraphics[page=#1,trim=#2,clip,width=\linewidth]{figure/ICLR27_Emotion_prompt_templates.pdf}\par\nointerlineskip}

\begin{figure}[p]
  \centering
  \promptpanel{1}{35 117.0 35 154.5}{(a)~No-age baseline}
  \promptpanel{2}{35 99.0 35 137.0}{(b)~Young adult (18--29)}
  \promptpanel{3}{35 99.0 35 137.0}{(c)~Middle adult (30--59)}
  \promptpanel{4}{35 99.0 35 137.0}{(d)~Older adult (60+)}
  \caption{\textbf{Prompt templates for the main study.} (a)~No-age baseline; (b--d)~young (18--29), middle (30--59) and older (60+) adult personas. Each request contains the image and this single user message, with no system message. Highlighted text is the only part that differs from the baseline; the three persona templates differ from one another only in the age range.}
  \label{fig:prompts-main}
\end{figure}

\subsubsection{Field order and label vocabulary}

Fields are generated in the order valence, arousal, description, emotion. The two ratings are therefore produced before any generated text, whereas the label is conditioned on the model's own description; label shifts are properties of this joint elicitation and are not treated as independent confirmation of the rating shift. The vocabulary comprises the basic-level categories of \citet{shaver1987emotion}, disgust \citep{ekman1992argument}, and neutral; without neutral, a model with no emotional response to an image would be forced into a spurious category.

\subsubsection{Rating resolution}

\label{app:lattice}
With reference ages $(23.5,44.5,67.0)$ the trend in Eq.~\ref{eq:weights} weights the three ratings by $(-43,-1,44)/1893$ times $\sigma_a$ (Appendix~\ref{app:estimands}). If ratings are integers, the smallest non-zero $|\Delta M|$ is therefore about $0.30$ scale points per SD of age, which exceeds the largest human $|\Delta H|$ on valence ($0.26$; mean $0.09$). In pilot runs integer replies made the comparison vacuous; permitting one decimal place raised decimal usage to 94\% and removed the artifact.

\subsubsection{Models, decoding and execution}

\label{app:models}

\begin{table}[tbp]
\centering\small
\caption{Evaluated models. The Qwen3-VL group (8B, 32B, 235B-A22B) is the controlled scale ladder: same developer, generation and instruction-tuning recipe. The 8B Thinking variant isolates inference-time reasoning at fixed parameter count.}
\label{tab:models}
\begin{tabular}{llr}
\toprule
\rowcolor{titlebar}
Developer & Model & Nominal scale \\
\midrule
Alibaba   & Qwen3-VL-8B-Instruct        & 8B \\
Alibaba   & Qwen3-VL-8B-Thinking        & 8B \\
Alibaba   & Qwen3-VL-32B-Instruct       & 32B \\
Alibaba   & Qwen3-VL-235B-A22B-Instruct & 235B (22B active) \\
Alibaba   & Qwen3.8-27B                 & 27B \\
Google    & Gemma-3-27B-IT              & 27B \\
Google    & Gemma-4-31B-IT              & 31B \\
Google    & Gemini-3-Flash              & undisclosed \\
Google    & Gemini-3.8-Flash            & undisclosed \\
Meta      & Llama-4-Scout               & 109B (17B active) \\
Z-AI      & GLM-4.5V                    & 106B (12B active) \\
OpenAI    & GPT-5-mini                  & undisclosed \\
OpenAI    & GPT-5.2                     & undisclosed \\
OpenAI    & GPT-5.6-Luna-Pro            & undisclosed \\
Anthropic & Claude-Sonnet-4.6           & undisclosed \\
Anthropic & Claude-Opus-5               & undisclosed \\
xAI       & Grok-4.6                    & undisclosed \\
\bottomrule
\end{tabular}
\end{table}

We access all models through one commercial routing API. We send each model $900\times4=3{,}600$ requests with temperature 0, a single completion and a 1{,}500-token output budget, chosen after observing 513-token mean completions from a verbose model against an initial 600-token cap. Temperature 0 removes sampling noise but not serving nondeterminism, and the protocol does not estimate run-to-run variance. We excluded reasoning-mode endpoints whose hidden token stream broke JSON-only output (one returned empty completions on 24 of 80 pilot requests with a length-truncation finish reason), retaining Qwen3-VL-8B-Thinking as a deliberate comparison. We store every response with its prompt hash, model identifier, decoding parameters, raw text and status in a content-addressed cache. This lets us resume interrupted runs without re-issuing completed requests. Cross-developer differences confound parameter count with data, architecture and alignment procedure, and we do not read them as scale effects.

\subsection{Estimands and Derivations}

\label{app:estimands}

\subsubsection{Why list effects and image effects differ}

Integrating over the participant intercept in the human-rating model gives
\begin{equation}
 \E[y^{(d)}\mid i,z]=\alpha_{\ell(i)}^{(d)}+v_i^{(d)}+\bigl(\beta_{\ell(i)}^{(d)}+s_i^{(d)}\bigr)z.
\end{equation}
The target $s_i$ omits the common list slope $\beta_\ell$. Write the model's raw trend as $\Delta M_{im}^{\text{raw}}=b_{\ell(i),m}+q_{im}$, where $b_{\ell,m}$ is a list-level shift (including any uniform persona shift) and $q_{im}$ the image-specific part. Centring within list on the retained images,
\begin{equation}
 \Delta M_{im}=\Delta M_{im}^{\text{raw}}-\frac{1}{n_{\ell(i),m}}\sum_{j:\ell(j)=\ell(i)}\Delta M_{jm}^{\text{raw}}
 =q_{im}-\frac{1}{n_{\ell(i),m}}\sum_{j:\ell(j)=\ell(i)}q_{jm},
\end{equation}
so list-level shifts cancel exactly and $\Delta M$ matches the list-relative definition of $\Delta H$. Without this step the two quantities are different estimands.

\subsubsection{Partial correlation}

Let $B$ be the $n\times3$ matrix with rows $(1,\bar H_i^{(V)},(\bar H_i^{(V)})^2)$ and $R=I-B(B^\top B)^{-1}B^\top$. The headline statistic is
\begin{equation}
 r_{\text{part}}=\frac{(R\,\Delta M)^\top(R\,\Delta H)}{\|R\,\Delta M\|_2\,\|R\,\Delta H\|_2}.
\end{equation}
We interpret this statistic as ``correspondence beyond a quadratic valence control''. It does not establish independence from all stimulus properties. Neither raw nor partial correlation measures agreement in magnitude: both are invariant to positive rescaling of one vector, which is why magnitudes are reported separately as contrasts (Appendix~\ref{app:pooled}).

\subsubsection{Reference ages}

For $(a_Y,a_M,a_O)=(23.5,44.5,67.0)$, $\bar a_G=45$ and
\begin{equation}
 \Delta M_{im}^{\text{raw}}=\frac{\sigma_a}{1893}\bigl(-43M_{imY}-M_{imM}+44M_{imO}\bigr).
 \label{eq:weights}
\end{equation}
The middle persona receives almost no weight because its reference age lies at the mean of the three, so the trend is essentially a scaled older-minus-young contrast. The older band is open-ended; placing it at 67.0, the midpoint of the observed 60--74 range, keeps the model and human age units on the same participants. Because a correlation is invariant to the common scale factor, the choice of reference ages affects $r_{\text{part}}$ only through the small middle-persona weight. Appendix~\ref{app:shape} shows that the response is in any case a step at 60 rather than a gradient.

\subsection{Coverage, Failures and Robustness}

\label{app:coverage}

\subsubsection{Response coverage and failure causes}

Of 61{,}200 scheduled requests, 61{,}042 returned a valid rating (99.7\%; Table~\ref{tab:coverage}). Failures are reported by cause (Table~\ref{tab:coverage}): service faults, parse faults, explicit refusals, and silent empty completions. GLM-4.5V returned a well-formed response with an empty completion body for 96 image--condition pairs, 92 of them on sensitive-flagged images, spread near-uniformly across conditions ($26/24/24/22$): content filtering that presents as an infrastructure fault. Of the $\eFailTotal$ requests without a valid rating, $\eFailYoung$ fell under the young persona and $\eFailOlder$ under the older, and no model's failures depart from uniformity across conditions (smallest $p=\eFailMinPchi$; Table~\ref{tab:failures}). These checks do not suggest that differences in non-response across age conditions explain the persona effects.

\begin{table}[tbp]
\centering
\small
\caption{Coverage and failure taxonomy. Each model was issued $900\times4=3{,}600$ requests. Failures are separated by cause: \emph{service} (transport, timeout), \emph{parse} (non-conforming output), \emph{refusal} (an explicit decline), and \emph{silent filtering} (HTTP 200 with an empty completion body). Only the last two are findings.}
\label{tab:coverage}
\begin{tabular}{llrrrrrrrr}
\toprule
\rowcolor{titlebar}
Model & Vendor & Scale & Valid & Cov. & Svc. & Parse & Refus. & Filt. & \$ \\
\midrule
Grok-4.6 & xAI & -- & 3,600 & 100.0\% & 0 & 0 & 0 & 0 & 21.42 \\
Qwen3-VL-8B & Alibaba & 8B & 3,600 & 100.0\% & 0 & 0 & 0 & 0 & 0.90 \\
Gemini-3.8-Flash & Google & -- & 3,600 & 100.0\% & 0 & 0 & 0 & 0 & 8.74 \\
Gemma-4-31B & Google & 31B & 3,600 & 100.0\% & 0 & 0 & 0 & 0 & 0.25 \\
Qwen3-VL-32B & Alibaba & 32B & 3,600 & 100.0\% & 0 & 0 & 0 & 0 & 0.67 \\
Llama-4-Scout & Meta & 109B & 3,600 & 100.0\% & 0 & 0 & 0 & 0 & 0.50 \\
Gemma-3-27B & Google & 27B & 3,599 & 100.0\% & 0 & 1 & 0 & 0 & 0.24 \\
Qwen3-VL-235B & Alibaba & 235B & 3,599 & 100.0\% & 0 & 1 & 0 & 0 & 2.05 \\
GPT-5.2 & OpenAI & -- & 3,597 & 99.9\% & 3 & 0 & 0 & 0 & 8.15 \\
Qwen3.8-27B & Alibaba & 27B & 3,597 & 99.9\% & 0 & 0 & 0 & 3 & 3.78 \\
Claude-Sonnet-4.6 & Anthropic & -- & 3,594 & 99.8\% & 4 & 0 & 2 & 0 & 9.34 \\
Claude-Opus-5 & Anthropic & -- & 3,594 & 99.8\% & 4 & 1 & 1 & 0 & 18.61 \\
Gemini-3-Flash & Google & -- & 3,593 & 99.8\% & 1 & 6 & 0 & 0 & 3.16 \\
Qwen3-VL-8B-Think & Alibaba & 8B & 3,592 & 99.8\% & 1 & 4 & 0 & 3 & 8.18 \\
GPT-5.6-Luna-Pro & OpenAI & -- & 3,590 & 99.7\% & 9 & 1 & 0 & 0 & 4.40 \\
GPT-5-mini & OpenAI & -- & 3,583 & 99.5\% & 6 & 6 & 0 & 5 & 2.62 \\
GLM-4.5V & Z-AI & 106B & 3,504 & 97.3\% & 0 & 0 & 0 & 96 & 4.28 \\
\textbf{Total} &  &  & \textbf{61,042} & \textbf{99.7\%} & \textbf{28} & \textbf{20} & \textbf{3} & \textbf{107} & \textbf{97.30} \\
\bottomrule
\end{tabular}
\\[2pt]
\begin{minipage}{\linewidth}\footnotesize Silent filtering is concentrated in GLM-4.5V, where 92 of 96 affected pairs are sensitive-flagged images and the loss is near-uniform across conditions (26/24/24/22), so it reduces coverage without biasing the young--older contrast.\end{minipage}
\end{table}

\begin{table}[tbp]
\centering
\small
\caption{Non-response by condition. Counts of requests that returned no valid rating (any failure class), split by prompt condition, with the number falling on sensitive-flagged images. $p_{\chi^2}$ tests uniformity over the four conditions (models with at least eight failures); $p_{O\,\text{vs}\,Y}$ is an exact binomial test of older against young among their failures. \emph{off-vocab} counts valid replies whose emotion label fell outside the eight permitted categories. No model's non-response is condition-asymmetric.}
\label{tab:failures}
\begin{tabular}{lrrrrrrrrr}
\toprule
\rowcolor{titlebar}
Model & base & young & middle & older & total & sens. & $p_{\chi^2}$ & $p_{O\,\text{vs}\,Y}$ & off-vocab Y/O \\
\midrule
GLM-4.5V & 26 & 24 & 24 & 22 & 96 & 92 & 0.95 & 0.88 & 28/29 \\
GPT-5-mini & 1 & 3 & 5 & 8 & 17 & 10 & 0.10 & 0.23 & 1/0 \\
GPT-5.6-Luna-Pro & 3 & 3 & 1 & 3 & 10 & 6 & 0.75 & 1.00 & 0/0 \\
Qwen3-VL-8B-Think & 1 & 3 & 3 & 1 & 8 & 6 & 0.57 & -- & 0/0 \\
Gemini-3-Flash & 1 & 0 & 2 & 4 & 7 & 2 & -- & -- & 0/2 \\
Claude-Opus-5 & 3 & 1 & 1 & 1 & 6 & 2 & -- & -- & 0/0 \\
Claude-Sonnet-4.6 & 1 & 1 & 2 & 2 & 6 & 2 & -- & -- & 0/0 \\
GPT-5.2 & 1 & 0 & 2 & 0 & 3 & 3 & -- & -- & 0/0 \\
Qwen3.8-27B & 0 & 1 & 1 & 1 & 3 & 0 & -- & -- & 0/0 \\
Gemma-3-27B & 1 & 0 & 0 & 0 & 1 & 1 & -- & -- & 0/0 \\
Qwen3-VL-235B & 0 & 0 & 0 & 1 & 1 & 1 & -- & -- & 0/0 \\
Gemma-4-31B & 0 & 0 & 0 & 0 & 0 & 0 & -- & -- & 0/0 \\
Gemini-3.8-Flash & 0 & 0 & 0 & 0 & 0 & 0 & -- & -- & 12/11 \\
Llama-4-Scout & 0 & 0 & 0 & 0 & 0 & 0 & -- & -- & 0/1 \\
Grok-4.6 & 0 & 0 & 0 & 0 & 0 & 0 & -- & -- & 0/0 \\
Qwen3-VL-8B & 0 & 0 & 0 & 0 & 0 & 0 & -- & -- & 0/0 \\
Qwen3-VL-32B & 0 & 0 & 0 & 0 & 0 & 0 & -- & -- & 0/0 \\
\textbf{Total} & 38 & 36 & 41 & 43 & 158 & 125 &  &  & 41/43 \\
\bottomrule
\end{tabular}
\end{table}

\subsubsection{Robustness on the common image subset}

Models answered slightly different image sets. We recompute every valence statistic on the 846 images for which all 17 models returned a valid rating in all four conditions. No conclusion changes: the largest change in any model's partial correlation is $0.032$, and the two sets of values correlate at $r=0.957$ (Table~\ref{tab:subset}).

\begin{table}[tbp]
\centering
\small
\caption{Robustness to differing coverage: the headline valence statistic computed on each model's own valid images, and on the common subset of images rated by every model under every condition. Agreement between the two columns shows that cross-model comparisons are not driven by which images each model happened to answer.}
\label{tab:subset}
\begin{tabular}{lrrrrr}
\toprule
\rowcolor{titlebar}
Model & $n$ all & $r_{\text{part}}$ & $n$ common & $r_{\text{part}}$ & $\Delta$ \\
\midrule
Claude-Opus-5 & 899 & 0.058 & 846 & 0.089 & 0.031 \\
Claude-Sonnet-4.6 & 898 & 0.086 & 846 & 0.105 & 0.019 \\
Gemini-3-Flash & 894 & 0.065 & 846 & 0.084 & 0.019 \\
Gemini-3.8-Flash & 900 & 0.059 & 846 & 0.068 & 0.009 \\
Gemma-3-27B & 900 & 0.009 & 846 & 0.032 & 0.023 \\
Gemma-4-31B & 900 & 0.042 & 846 & 0.063 & 0.020 \\
GLM-4.5V & 873 & 0.093 & 846 & 0.088 & -0.005 \\
GPT-5-mini & 884 & 0.037 & 846 & 0.060 & 0.023 \\
GPT-5.2 & 898 & -0.030 & 846 & -0.012 & 0.018 \\
GPT-5.6-Luna-Pro & 894 & 0.006 & 846 & 0.033 & 0.028 \\
Grok-4.6 & 900 & -0.001 & 846 & 0.020 & 0.021 \\
Llama-4-Scout & 900 & -0.017 & 846 & -0.023 & -0.006 \\
Qwen3-VL-235B & 899 & 0.072 & 846 & 0.096 & 0.023 \\
Qwen3-VL-32B & 900 & 0.002 & 846 & 0.016 & 0.013 \\
Qwen3-VL-8B & 900 & 0.001 & 846 & 0.003 & 0.002 \\
Qwen3-VL-8B-Think & 893 & 0.044 & 846 & 0.076 & 0.032 \\
Qwen3.8-27B & 897 & 0.024 & 846 & 0.048 & 0.024 \\
\bottomrule
\end{tabular}
\\[2pt]
\begin{minipage}{\linewidth}\footnotesize The common subset is enriched for non-sensitive images relative to the full set, so it is the conservative comparison for effects that concentrate on sensitive content.\end{minipage}
\end{table}

\addtocontents{toc}{\protect\apxpart{Part II}{Extended Results}}
\FloatBarrierAbove
\section{Numerical Ratings and Human Alignment}

\label{app:numerical-results}

We first compare no-age ratings with pooled human ratings. We then describe the size and shape of the persona shifts, compare them with human age differences, and examine the relation between valence and arousal. We keep pooled age contrasts separate from image-level alignment because they answer different questions.

\subsection{Baseline agreement and calibration}

\label{app:baseline}

\subsubsection{Agreement with pooled human ratings}

\begin{figure}[tbp]
\centering
\includegraphics[width=0.62\linewidth]{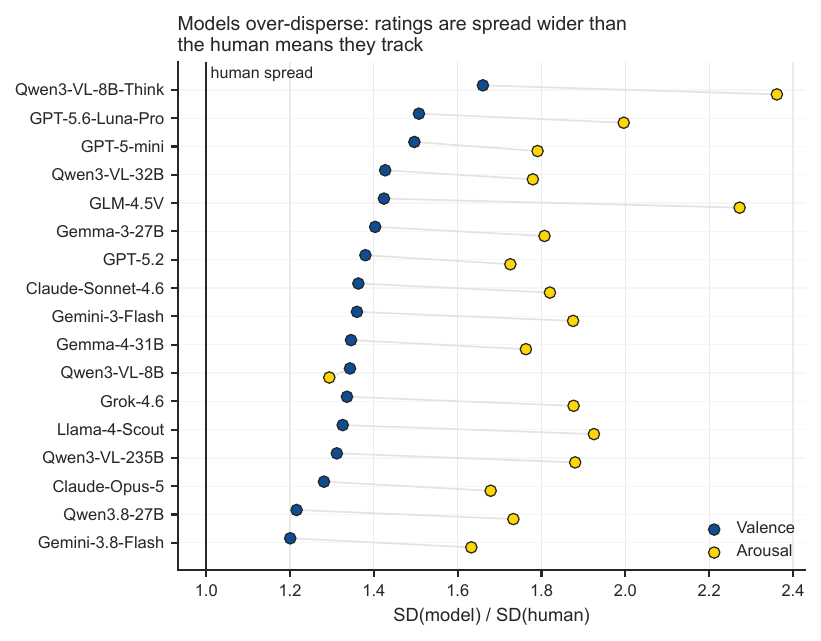}
\caption{Ratio of model rating SD to human rating SD. All 17 models on both dimensions lie to the right of parity.}
\label{fig:dispersion}
\end{figure}

Concordance is lower than correlation for every model (CCC $0.731$--$0.892$ for valence) because all models over-disperse (Figure~\ref{fig:dispersion}); mean absolute error ranges from $0.50$ to $1.03$ points.

\begin{table}[tbp]
\centering
\small
\caption{Baseline affective competence, valence: the no-age condition against pooled human image means. Correlation measures preservation of image ordering; CCC, MAE and the calibration slope measure agreement in absolute level. A slope below 1 and an SD ratio below 1 indicate compression toward the scale midpoint.}
\label{tab:baseline_valence}
\begin{tabular}{lrrrrrrrr}
\toprule
\rowcolor{titlebar}
Model & $n$ & $r$ & $\rho$ & CCC & MAE & RMSE & Slope & SD ratio \\
\midrule
Claude-Opus-5 & 897 & 0.923 & 0.932 & 0.877 & 0.574 & 0.708 & 1.182 & 1.281 \\
Gemini-3.8-Flash & 900 & 0.919 & 0.930 & 0.892 & 0.499 & 0.634 & 1.103 & 1.200 \\
Gemini-3-Flash & 899 & 0.912 & 0.924 & 0.867 & 0.599 & 0.756 & 1.241 & 1.360 \\
Gemma-4-31B & 900 & 0.908 & 0.920 & 0.867 & 0.601 & 0.750 & 1.222 & 1.345 \\
Qwen3-VL-235B & 900 & 0.902 & 0.903 & 0.827 & 0.696 & 0.864 & 1.184 & 1.312 \\
Claude-Sonnet-4.6 & 899 & 0.900 & 0.908 & 0.858 & 0.627 & 0.783 & 1.227 & 1.363 \\
GLM-4.5V & 874 & 0.895 & 0.904 & 0.841 & 0.679 & 0.845 & 1.275 & 1.424 \\
GPT-5.2 & 899 & 0.891 & 0.906 & 0.846 & 0.651 & 0.820 & 1.230 & 1.380 \\
Grok-4.6 & 900 & 0.891 & 0.908 & 0.836 & 0.680 & 0.839 & 1.190 & 1.336 \\
GPT-5.6-Luna-Pro & 897 & 0.890 & 0.906 & 0.809 & 0.807 & 0.970 & 1.342 & 1.507 \\
Qwen3-VL-8B & 900 & 0.885 & 0.887 & 0.810 & 0.739 & 0.916 & 1.189 & 1.343 \\
Llama-4-Scout & 900 & 0.878 & 0.884 & 0.839 & 0.624 & 0.819 & 1.164 & 1.326 \\
GPT-5-mini & 899 & 0.877 & 0.903 & 0.797 & 0.845 & 0.999 & 1.313 & 1.497 \\
Qwen3-VL-32B & 900 & 0.866 & 0.883 & 0.775 & 0.847 & 1.040 & 1.235 & 1.427 \\
Gemma-3-27B & 899 & 0.863 & 0.885 & 0.816 & 0.731 & 0.908 & 1.211 & 1.403 \\
Qwen3.8-27B & 900 & 0.855 & 0.865 & 0.834 & 0.589 & 0.788 & 1.039 & 1.216 \\
Qwen3-VL-8B-Think & 899 & 0.837 & 0.840 & 0.731 & 1.027 & 1.240 & 1.389 & 1.660 \\
\bottomrule
\end{tabular}
\end{table}

\begin{table}[tbp]
\centering
\small
\caption{Baseline affective competence, arousal: the no-age condition against pooled human image means. Correlation measures preservation of image ordering; CCC, MAE and the calibration slope measure agreement in absolute level. A slope below 1 and an SD ratio below 1 indicate compression toward the scale midpoint.}
\label{tab:baseline_arousal}
\begin{tabular}{lrrrrrrrr}
\toprule
\rowcolor{titlebar}
Model & $n$ & $r$ & $\rho$ & CCC & MAE & RMSE & Slope & SD ratio \\
\midrule
Claude-Sonnet-4.6 & 899 & 0.709 & 0.696 & 0.571 & 0.954 & 1.167 & 1.291 & 1.820 \\
Claude-Opus-5 & 897 & 0.706 & 0.698 & 0.550 & 0.952 & 1.162 & 1.186 & 1.679 \\
GPT-5.2 & 899 & 0.700 & 0.696 & 0.573 & 0.912 & 1.123 & 1.207 & 1.726 \\
Gemini-3.8-Flash & 900 & 0.697 & 0.688 & 0.616 & 0.798 & 0.997 & 1.137 & 1.633 \\
Qwen3-VL-32B & 900 & 0.679 & 0.688 & 0.518 & 1.006 & 1.256 & 1.208 & 1.779 \\
GPT-5.6-Luna-Pro & 897 & 0.677 & 0.665 & 0.477 & 1.190 & 1.438 & 1.351 & 1.996 \\
Gemini-3-Flash & 899 & 0.676 & 0.670 & 0.557 & 0.992 & 1.191 & 1.268 & 1.875 \\
GLM-4.5V & 874 & 0.668 & 0.680 & 0.478 & 1.252 & 1.488 & 1.519 & 2.273 \\
GPT-5-mini & 899 & 0.667 & 0.663 & 0.512 & 1.040 & 1.262 & 1.194 & 1.790 \\
Gemma-3-27B & 899 & 0.666 & 0.676 & 0.510 & 1.028 & 1.272 & 1.204 & 1.807 \\
Llama-4-Scout & 900 & 0.665 & 0.638 & 0.543 & 1.007 & 1.229 & 1.280 & 1.925 \\
Grok-4.6 & 900 & 0.661 & 0.658 & 0.516 & 1.046 & 1.276 & 1.240 & 1.876 \\
Qwen3-VL-235B & 900 & 0.638 & 0.635 & 0.482 & 1.089 & 1.345 & 1.200 & 1.880 \\
Gemma-4-31B & 900 & 0.638 & 0.645 & 0.547 & 0.929 & 1.143 & 1.125 & 1.763 \\
Qwen3.8-27B & 899 & 0.598 & 0.609 & 0.516 & 0.964 & 1.169 & 1.036 & 1.733 \\
Qwen3-VL-8B & 900 & 0.591 & 0.596 & 0.405 & 1.066 & 1.255 & 0.764 & 1.294 \\
Qwen3-VL-8B-Think & 899 & 0.526 & 0.525 & 0.362 & 1.494 & 1.750 & 1.242 & 2.361 \\
\bottomrule
\end{tabular}
\end{table}

\subsubsection{Agreement within content subsets}

Competence is not lower on the flagged subset: baseline valence agreement is $r=\eCompRsensMean$ within flagged images against $\eCompRnonsensMean$ within unflagged ones (Table~\ref{tab:competence_split}), so the persona shift on flagged images occurs despite strong baseline agreement on those images.

\begin{table}[tbp]
\centering
\small
\caption{Baseline valence agreement with human image means within content subsets: Pearson $r$ on the 167 sensitive-flagged and 733 unflagged images and within each category, and the mean signed error (model minus human, rating points) on the two flag subsets. Correlations within a subset are lower than the pooled $r$ because the range of human valence is narrower, not because the models are less competent there; the flagged subset is rated more negative than humans rate it by every model.}
\label{tab:competence_split}
\begin{tabular}{lrrrrrrrr}
\toprule
\rowcolor{titlebar}
Model & sens. & non-sens. & Person & Scene & Object & Animal & bias sens. & bias non-sens. \\
\midrule
Claude-Opus-5 & 0.94 & 0.91 & 0.92 & 0.93 & 0.93 & 0.93 & +0.17 & +0.32 \\
Gemini-3.8-Flash & 0.93 & 0.91 & 0.92 & 0.91 & 0.94 & 0.93 & +0.17 & +0.23 \\
Gemini-3-Flash & 0.92 & 0.90 & 0.91 & 0.92 & 0.94 & 0.92 & -0.03 & +0.18 \\
Gemma-4-31B & 0.91 & 0.89 & 0.90 & 0.91 & 0.93 & 0.92 & -0.13 & +0.15 \\
Qwen3-VL-235B & 0.90 & 0.89 & 0.90 & 0.91 & 0.92 & 0.89 & +0.34 & +0.49 \\
Claude-Sonnet-4.6 & 0.92 & 0.88 & 0.90 & 0.91 & 0.90 & 0.91 & -0.01 & +0.08 \\
GLM-4.5V & 0.90 & 0.88 & 0.88 & 0.91 & 0.91 & 0.91 & -0.16 & +0.12 \\
GPT-5.2 & 0.87 & 0.88 & 0.88 & 0.90 & 0.93 & 0.91 & -0.07 & +0.11 \\
Grok-4.6 & 0.90 & 0.88 & 0.89 & 0.89 & 0.91 & 0.89 & +0.22 & +0.33 \\
GPT-5.6-Luna-Pro & 0.88 & 0.88 & 0.88 & 0.90 & 0.90 & 0.90 & -0.00 & +0.32 \\
Qwen3-VL-8B & 0.87 & 0.87 & 0.88 & 0.91 & 0.91 & 0.87 & +0.33 & +0.47 \\
Llama-4-Scout & 0.87 & 0.86 & 0.86 & 0.91 & 0.89 & 0.88 & -0.02 & +0.20 \\
GPT-5-mini & 0.88 & 0.86 & 0.87 & 0.89 & 0.90 & 0.88 & +0.07 & +0.33 \\
Qwen3-VL-32B & 0.89 & 0.85 & 0.87 & 0.88 & 0.87 & 0.88 & +0.46 & +0.48 \\
Gemma-3-27B & 0.86 & 0.84 & 0.86 & 0.86 & 0.89 & 0.88 & -0.25 & +0.03 \\
Qwen3.8-27B & 0.81 & 0.85 & 0.83 & 0.88 & 0.91 & 0.85 & -0.07 & -0.16 \\
Qwen3-VL-8B-Think & 0.81 & 0.83 & 0.82 & 0.87 & 0.87 & 0.82 & +0.02 & +0.31 \\
\textbf{Mean} & \textbf{0.89} & \textbf{0.87} & \textbf{0.88} & \textbf{0.90} & \textbf{0.91} & \textbf{0.89} & \textbf{+0.06} & \textbf{+0.23} \\
\bottomrule
\end{tabular}
\end{table}

\subsection{Changes under the age personas}

\label{app:shift}

\subsubsection{Mean ratings and older-minus-young contrasts}

\begin{figure}[tbp]
\centering
\includegraphics[width=0.62\linewidth]{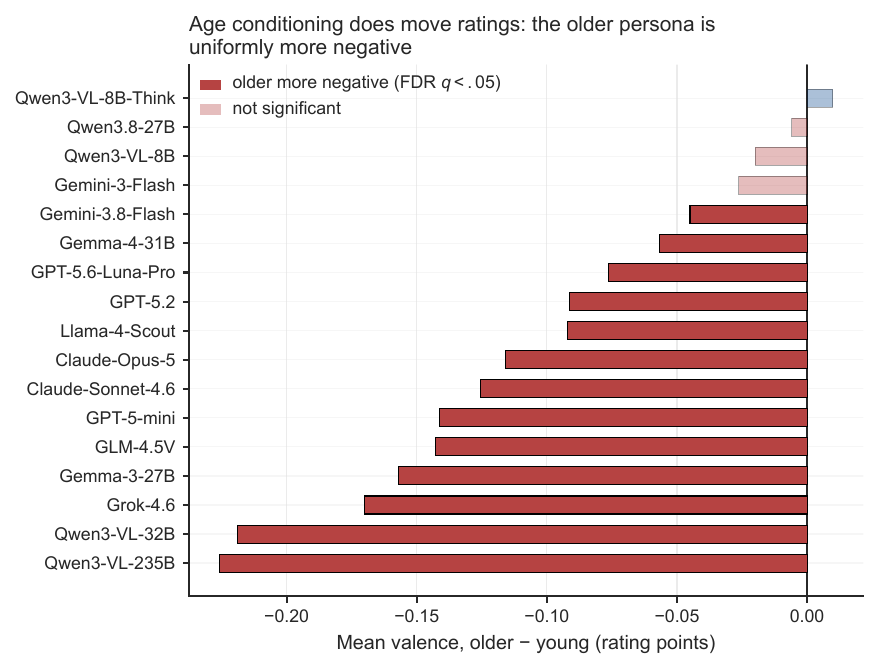}
\caption{Mean valence under the older persona minus the young persona, paired on images rated in both conditions. Solid bars survive Benjamini--Hochberg correction.}
\label{fig:shift}
\end{figure}

We report mean ratings under all four conditions in Tables~\ref{tab:conditions_valence} and~\ref{tab:conditions_arousal}. Figure~\ref{fig:shift} shows the paired older-minus-young valence contrast for each model.

\begin{table}[tbp]
\centering
\small
\caption{Mean valence by age condition, and the paired older-minus-young contrast on images rated in both. \emph{Moved} is the share of images whose rating changed at all between the two conditions.}
\label{tab:conditions_valence}
\begin{tabular}{lrrrrrrr}
\toprule
\rowcolor{titlebar}
Model & Base & Young & Middle & Older & $\Delta_{O-Y}$ & $d_z$ & Moved \\
\midrule
Qwen3-VL-235B & 4.80 & 4.88 & 4.85 & 4.66 & -0.226 & -0.32 & 35\% \\
Qwen3-VL-32B & 4.81 & 4.77 & 4.77 & 4.55 & -0.219 & -0.32 & 34\% \\
Grok-4.6 & 4.64 & 4.76 & 4.68 & 4.59 & -0.170 & -0.24 & 77\% \\
Gemma-3-27B & 4.31 & 4.37 & 4.45 & 4.21 & -0.157 & -0.21 & 38\% \\
GLM-4.5V & 4.42 & 4.51 & 4.46 & 4.36 & -0.143 & -0.17 & 72\% \\
GPT-5-mini & 4.61 & 4.74 & 4.70 & 4.59 & -0.141 & -0.19 & 68\% \\
Claude-Sonnet-4.6 & 4.40 & 4.48 & 4.43 & 4.36 & -0.126 & -0.20 & 49\% \\
Claude-Opus-5 & 4.62 & 4.64 & 4.64 & 4.53 & -0.116 & -0.30 & 69\% \\
Llama-4-Scout & 4.49 & 4.48 & 4.48 & 4.39 & -0.092 & -0.17 & 35\% \\
GPT-5.2 & 4.41 & 4.41 & 4.41 & 4.32 & -0.091 & -0.14 & 75\% \\
GPT-5.6-Luna-Pro & 4.60 & 4.65 & 4.60 & 4.57 & -0.076 & -0.17 & 51\% \\
Gemma-4-31B & 4.43 & 4.56 & 4.57 & 4.51 & -0.057 & -0.10 & 50\% \\
Gemini-3.8-Flash & 4.55 & 4.37 & 4.35 & 4.32 & -0.045 & -0.09 & 78\% \\
Gemini-3-Flash & 4.47 & 4.58 & 4.56 & 4.55 & \ns{-0.026} & -0.04 & 60\% \\
Qwen3-VL-8B & 4.78 & 4.89 & 4.86 & 4.87 & \ns{-0.020} & -0.06 & 26\% \\
Qwen3.8-27B & 4.19 & 4.21 & 4.17 & 4.20 & \ns{-0.006} & -0.01 & 43\% \\
Qwen3-VL-8B-Think & 4.58 & 4.68 & 4.72 & 4.68 & \ns{0.010} & 0.01 & 60\% \\
\bottomrule
\end{tabular}
\\[2pt]
\begin{minipage}{\linewidth}\footnotesize Older$-$young: paired $t$-test, Benjamini--Hochberg adjusted across models within dimension; entries in \textcolor{black!45}{grey} are not significant ($q\ge.05$), all others have $q<.05$.\end{minipage}
\end{table}

\begin{table}[tbp]
\centering
\small
\caption{Mean arousal by age condition, and the paired older-minus-young contrast on images rated in both. \emph{Moved} is the share of images whose rating changed at all between the two conditions.}
\label{tab:conditions_arousal}
\begin{tabular}{lrrrrrrr}
\toprule
\rowcolor{titlebar}
Model & Base & Young & Middle & Older & $\Delta_{O-Y}$ & $d_z$ & Moved \\
\midrule
GPT-5-mini & 4.23 & 4.41 & 4.28 & 3.89 & -0.518 & -0.83 & 85\% \\
GLM-4.5V & 3.99 & 4.06 & 3.95 & 3.58 & -0.486 & -0.44 & 87\% \\
Qwen3-VL-8B-Think & 4.11 & 4.32 & 4.16 & 3.88 & -0.441 & -0.42 & 74\% \\
Gemma-4-31B & 3.73 & 3.90 & 3.73 & 3.47 & -0.429 & -0.77 & 62\% \\
Gemma-3-27B & 4.23 & 4.31 & 4.31 & 3.91 & -0.402 & -0.62 & 57\% \\
Gemini-3-Flash & 3.83 & 4.01 & 3.85 & 3.61 & -0.395 & -0.59 & 72\% \\
Grok-4.6 & 4.11 & 4.22 & 4.05 & 3.83 & -0.389 & -0.71 & 85\% \\
Qwen3-VL-8B & 4.55 & 4.50 & 4.41 & 4.21 & -0.296 & -0.83 & 51\% \\
Qwen3-VL-32B & 4.26 & 4.39 & 4.34 & 4.09 & -0.295 & -0.56 & 42\% \\
Gemini-3.8-Flash & 3.81 & 4.00 & 3.93 & 3.73 & -0.275 & -0.64 & 88\% \\
Claude-Sonnet-4.6 & 4.05 & 4.14 & 4.03 & 3.86 & -0.272 & -0.57 & 54\% \\
Qwen3-VL-235B & 4.23 & 4.30 & 4.25 & 4.06 & -0.245 & -0.48 & 55\% \\
GPT-5.2 & 4.08 & 4.17 & 4.05 & 3.94 & -0.231 & -0.52 & 84\% \\
Claude-Opus-5 & 4.25 & 4.24 & 4.19 & 4.03 & -0.208 & -0.68 & 76\% \\
GPT-5.6-Luna-Pro & 4.35 & 4.39 & 4.35 & 4.20 & -0.193 & -0.42 & 63\% \\
Llama-4-Scout & 3.75 & 3.78 & 3.71 & 3.61 & -0.164 & -0.26 & 41\% \\
Qwen3.8-27B & 3.56 & 3.56 & 3.50 & 3.41 & -0.157 & -0.29 & 52\% \\
\bottomrule
\end{tabular}
\\[2pt]
\begin{minipage}{\linewidth}\footnotesize Older$-$young: paired $t$-test, Benjamini--Hochberg adjusted across models within dimension; entries in \textcolor{black!45}{grey} are not significant ($q\ge.05$), all others have $q<.05$.\end{minipage}
\end{table}

\subsubsection{Frequency and size of rating changes}

Between 26\% and 78\% of images change their valence rating at all between the young and older personas, so the shift is a broad re-rating rather than a few large moves; on average $\eMoveZeroMeanValence\%$ of an image's ratings are identical under the two personas and $\eMoveGtOneMeanValence\%$ differ by more than one point (Tables~\ref{tab:movement_valence} and~\ref{tab:movement_arousal}).

\begin{table}[tbp]
\centering
\caption{Distribution of the per-image older-minus-young change, valence. Columns 2--6 are the percentage of images whose absolute change falls in each band; \emph{neg.} is the share of moved images that moved downward; \emph{dec.} is the share of ratings using a decimal under the baseline and older conditions; SD is the across-image spread of ratings under each persona. The older persona is a re-rating of many images by small amounts, not a few large moves, and it does not compress the rating range.}
\label{tab:movement_valence}
\resizebox{\linewidth}{!}{
\begin{tabular}{lrrrrrrrrrr}
\toprule
\rowcolor{titlebar}
Model & 0 & $\le$.25 & $\le$.5 & $\le$1 & $>$1 & neg. & mean$|\Delta|$ & max & dec.\,\% $\varnothing$/O & SD Y/O \\
\midrule
Qwen3-VL-8B & 74 & 0 & 21 & 4 & 0 & 55 & 0.15 & 3.0 & 65/75 & 1.61/1.57 \\
Qwen3-VL-32B & 66 & 0 & 13 & 13 & 8 & 75 & 0.36 & 3.0 & 70/82 & 1.73/1.68 \\
Qwen3-VL-235B & 65 & 0 & 14 & 13 & 8 & 71 & 0.36 & 3.5 & 82/83 & 1.62/1.68 \\
Llama-4-Scout & 65 & 0 & 19 & 12 & 4 & 64 & 0.28 & 3.0 & 41/30 & 1.54/1.59 \\
Gemma-3-27B & 62 & 3 & 19 & 9 & 7 & 57 & 0.34 & 5.0 & 80/87 & 1.75/1.71 \\
Qwen3.8-27B & 57 & 2 & 32 & 7 & 3 & 46 & 0.27 & 3.0 & 49/52 & 1.52/1.50 \\
Claude-Sonnet-4.6 & 51 & 3 & 23 & 18 & 5 & 56 & 0.36 & 3.5 & 84/80 & 1.64/1.71 \\
Gemma-4-31B & 50 & 6 & 26 & 12 & 6 & 52 & 0.32 & 3.7 & 51/72 & 1.64/1.70 \\
GPT-5.6-Luna-Pro & 49 & 9 & 34 & 6 & 3 & 59 & 0.27 & 3.0 & 53/43 & 1.87/1.82 \\
Gemini-3-Flash & 40 & 4 & 33 & 17 & 6 & 43 & 0.35 & 3.7 & 91/93 & 1.71/1.83 \\
Qwen3-VL-8B-Think & 40 & 0 & 40 & 10 & 10 & 50 & 0.50 & 5.0 & 33/47 & 1.96/1.91 \\
GPT-5-mini & 32 & 28 & 24 & 9 & 7 & 61 & 0.38 & 4.5 & 72/66 & 1.83/1.82 \\
Claude-Opus-5 & 31 & 42 & 15 & 8 & 4 & 63 & 0.23 & 2.2 & 89/90 & 1.54/1.58 \\
GLM-4.5V & 28 & 21 & 26 & 14 & 11 & 58 & 0.49 & 4.5 & 78/71 & 1.72/1.76 \\
GPT-5.2 & 25 & 35 & 23 & 9 & 7 & 47 & 0.37 & 3.4 & 87/86 & 1.73/1.75 \\
Grok-4.6 & 23 & 32 & 24 & 13 & 9 & 58 & 0.41 & 4.0 & 83/89 & 1.61/1.69 \\
Gemini-3.8-Flash & 22 & 40 & 20 & 12 & 5 & 45 & 0.30 & 2.4 & 90/93 & 1.07/1.13 \\
\bottomrule
\end{tabular}}
\end{table}

\begin{table}[tbp]
\centering
\small
\caption{Distribution of the per-image older-minus-young change, arousal. Columns 2--6 are the percentage of images whose absolute change falls in each band; \emph{neg.} is the share of moved images that moved downward; \emph{dec.} is the share of ratings using a decimal under the baseline and older conditions; SD is the across-image spread of ratings under each persona. The older persona is a re-rating of many images by small amounts, not a few large moves, and it does not compress the rating range.}
\label{tab:movement_arousal}
\resizebox{\linewidth}{!}{
\begin{tabular}{lrrrrrrrrrr}
\toprule
\rowcolor{titlebar}
Model & 0 & $\le$.25 & $\le$.5 & $\le$1 & $>$1 & neg. & mean$|\Delta|$ & max & dec.\,\% $\varnothing$/O & SD Y/O \\
\midrule
Llama-4-Scout & 59 & 0 & 18 & 17 & 7 & 68 & 0.38 & 3.0 & 64/70 & 1.59/1.58 \\
Qwen3-VL-32B & 58 & 0 & 21 & 18 & 4 & 90 & 0.35 & 3.0 & 25/17 & 1.44/1.33 \\
Qwen3-VL-8B & 49 & 0 & 40 & 10 & 1 & 97 & 0.31 & 1.5 & 58/43 & 1.07/1.01 \\
Qwen3.8-27B & 48 & 1 & 36 & 10 & 5 & 68 & 0.36 & 2.5 & 38/38 & 1.41/1.34 \\
Claude-Sonnet-4.6 & 46 & 4 & 28 & 19 & 3 & 85 & 0.35 & 2.7 & 80/79 & 1.49/1.39 \\
Qwen3-VL-235B & 45 & 12 & 14 & 23 & 6 & 81 & 0.36 & 2.0 & 61/50 & 1.54/1.53 \\
Gemma-3-27B & 43 & 6 & 25 & 18 & 9 & 91 & 0.45 & 4.8 & 53/42 & 1.52/1.37 \\
Gemma-4-31B & 38 & 8 & 20 & 23 & 11 & 92 & 0.47 & 3.0 & 29/51 & 1.40/1.22 \\
GPT-5.6-Luna-Pro & 37 & 12 & 33 & 14 & 4 & 72 & 0.34 & 2.0 & 57/47 & 1.63/1.58 \\
Gemini-3-Flash & 28 & 4 & 27 & 29 & 12 & 82 & 0.52 & 3.5 & 96/96 & 1.52/1.56 \\
Qwen3-VL-8B-Think & 26 & 0 & 33 & 18 & 23 & 71 & 0.81 & 4.5 & 48/57 & 1.78/1.72 \\
Claude-Opus-5 & 24 & 44 & 21 & 8 & 3 & 87 & 0.25 & 1.6 & 86/85 & 1.34/1.30 \\
GPT-5.2 & 16 & 32 & 30 & 17 & 5 & 76 & 0.36 & 2.1 & 92/91 & 1.43/1.38 \\
GPT-5-mini & 15 & 16 & 27 & 25 & 16 & 88 & 0.59 & 3.5 & 74/66 & 1.45/1.36 \\
Grok-4.6 & 15 & 22 & 29 & 21 & 12 & 82 & 0.48 & 2.7 & 79/86 & 1.56/1.50 \\
GLM-4.5V & 13 & 15 & 27 & 19 & 26 & 75 & 0.81 & 5.8 & 80/71 & 1.84/1.68 \\
Gemini-3.8-Flash & 12 & 30 & 35 & 17 & 5 & 85 & 0.38 & 2.7 & 90/95 & 0.89/0.83 \\
\bottomrule
\end{tabular}}
\end{table}

\subsubsection{Changes by sensitive flag and image category}

By category, Person images carry the largest valence shift ($\eCatPersonValence$, negative for all 17 models) and Object and Animal images none ($\eCatObjectValence$, $\eCatAnimalValence$), while the arousal shift is present in every category (Table~\ref{tab:category}). Table~\ref{tab:sensitive} reports the shift by sensitive flag.

\begin{table}[tbp]
\centering
\small
\caption{Valence age effect split by the OASIS sensitive-content flag (167 flagged, 733 unflagged images). A larger effect on the flagged subset indicates that age conditioning acts mainly where content is aversive.}
\label{tab:sensitive}
\begin{tabular}{lrrrr}
\toprule
\rowcolor{titlebar}
Model & $\Delta_{O-Y}$ sens. & $\Delta_{O-Y}$ non-sens. & $\overline{\Delta M}$ sens. & $\overline{\Delta M}$ non-sens. \\
\midrule
Qwen3.8-27B & -0.117 & 0.020 & -0.033 & 0.006 \\
Qwen3-VL-8B & -0.129 & 0.005 & -0.036 & 0.001 \\
GPT-5.6-Luna-Pro & -0.183 & -0.052 & -0.052 & -0.015 \\
Gemini-3.8-Flash & -0.250 & 0.002 & -0.071 & 0.001 \\
Llama-4-Scout & -0.284 & -0.048 & -0.080 & -0.014 \\
Qwen3-VL-8B-Think & -0.320 & 0.084 & -0.084 & 0.024 \\
Claude-Opus-5 & -0.395 & -0.052 & -0.112 & -0.015 \\
Gemini-3-Flash & -0.405 & 0.060 & -0.115 & 0.017 \\
GLM-4.5V & -0.421 & -0.089 & -0.119 & -0.025 \\
Gemma-4-31B & -0.421 & 0.026 & -0.119 & 0.007 \\
GPT-5-mini & -0.446 & -0.074 & -0.113 & -0.021 \\
GPT-5.2 & -0.469 & -0.005 & -0.133 & -0.002 \\
Gemma-3-27B & -0.476 & -0.084 & -0.135 & -0.025 \\
Claude-Sonnet-4.6 & -0.486 & -0.044 & -0.137 & -0.012 \\
Grok-4.6 & -0.565 & -0.080 & -0.159 & -0.022 \\
Qwen3-VL-32B & -0.626 & -0.126 & -0.177 & -0.036 \\
Qwen3-VL-235B & -0.632 & -0.134 & -0.178 & -0.038 \\
\bottomrule
\end{tabular}
\end{table}

\begin{table}[tbp]
\centering
\small
\caption{Older-minus-young persona shift by OASIS content category (Person 346, Scene 220, Object 200, Animal 134 images). Significance: one-sample $t$-test, Benjamini--Hochberg over the 68 model$\times$category tests within dimension; model entries in \textcolor{black!45}{grey} are not significant ($q\ge.05$), all others have $q<.05$. The human row is the pooled older-minus-young rater contrast of Table~\ref{tab:human_vs_model} restricted to the category. Person images, which hold most of the sensitive-flagged content, carry the largest valence shift; the arousal shift is present in every category.}
\label{tab:category}

\resizebox{\linewidth}{!}{
\begin{tabular}{lrrrrrrrr}
\toprule
\rowcolor{titlebar}
 & \multicolumn{4}{c}{Valence} & \multicolumn{4}{c}{Arousal} \\
\cmidrule(lr){2-5}\cmidrule(lr){6-9}
\rowcolor{titlebar}
Model & Person & Scene & Object & Animal & Person & Scene & Object & Animal \\
\midrule
Qwen3-VL-32B & -0.49 & -0.13 & \ns{+0.04} & \ns{-0.06} & -0.35 & -0.29 & -0.20 & -0.30 \\
Qwen3-VL-235B & -0.38 & -0.26 & \ns{-0.05} & \ns{-0.03} & -0.23 & -0.31 & -0.27 & -0.15 \\
Gemma-3-27B & -0.37 & \ns{-0.05} & \ns{-0.01} & \ns{+0.01} & -0.54 & -0.36 & -0.30 & -0.26 \\
Grok-4.6 & -0.36 & -0.18 & +0.10 & \ns{-0.07} & -0.43 & -0.39 & -0.30 & -0.42 \\
GLM-4.5V & -0.35 & -0.17 & +0.10 & \ns{+0.05} & -0.67 & -0.39 & -0.32 & -0.47 \\
GPT-5-mini & -0.30 & -0.13 & \ns{+0.01} & \ns{+0.01} & -0.58 & -0.53 & -0.48 & -0.39 \\
Claude-Sonnet-4.6 & -0.28 & -0.18 & +0.09 & \ns{+0.04} & -0.33 & -0.29 & -0.19 & -0.23 \\
GPT-5.2 & -0.26 & -0.10 & +0.13 & \ns{+0.03} & -0.28 & -0.17 & -0.22 & -0.23 \\
Claude-Opus-5 & -0.23 & -0.12 & \ns{+0.03} & -0.05 & -0.26 & -0.19 & -0.16 & -0.17 \\
Gemma-4-31B & -0.19 & -0.10 & +0.14 & +0.07 & -0.51 & -0.45 & -0.26 & -0.44 \\
Llama-4-Scout & -0.18 & -0.10 & \ns{+0.03} & \ns{-0.03} & -0.20 & -0.09 & -0.22 & -0.11 \\
Qwen3-VL-8B-Think & -0.18 & \ns{+0.06} & +0.13 & +0.22 & -0.51 & -0.48 & -0.29 & -0.43 \\
Gemini-3.8-Flash & -0.13 & \ns{-0.06} & \ns{+0.04} & +0.07 & -0.28 & -0.30 & -0.21 & -0.31 \\
GPT-5.6-Luna-Pro & -0.12 & -0.09 & \ns{+0.00} & \ns{-0.05} & -0.23 & -0.13 & -0.21 & -0.16 \\
Gemini-3-Flash & -0.10 & \ns{-0.08} & +0.10 & +0.06 & -0.45 & -0.39 & -0.31 & -0.37 \\
Qwen3.8-27B & -0.07 & \ns{-0.02} & +0.09 & \ns{+0.03} & -0.18 & -0.15 & -0.13 & -0.15 \\
Qwen3-VL-8B & -0.06 & \ns{-0.00} & \ns{+0.02} & \ns{-0.01} & -0.33 & -0.33 & -0.24 & -0.25 \\
\midrule
\textbf{Model mean} & \textbf{-0.24} & \textbf{-0.10} & \textbf{+0.06} & \textbf{+0.02} & \textbf{-0.37} & \textbf{-0.31} & \textbf{-0.25} & \textbf{-0.28} \\
OASIS raters & +0.23 & -0.00 & +0.01 & +0.13 & +0.02 & +0.17 & -0.23 & +0.30 \\
\bottomrule
\end{tabular}}
\end{table}

\subsection{Shape of the age response}

\label{app:shape}

\begin{figure}[tbp]
  \centering
  \includegraphics[width=0.9 \linewidth]{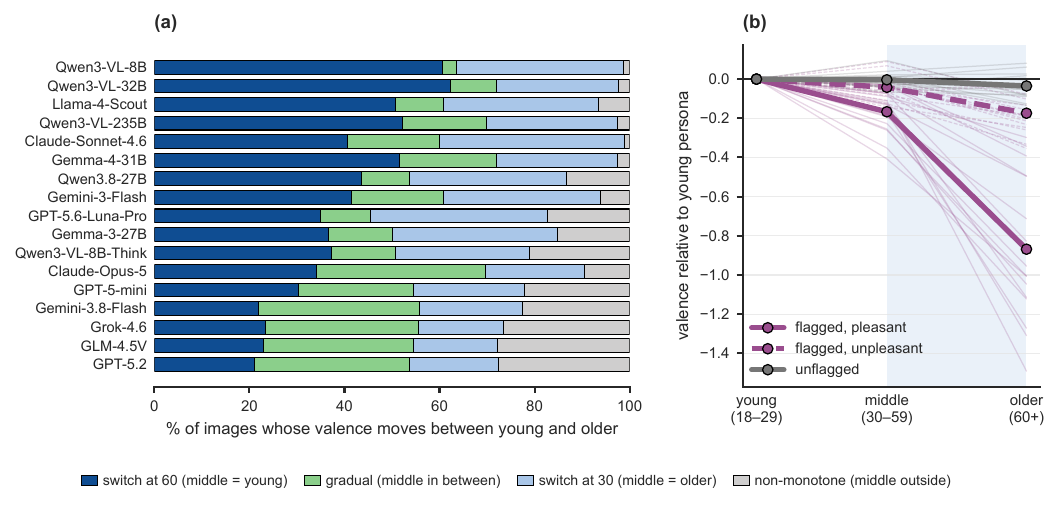}
\caption{\textbf{Age is treated as a step, not a gradient.}
(a)~Among images with at least $0.1$ valence change from young to older personas, we classify the middle rating as a switch at 30, switch at 60, gradual, or non-monotone using a $0.05$ tolerance. For the median model, 72\% of changes are step-like and only 19\% are gradual; step changes dominate in all 17 models.
(b)~Mean valence relative to the young persona across age bands. On flagged pleasant images, most of the decline occurs from middle to older age ($-0.17\rightarrow-0.87$), while unflagged images change little ($-0.04$ at 60+).}
  \label{fig:step}
\end{figure}

\subsubsection{Which age persona the baseline resembles}

\begin{figure}[tbp]
\centering
\includegraphics[width=0.92\linewidth]{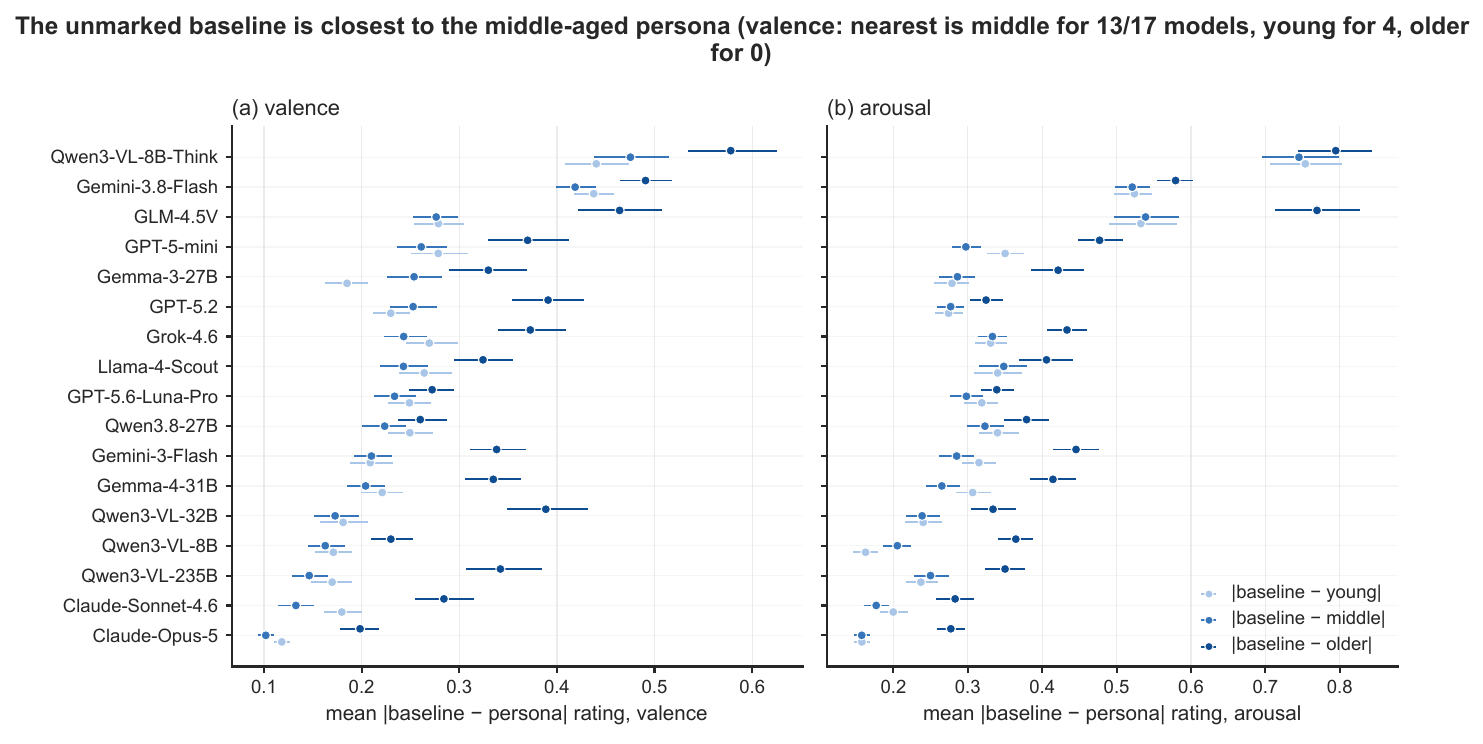}
\caption{Mean absolute difference between the baseline rating and each persona's rating on the same image, with 95\% image-bootstrap intervals.}
\label{fig:unmarked_default}
\end{figure}

The no-age prompt is closest to the middle persona for $\eDefNearMiddleValence$ of 17 models on valence (young for $\eDefNearYoungValence$, older for none) and for $\eDefNearMiddleArousal$ on arousal (Tables~\ref{tab:default_valence} and~\ref{tab:default_arousal}, Figure~\ref{fig:unmarked_default}). The mean absolute distance from the baseline is $\eDefAbsYoungValence$ to the young rating, $\eDefAbsMiddleValence$ to the middle rating and $\eDefAbsOlderValence$ to the older rating. Placing the baseline mean on the line through the three band means gives an implied age with median $\eDefImpliedMedianValence$ years across the $\eDefImpliedNValence$ models whose span allows the estimate.

\begin{table}[tbp]
\centering
\small
\caption{Which persona the no-age baseline resembles, valence. $|b-X|$ is the mean absolute difference between the baseline rating and the rating under persona $X$ on the same image; \emph{nearest} is the smallest of the three. $p_{Y\,\text{vs}\,M}$ is a Wilcoxon signed-rank test of $|b-Y|$ against $|b-M|$ over images; values of $p\ge.05$ are shown in \textcolor{black!45}{grey}. \emph{Implied age} places the baseline mean on the line through the three band means (omitted when the young--older span is under 0.05 points). Nearest band across models: young 4, middle 13, older 0.}
\label{tab:default_valence}
\resizebox{\linewidth}{!}{
\begin{tabular}{lrrrrlrrr}
\toprule
\rowcolor{titlebar}
Model & $n$ & $|b-Y|$ & $|b-M|$ & $|b-O|$ & nearest & $p_{Y\,\text{vs}\,M}$ & $M-b$ & implied age \\
\midrule
Claude-Opus-5 & 897 & 0.118 & 0.102 & 0.198 & middle & $<$0.001 & +0.016 & 39 \\
Claude-Sonnet-4.6 & 898 & 0.180 & 0.133 & 0.284 & middle & $<$0.001 & +0.035 & 55 \\
Qwen3-VL-235B & 899 & 0.170 & 0.146 & 0.342 & middle & \ns{0.414} & +0.052 & 45 \\
Qwen3-VL-8B & 900 & 0.171 & 0.163 & 0.230 & middle & \ns{0.410} & +0.087 & -- \\
Qwen3-VL-32B & 900 & 0.181 & 0.173 & 0.389 & middle & \ns{0.453} & -0.036 & 24 \\
Gemma-4-31B & 900 & 0.221 & 0.204 & 0.335 & middle & 0.019 & +0.138 & 120 \\
Gemini-3-Flash & 893 & 0.209 & 0.210 & 0.338 & young & \ns{0.195} & +0.086 & -- \\
Qwen3.8-27B & 897 & 0.249 & 0.224 & 0.260 & middle & 0.019 & -0.019 & -- \\
GPT-5.6-Luna-Pro & 894 & 0.249 & 0.234 & 0.272 & middle & \ns{0.169} & +0.019 & 53 \\
Llama-4-Scout & 900 & 0.264 & 0.243 & 0.324 & middle & \ns{0.153} & -0.010 & 26 \\
Grok-4.6 & 900 & 0.269 & 0.243 & 0.373 & middle & \ns{0.080} & +0.035 & 54 \\
GPT-5.2 & 898 & 0.230 & 0.253 & 0.391 & young & \ns{0.758} & -0.006 & 31 \\
Gemma-3-27B & 899 & 0.185 & 0.254 & 0.330 & young & 0.001 & +0.141 & 55 \\
GPT-5-mini & 884 & 0.279 & 0.261 & 0.370 & middle & \ns{0.844} & +0.086 & 65 \\
GLM-4.5V & 873 & 0.279 & 0.276 & 0.464 & middle & \ns{0.545} & +0.052 & 56 \\
Gemini-3.8-Flash & 900 & 0.438 & 0.419 & 0.491 & middle & 0.037 & -0.199 & -- \\
Qwen3-VL-8B-Think & 892 & 0.440 & 0.475 & 0.578 & young & \ns{0.254} & +0.140 & -- \\
\bottomrule
\end{tabular}}
\end{table}

\begin{table}[tbp]
\centering
\small
\caption{Which persona the no-age baseline resembles, arousal. $|b-X|$ is the mean absolute difference between the baseline rating and the rating under persona $X$ on the same image; \emph{nearest} is the smallest of the three. $p_{Y\,\text{vs}\,M}$ is a Wilcoxon signed-rank test of $|b-Y|$ against $|b-M|$ over images; values of $p\ge.05$ are shown in \textcolor{black!45}{grey}. \emph{Implied age} places the baseline mean on the line through the three band means (omitted when the young--older span is under 0.05 points). Nearest band across models: young 7, middle 10, older 0.}
\label{tab:default_arousal}
\resizebox{\linewidth}{!}{
\begin{tabular}{lrrrrlrrr}
\toprule
\rowcolor{titlebar}
Model & $n$ & $|b-Y|$ & $|b-M|$ & $|b-O|$ & nearest & $p_{Y\,\text{vs}\,M}$ & $M-b$ & implied age \\
\midrule
Claude-Opus-5 & 897 & 0.158 & 0.157 & 0.277 & middle & \ns{0.508} & -0.060 & 25 \\
Claude-Sonnet-4.6 & 898 & 0.200 & 0.177 & 0.283 & middle & \ns{0.090} & -0.015 & 39 \\
Qwen3-VL-8B & 900 & 0.163 & 0.205 & 0.365 & young & $<$0.001 & -0.137 & 20 \\
Qwen3-VL-32B & 900 & 0.240 & 0.239 & 0.334 & middle & \ns{0.943} & +0.084 & 47 \\
Qwen3-VL-235B & 899 & 0.237 & 0.250 & 0.350 & young & \ns{0.255} & +0.024 & 41 \\
Gemma-4-31B & 900 & 0.307 & 0.265 & 0.414 & middle & $<$0.001 & +0.000 & 42 \\
GPT-5.2 & 898 & 0.274 & 0.277 & 0.324 & young & \ns{0.311} & -0.020 & 41 \\
Gemini-3-Flash & 893 & 0.315 & 0.285 & 0.445 & middle & 0.019 & +0.019 & 44 \\
Gemma-3-27B & 899 & 0.279 & 0.286 & 0.421 & young & \ns{0.572} & +0.079 & 39 \\
GPT-5-mini & 884 & 0.350 & 0.297 & 0.477 & middle & $<$0.001 & +0.052 & 42 \\
GPT-5.6-Luna-Pro & 894 & 0.319 & 0.298 & 0.339 & middle & \ns{0.067} & -0.000 & 37 \\
Qwen3.8-27B & 895 & 0.340 & 0.323 & 0.379 & middle & \ns{0.196} & -0.058 & 26 \\
Grok-4.6 & 900 & 0.331 & 0.333 & 0.433 & young & \ns{0.459} & -0.065 & 36 \\
Llama-4-Scout & 900 & 0.340 & 0.348 & 0.406 & young & \ns{0.777} & -0.039 & 33 \\
Gemini-3.8-Flash & 900 & 0.524 & 0.521 & 0.579 & middle & \ns{0.747} & +0.125 & 58 \\
GLM-4.5V & 873 & 0.533 & 0.539 & 0.769 & young & \ns{0.411} & -0.044 & 34 \\
Qwen3-VL-8B-Think & 892 & 0.754 & 0.745 & 0.795 & middle & \ns{0.919} & +0.050 & 46 \\
\bottomrule
\end{tabular}}
\end{table}

\subsubsection{Position of the middle persona}

\begin{figure}[tbp]
\centering
\includegraphics[width=\linewidth]{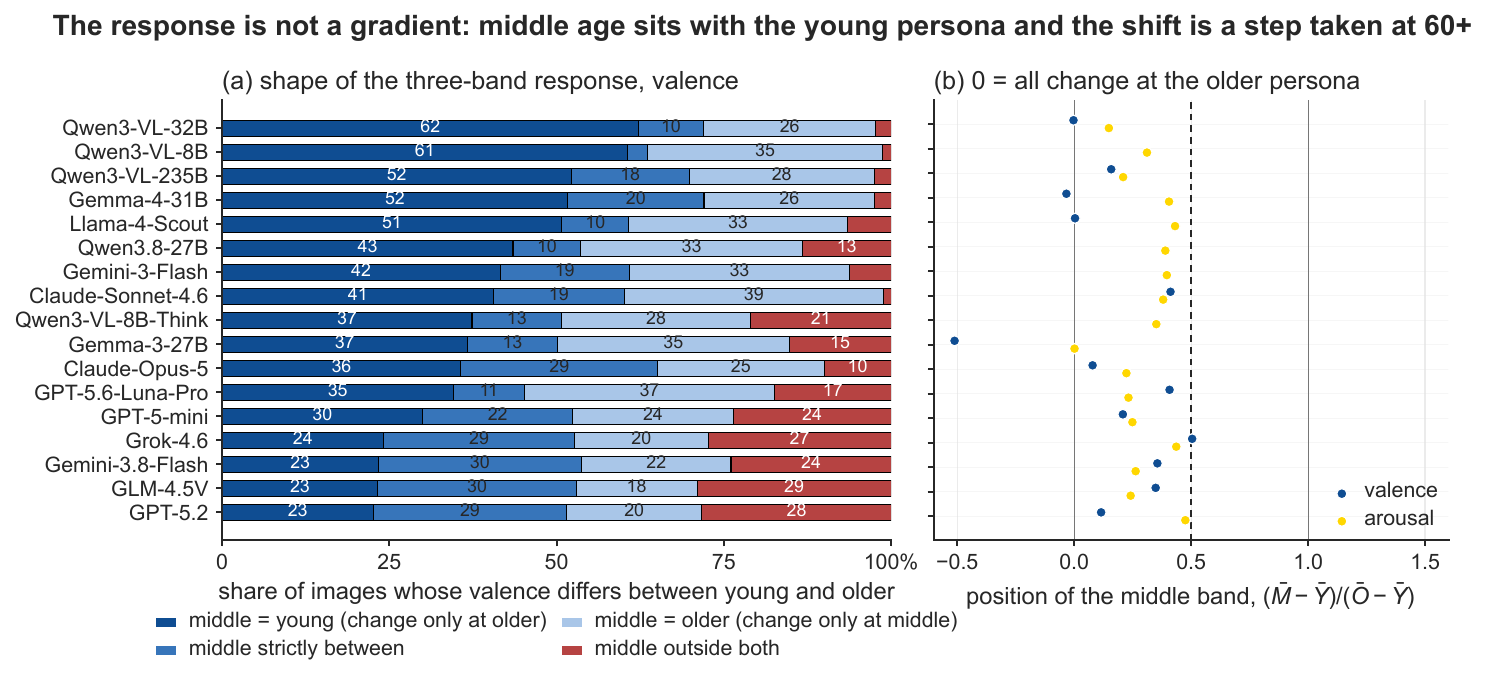}
\caption{Shape of the three-band response. (a) Among images that move between young and older, where the middle rating lies. (b) The position of the middle band between the other two at the level of means.}
\label{fig:monotonicity}
\end{figure}

Among images whose valence differs between the young and older personas, the middle rating equals the young rating for $\eMonoStepOlderMeanValence\%$ on average, lies strictly between the endpoints for $\eMonoBetweenMeanValence\%$, and lies outside them for $\eMonoOutsideMeanValence\%$ (Table~\ref{tab:monotonic_valence}; arousal in Table~\ref{tab:monotonic_arousal}; Figure~\ref{fig:monotonicity}). At the level of means, the middle band's position between young and older has a median of $\eMonoPosMedianValence$ on valence and $\eMonoPosMedianArousal$ on arousal, where $0$ places all change at the older persona. The persona effect is an old/not-old distinction taken at 60. This does not affect the alignment conclusion, which concerns the image-level pattern, but it means the trend of Eq.~\ref{eq:weights} should be read as an older-minus-young contrast.

\begin{table}[tbp]
\centering
\small
\caption{Shape of the three-band age response, valence. Among images whose rating differs between young and older (\emph{moved}), \emph{between} is the share with the middle rating strictly inside the two endpoints, step$_{M\to O}$ the share where middle equals young (all change occurs at the older persona), step$_{Y\to M}$ the share where middle equals older, and \emph{outside} the share with middle beyond both. \emph{bump} is the share of all images where young and older agree but middle differs. \emph{position} is $(\bar M-\bar Y)/(\bar O-\bar Y)$ at the mean level (0: change confined to the older persona; 1: confined to the young persona; omitted when the span is under 0.03). $r$ correlates the two successive steps over images.}
\label{tab:monotonic_valence}
\resizebox{\linewidth}{!}{
\begin{tabular}{lrrrrrrrr}
\toprule
\rowcolor{titlebar}
Model & moved & between & step$_{M\to O}$ & step$_{Y\to M}$ & outside & bump & position & $r$ \\
\midrule
Gemma-3-27B & 341 & 13\% & 37\% & 35\% & 15\% & 12\% & -0.51 & -0.22 \\
Gemma-4-31B & 447 & 20\% & 52\% & 26\% & 2\% & 5\% & -0.03 & 0.15 \\
Qwen3-VL-32B & 310 & 10\% & 62\% & 26\% & 2\% & 3\% & -0.00 & 0.04 \\
Llama-4-Scout & 319 & 10\% & 51\% & 33\% & 7\% & 10\% & 0.00 & -0.32 \\
Claude-Opus-5 & 619 & 29\% & 36\% & 25\% & 10\% & 11\% & 0.08 & 0.16 \\
GPT-5.2 & 672 & 29\% & 23\% & 20\% & 28\% & 14\% & 0.12 & -0.10 \\
Qwen3-VL-235B & 318 & 18\% & 52\% & 28\% & 3\% & 5\% & 0.16 & 0.03 \\
GPT-5-mini & 597 & 22\% & 30\% & 24\% & 24\% & 14\% & 0.21 & -0.13 \\
GLM-4.5V & 631 & 30\% & 23\% & 18\% & 29\% & 12\% & 0.35 & -0.01 \\
Gemini-3.8-Flash & 698 & 30\% & 23\% & 22\% & 24\% & 12\% & 0.36 & -0.13 \\
GPT-5.6-Luna-Pro & 454 & 11\% & 35\% & 37\% & 17\% & 13\% & 0.41 & -0.42 \\
Claude-Sonnet-4.6 & 436 & 19\% & 41\% & 39\% & 1\% & 3\% & 0.41 & 0.20 \\
Grok-4.6 & 691 & 29\% & 24\% & 20\% & 27\% & 12\% & 0.50 & -0.07 \\
Gemini-3-Flash & 534 & 19\% & 42\% & 33\% & 6\% & 7\% & -- & 0.06 \\
Qwen3-VL-8B & 231 & 3\% & 61\% & 35\% & 1\% & 6\% & -- & -0.25 \\
Qwen3-VL-8B-Think & 538 & 13\% & 37\% & 28\% & 21\% & 17\% & -- & -0.44 \\
Qwen3.8-27B & 384 & 10\% & 43\% & 33\% & 13\% & 10\% & -- & -0.33 \\
\bottomrule
\end{tabular}}
\end{table}

\begin{table}[tbp]
\centering
\small
\caption{Shape of the three-band age response, arousal. Among images whose rating differs between young and older (\emph{moved}), \emph{between} is the share with the middle rating strictly inside the two endpoints, step$_{M\to O}$ the share where middle equals young (all change occurs at the older persona), step$_{Y\to M}$ the share where middle equals older, and \emph{outside} the share with middle beyond both. \emph{bump} is the share of all images where young and older agree but middle differs. \emph{position} is $(\bar M-\bar Y)/(\bar O-\bar Y)$ at the mean level (0: change confined to the older persona; 1: confined to the young persona; omitted when the span is under 0.03). $r$ correlates the two successive steps over images.}
\label{tab:monotonic_arousal}
\resizebox{\linewidth}{!}{
\begin{tabular}{lrrrrrrrr}
\toprule
\rowcolor{titlebar}
Model & moved & between & step$_{M\to O}$ & step$_{Y\to M}$ & outside & bump & position & $r$ \\
\midrule
Gemma-3-27B & 515 & 20\% & 44\% & 22\% & 14\% & 14\% & 0.00 & -0.22 \\
Qwen3-VL-32B & 382 & 11\% & 60\% & 26\% & 3\% & 5\% & 0.15 & -0.13 \\
Qwen3-VL-235B & 498 & 12\% & 50\% & 27\% & 11\% & 9\% & 0.21 & -0.28 \\
Claude-Opus-5 & 681 & 25\% & 40\% & 21\% & 14\% & 10\% & 0.22 & -0.06 \\
GPT-5.6-Luna-Pro & 564 & 17\% & 29\% & 28\% & 26\% & 15\% & 0.23 & -0.43 \\
GLM-4.5V & 756 & 34\% & 14\% & 12\% & 40\% & 9\% & 0.24 & -0.37 \\
GPT-5-mini & 747 & 37\% & 21\% & 12\% & 30\% & 8\% & 0.25 & -0.20 \\
Gemini-3.8-Flash & 790 & 38\% & 19\% & 15\% & 28\% & 7\% & 0.26 & -0.35 \\
Qwen3-VL-8B & 462 & 6\% & 60\% & 31\% & 3\% & 5\% & 0.31 & -0.35 \\
Qwen3-VL-8B-Think & 663 & 18\% & 29\% & 22\% & 31\% & 16\% & 0.35 & -0.44 \\
Claude-Sonnet-4.6 & 484 & 21\% & 38\% & 38\% & 3\% & 4\% & 0.38 & -0.05 \\
Qwen3.8-27B & 464 & 14\% & 33\% & 37\% & 16\% & 16\% & 0.39 & -0.41 \\
Gemini-3-Flash & 641 & 28\% & 33\% & 26\% & 13\% & 7\% & 0.40 & -0.05 \\
Gemma-4-31B & 557 & 22\% & 41\% & 32\% & 5\% & 5\% & 0.41 & -0.12 \\
Llama-4-Scout & 372 & 8\% & 48\% & 34\% & 9\% & 13\% & 0.43 & -0.46 \\
Grok-4.6 & 769 & 34\% & 16\% & 13\% & 37\% & 10\% & 0.44 & -0.32 \\
GPT-5.2 & 753 & 31\% & 15\% & 21\% & 33\% & 10\% & 0.48 & -0.31 \\
\bottomrule
\end{tabular}}
\end{table}

\subsection{Pooled human and model age contrasts}

\label{app:pooled}

The image-level analysis compares list-centred quantities and therefore removes any global shift. We can estimate the global shift from OASIS because every image has at least two raters aged 60 or over on its list: the mean over images of the older-rater mean minus the young-rater mean is a pooled contrast over $\eHumNValOlder$ (valence) and $\eHumNAroOlder$ (arousal) older raters, with a participant bootstrap stratified by list. It is not a per-image estimate and we do not use it as one.

\begin{figure}[tbp]
\centering
\includegraphics[width=\linewidth]{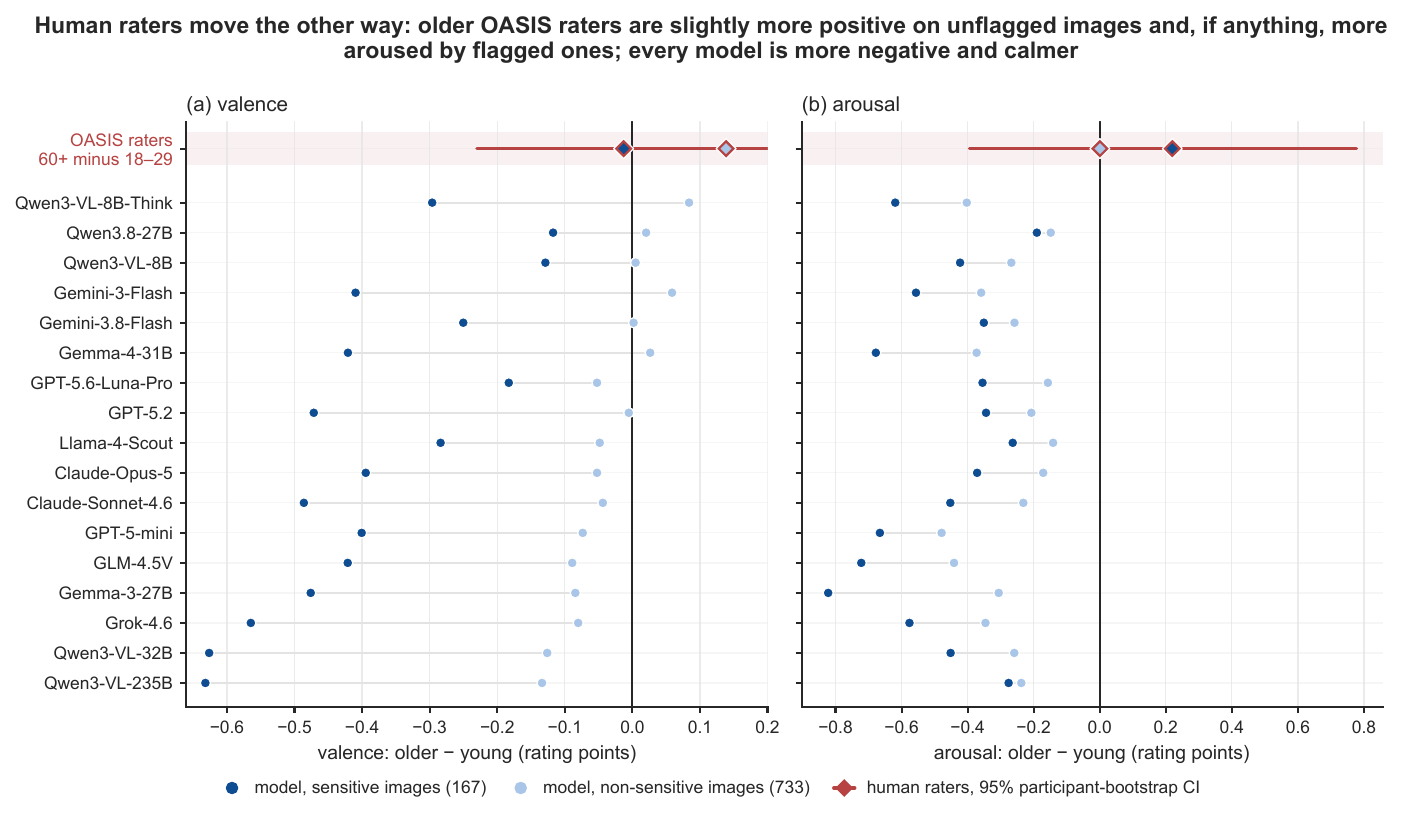}
\caption{Older-minus-young contrast in OASIS raters (diamonds, 95\% participant bootstrap) and in each model (circles), on the 167 flagged and 733 unflagged images, for valence (a) and arousal (b). The models sit below the human interval on almost every row.}
\label{fig:human_vs_model}
\end{figure}

Older OASIS raters are, if anything, slightly more positive than young raters: $\eHumValenceAll$ overall ($[\eHumValenceAllLo,\eHumValenceAllHi]$) and $\eHumValenceNonsens$ on unflagged images ($[\eHumValenceNonsensLo,\eHumValenceNonsensHi]$), with no difference on flagged images ($\eHumValenceSens$). None of these pooled intervals excludes zero, and we make no claim that the human shift is reliably positive. The claim is comparative: on unflagged valence the model contrasts run from $\eModValNonsensMin$ to $\eModValNonsensMax$, and $\eBelowValenceNonsens$ of 17 fall below the human interval; only $\eAgreeValenceNonsens$ models share the human sign there, and none does on arousal (Table~\ref{tab:human_vs_model}, Figure~\ref{fig:human_vs_model}). The three-band mean ratings show the same pattern: every model descends toward the older band while the human curve is flat or rises.

\begin{table}[tbp]
\centering
\small
\caption{Older-minus-young shift in the human raters and in every model, pooled over images. The human contrast is the mean over images of the mean rating given by raters aged 60 or over minus that given by raters aged 18--29 on the same image; the interval is a participant bootstrap stratified by list (1{,}000 replicates). Model rows are the paired older-minus-young persona contrast on the same images. The human contrast is small and its interval includes zero, but its sign is positive on unflagged images and on arousal; the models are negative on every subset, and most fall below the lower end of the human interval.}
\label{tab:human_vs_model}
\resizebox{\linewidth}{!}{
\begin{tabular}{lrrrrrr}
\toprule
\rowcolor{titlebar}
 & \multicolumn{3}{c}{Valence, older $-$ young} & \multicolumn{3}{c}{Arousal, older $-$ young} \\
\cmidrule(lr){2-4}\cmidrule(lr){5-7}
\rowcolor{titlebar}
Source & all & sens. & non-sens. & all & sens. & non-sens. \\
\midrule
\textbf{OASIS raters} & \textbf{+0.111} & \textbf{-0.013} & \textbf{+0.139} & \textbf{+0.041} & \textbf{+0.220} & \textbf{+0.000} \\
\quad 95\% CI & {\scriptsize [-0.05, 0.29]} & {\scriptsize [-0.23, 0.22]} & {\scriptsize [-0.03, 0.32]} & {\scriptsize [-0.35, 0.48]} & {\scriptsize [-0.26, 0.78]} & {\scriptsize [-0.39, 0.42]} \\
\midrule
Qwen3-VL-235B & -0.226 & -0.632 & -0.134 & -0.245 & -0.276 & -0.238 \\
Qwen3-VL-32B & -0.219 & -0.626 & -0.126 & -0.295 & -0.451 & -0.259 \\
Grok-4.6 & -0.170 & -0.565 & -0.080 & -0.389 & -0.576 & -0.346 \\
Gemma-3-27B & -0.157 & -0.476 & -0.084 & -0.402 & -0.822 & -0.306 \\
GLM-4.5V & -0.143 & -0.421 & -0.089 & -0.486 & -0.722 & -0.441 \\
GPT-5-mini & -0.132 & -0.401 & -0.073 & -0.512 & -0.666 & -0.479 \\
Claude-Sonnet-4.6 & -0.126 & -0.486 & -0.044 & -0.272 & -0.452 & -0.231 \\
Claude-Opus-5 & -0.116 & -0.395 & -0.052 & -0.208 & -0.371 & -0.171 \\
Llama-4-Scout & -0.092 & -0.284 & -0.048 & -0.164 & -0.263 & -0.141 \\
GPT-5.2 & -0.091 & -0.472 & \ns{-0.005} & -0.232 & -0.344 & -0.207 \\
GPT-5.6-Luna-Pro & -0.076 & -0.183 & -0.052 & -0.193 & -0.355 & -0.157 \\
Gemma-4-31B & -0.057 & -0.421 & \ns{+0.026} & -0.429 & -0.678 & -0.373 \\
Gemini-3.8-Flash & -0.045 & -0.250 & \ns{+0.002} & -0.275 & -0.351 & -0.258 \\
Gemini-3-Flash & \ns{-0.028} & -0.410 & +0.059 & -0.395 & -0.556 & -0.359 \\
Qwen3-VL-8B & \ns{-0.020} & -0.129 & \ns{+0.005} & -0.296 & -0.423 & -0.268 \\
Qwen3.8-27B & \ns{-0.005} & -0.117 & \ns{+0.020} & -0.156 & -0.190 & -0.148 \\
Qwen3-VL-8B-Think & \ns{+0.015} & -0.296 & +0.084 & -0.442 & -0.619 & -0.403 \\
\midrule
Models with the human sign & 1/17 & 17/17 & 6/17 & 0/17 & 0/17 & 0/17 \\
Models below the human 95\% CI & 12/17 & 14/17 & 10/17 & 7/17 & 16/17 & 3/17 \\
\bottomrule
\end{tabular}}
\\[2pt]\begin{minipage}{\linewidth}\footnotesize Human rater counts: valence 139 / 249 / 25 and arousal 140 / 252 / 16 in the young / middle / older bands; every image has at least two older raters on its list, so the pooled contrast is estimable even though per-image older means are not. Model significance: one-sample $t$-test on the paired difference, Benjamini--Hochberg within column; entries in \textcolor{black!45}{grey} are not significant ($q\ge.05$), all others have $q<.05$.\end{minipage}
\end{table}

\subsection{Image-level alignment with human age effects}

\label{app:alignment}

\subsubsection{Valence and arousal alignment}

Table~\ref{tab:alignment_valence} and Table~\ref{tab:alignment_arousal} give every alignment metric. On arousal the partial correlations are consistently negative, with 16 of 17 image-bootstrap intervals excluding zero; once raters are also resampled, none excludes zero (Table~\ref{tab:rb_rater_arousal}). Only $59/900$ images carry a supported human arousal effect. We therefore interpret the negative correlations cautiously: the arousal reference is weaker, and these results alone do not establish a reversed human age pattern.

\begin{figure}[tbp]
\centering
\includegraphics[width=\linewidth]{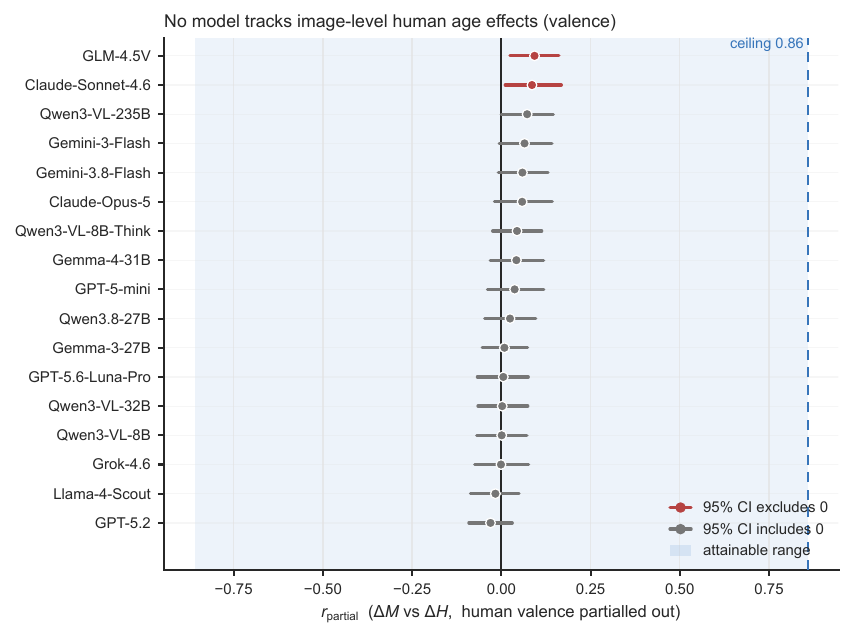}
\caption{Image-level alignment between each model's valence age effect and the human age effect, with image pleasantness partialled out. Bars are 95\% image-bootstrap intervals; the shaded band is the attainable range implied by the posterior SD of the human estimate.}
\label{fig:forest}
\end{figure}

\begin{table}[tbp]
\centering
\small
\caption{Age-shift alignment, valence: the model age effect $\Delta M$ against the human age effect $\Delta H$, both in rating points per $+1$\,SD of rater age. $r_{\text{part}}$ removes a quadratic function of pooled human valence from both vectors and is the headline quantity; $r_{\max}$ is the attenuation ceiling implied by the posterior SD of $\Delta H$ for the raw age effect; the ceiling for $r_{\text{part}}$ is $0.595$ (Table~\ref{tab:rb_ceiling_magnitude}).
\label{tab:alignment_valence}}
\begin{tabular}{lrrrcrrrr}
\toprule
\rowcolor{titlebar}
Model & $n$ & $r$ & $r_{\text{part}}$ & 95\% CI & $r_{\max}$ & Flat & Sign & SD$(\Delta M)$ \\
\midrule
GLM-4.5V & 873 & 0.030 & 0.093 & [0.02, 0.16] & 0.857 & 4\% & 0.53 & 0.240 \\
Claude-Sonnet-4.6 & 898 & 0.121 & 0.086 & [0.01, 0.17] & 0.859 & 0\% & 0.55 & 0.176 \\
Qwen3-VL-235B & 899 & 0.066 & 0.072 & [-0.00, 0.15] & 0.859 & 0\% & 0.56 & 0.195 \\
Gemini-3-Flash & 894 & 0.196 & 0.065 & [-0.01, 0.14] & 0.859 & 21\% & 0.60 & 0.166 \\
Gemini-3.8-Flash & 900 & 0.098 & 0.059 & [-0.01, 0.13] & 0.859 & 9\% & 0.55 & 0.134 \\
Claude-Opus-5 & 899 & 0.093 & 0.058 & [-0.02, 0.14] & 0.859 & 11\% & 0.54 & 0.109 \\
Qwen3-VL-8B-Think & 893 & -0.065 & 0.044 & [-0.02, 0.11] & 0.859 & 14\% & 0.45 & 0.225 \\
Gemma-4-31B & 900 & 0.044 & 0.042 & [-0.03, 0.12] & 0.859 & 16\% & 0.54 & 0.165 \\
GPT-5-mini & 884 & -0.035 & 0.037 & [-0.04, 0.12] & 0.859 & 4\% & 0.47 & 0.208 \\
Qwen3.8-27B & 897 & -0.013 & 0.024 & [-0.05, 0.10] & 0.859 & 30\% & 0.49 & 0.131 \\
Gemma-3-27B & 900 & -0.093 & 0.009 & [-0.05, 0.07] & 0.859 & 1\% & 0.45 & 0.211 \\
GPT-5.6-Luna-Pro & 894 & -0.106 & 0.006 & [-0.07, 0.08] & 0.857 & 0\% & 0.44 & 0.129 \\
Qwen3-VL-32B & 900 & -0.096 & 0.002 & [-0.07, 0.07] & 0.859 & 0\% & 0.43 & 0.193 \\
Qwen3-VL-8B & 900 & -0.071 & 0.001 & [-0.07, 0.07] & 0.859 & 73\% & 0.52 & 0.092 \\
Grok-4.6 & 900 & 0.037 & -0.001 & [-0.07, 0.08] & 0.859 & 1\% & 0.53 & 0.198 \\
Llama-4-Scout & 900 & 0.081 & -0.017 & [-0.09, 0.05] & 0.859 & 0\% & 0.52 & 0.148 \\
GPT-5.2 & 898 & -0.047 & -0.030 & [-0.09, 0.03] & 0.858 & 7\% & 0.51 & 0.188 \\
\bottomrule
\end{tabular}
\\[2pt]
\begin{minipage}{\linewidth}\footnotesize CI by resampling images, 2{,}000 replicates. \emph{Flat} is the share of images with $|\Delta M|<0.01$.\end{minipage}
\end{table}

\begin{table}[tbp]
\centering
\small
\caption{Age-shift alignment, arousal: the model age effect $\Delta M$ against the human age effect $\Delta H$, both in rating points per $+1$\,SD of rater age. $r_{\text{part}}$ removes a quadratic function of pooled human valence from both vectors and is the headline quantity; $r_{\max}$ is the attenuation ceiling implied by the posterior SD of $\Delta H$.}
\label{tab:alignment_arousal}
\begin{tabular}{lrrrcrrrr}
\toprule
\rowcolor{titlebar}
Model & $n$ & $r$ & $r_{\text{part}}$ & 95\% CI & $r_{\max}$ & Flat & Sign & SD$(\Delta M)$ \\
\midrule
Llama-4-Scout & 900 & -0.028 & -0.016 & [-0.08, 0.04] & 0.696 & 0\% & 0.53 & 0.170 \\
Qwen3-VL-235B & 899 & -0.032 & -0.067 & [-0.13, -0.01] & 0.696 & 2\% & 0.51 & 0.136 \\
GPT-5.6-Luna-Pro & 894 & -0.062 & -0.082 & [-0.14, -0.02] & 0.694 & 5\% & 0.48 & 0.124 \\
Qwen3-VL-8B-Think & 893 & -0.086 & -0.083 & [-0.15, -0.02] & 0.696 & 2\% & 0.47 & 0.278 \\
Gemini-3-Flash & 894 & -0.104 & -0.084 & [-0.15, -0.01] & 0.696 & 5\% & 0.48 & 0.178 \\
Qwen3-VL-8B & 900 & -0.128 & -0.108 & [-0.16, -0.05] & 0.696 & 0\% & 0.47 & 0.095 \\
Qwen3.8-27B & 896 & -0.067 & -0.108 & [-0.16, -0.05] & 0.696 & 0\% & 0.48 & 0.144 \\
GPT-5-mini & 884 & -0.084 & -0.108 & [-0.17, -0.05] & 0.695 & 8\% & 0.47 & 0.164 \\
GPT-5.2 & 898 & -0.100 & -0.118 & [-0.18, -0.06] & 0.696 & 8\% & 0.48 & 0.117 \\
Claude-Opus-5 & 899 & -0.086 & -0.132 & [-0.19, -0.07] & 0.696 & 23\% & 0.47 & 0.082 \\
Gemini-3.8-Flash & 900 & -0.141 & -0.149 & [-0.23, -0.07] & 0.696 & 11\% & 0.44 & 0.115 \\
GLM-4.5V & 873 & -0.164 & -0.164 & [-0.22, -0.11] & 0.695 & 4\% & 0.42 & 0.292 \\
Grok-4.6 & 900 & -0.154 & -0.167 & [-0.23, -0.10] & 0.696 & 8\% & 0.46 & 0.147 \\
Qwen3-VL-32B & 900 & -0.170 & -0.190 & [-0.25, -0.13] & 0.696 & 0\% & 0.45 & 0.139 \\
Gemma-3-27B & 900 & -0.145 & -0.223 & [-0.28, -0.17] & 0.696 & 2\% & 0.48 & 0.172 \\
Claude-Sonnet-4.6 & 898 & -0.198 & -0.236 & [-0.30, -0.17] & 0.696 & 6\% & 0.43 & 0.127 \\
Gemma-4-31B & 900 & -0.283 & -0.289 & [-0.35, -0.23] & 0.696 & 1\% & 0.42 & 0.148 \\
\bottomrule
\end{tabular}
\\[2pt]
\begin{minipage}{\linewidth}\footnotesize CI by resampling images, 2{,}000 replicates. \emph{Flat} is the share of images with $|\Delta M|<0.01$.\end{minipage}
\end{table}

\subsubsection{Persona-shift strength and human alignment}

We find no association between stereotype strength and alignment across models ($r=-0.038$, $p=0.88$).

\subsubsection{Model identity and the structure of image ratings}

\begin{figure}[tbp]
\centering
\includegraphics[width=\linewidth]{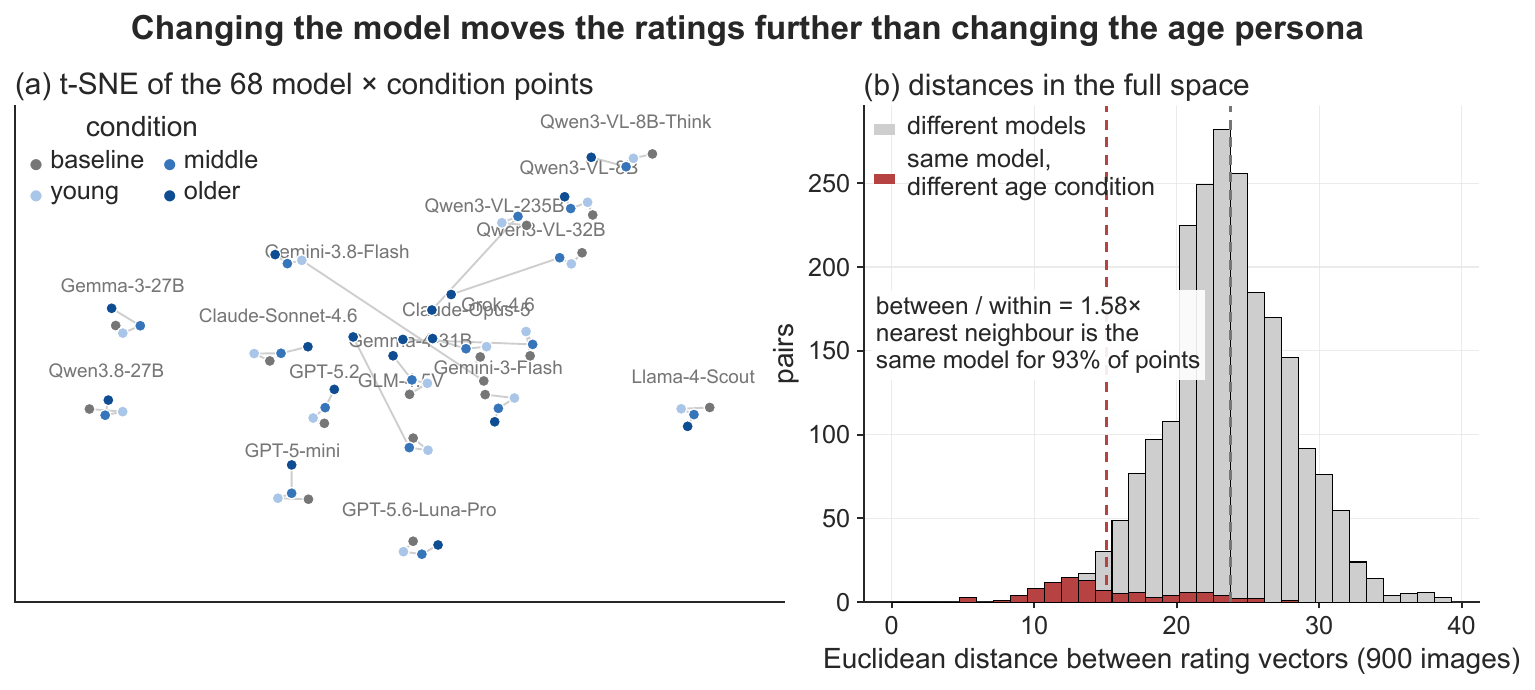}
\caption{Each point is one model under one condition, positioned by its 900 image ratings. Changing the model moves the ratings further than changing the persona: the nearest neighbour is the same model for 92.6\% of points.}
\label{fig:modelspace}
\end{figure}

In the space of ratings, placing each of the 68 model$\times$condition cells at its 900-dimensional vector of image ratings, the mean distance between two conditions of the same model is $15.1$ and between different models $23.8$ (ratio $1.58$), and for 92.6\% of cells the nearest neighbour is the same model under a different persona (Figure~\ref{fig:modelspace}): changing the model moves the ratings further than changing the persona. The images themselves, embedded in the same space, organise by human valence and separate flagged content.

\subsection{Valence--arousal geometry}

\label{app:geometry}

\begin{figure}[tbp]
\centering
\includegraphics[width=\linewidth]{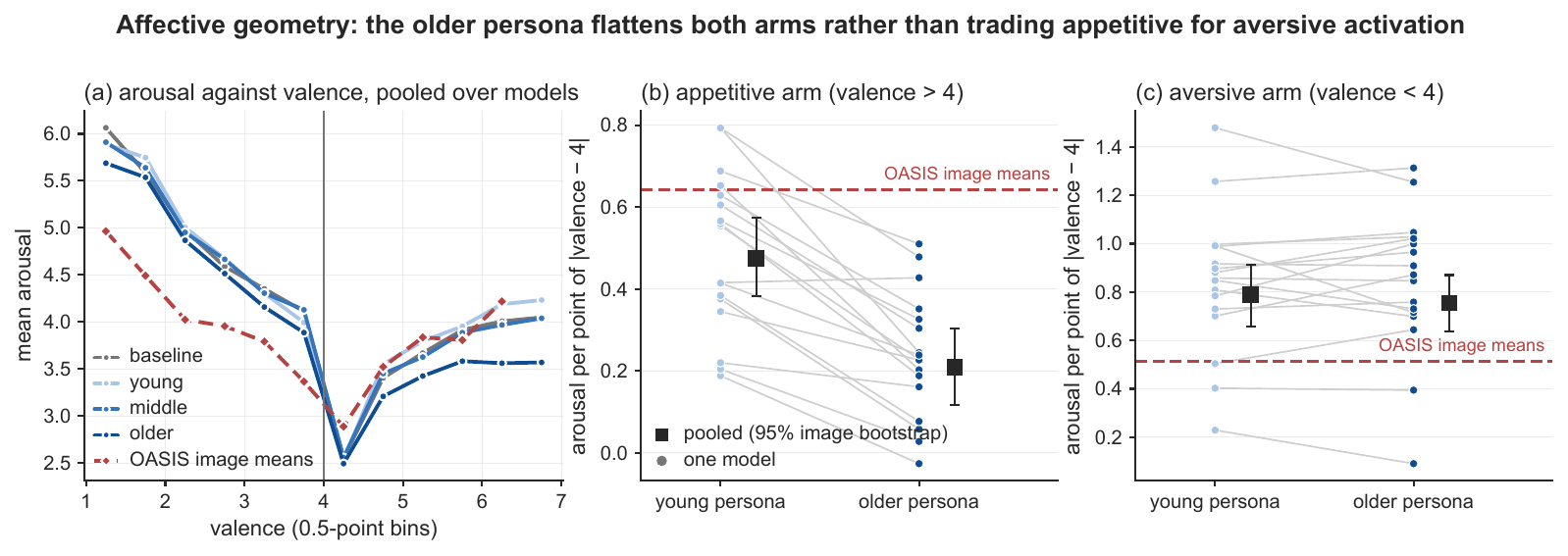}
\caption{Arousal against valence by condition, pooled over models, with the OASIS image means (a); appetitive (b) and aversive (c) arm slopes under the young and older personas, one line per model, pooled estimate with 95\% image-bootstrap interval.}
\label{fig:geometry}
\end{figure}

\citet{keil2009changes} predict that with age the appetitive arm of the valence--arousal relation flattens and the aversive arm steepens. Using each model's own valence and arousal from the same completion (Table~\ref{tab:geometry}, Figure~\ref{fig:geometry}), the appetitive arm flattens under the older persona for $\eGeoAppFlatter$ of 17 models, from a pooled slope of $\eGeoAppYoung$ ($[\eGeoAppLoYoung,\eGeoAppHiYoung]$) to $\eGeoAppOlder$ ($[\eGeoAppLoOlder,\eGeoAppHiOlder]$) arousal points per point of valence. The aversive-arm result depends on the statistic: the pooled regression slope barely changes ($\eGeoAveYoung$ against $\eGeoAveOlder$, $\eGeoAveSteeper$ of 17 models up), whereas the correlation-based coupling rises for all 17 (Appendix~\ref{app:rb-geometry}); $\eGeoKF$ models show both halves of the predicted pattern by the slope measure. Because the two measures disagree and OASIS cannot measure within-person coupling, we treat this analysis as exploratory and draw no conclusion from it. The pooled human image means have arm slopes of $\eGeoAppHuman$ and $\eGeoAveHuman$; these are across-image slopes from disjoint rater groups, not the within-person coupling of the original study, and are shown for scale only.

\begin{table}[tbp]
\centering
\small
\caption{Affective geometry by persona. For each model and condition, arousal is regressed on distance from the neutral point separately on the pleasant side (\emph{appetitive arm}, images with valence $>4$) and the unpleasant side (\emph{aversive arm}, valence $<4$), using that model's own valence and arousal from the same completion. \citet{keil2009changes} predict that with age the appetitive arm flattens and the aversive arm steepens; \emph{K\&F} marks models showing both. The human row uses the pooled OASIS image means, which come from disjoint rater groups and are therefore an across-image geometry rather than a within-rater one.}
\label{tab:geometry}
\begin{tabular}{lrrrrrrc}
\toprule
\rowcolor{titlebar}
 & \multicolumn{3}{c}{Appetitive arm} & \multicolumn{3}{c}{Aversive arm} & \\
\cmidrule(lr){2-4}\cmidrule(lr){5-7}
\rowcolor{titlebar}
Model & young & older & $\Delta$ & young & older & $\Delta$ & K\&F \\
\midrule
Claude-Opus-5 & 0.19 & -0.03 & -0.21 & 0.81 & 0.70 & -0.11 & no \\
Claude-Sonnet-4.6 & 0.65 & 0.20 & -0.45 & 0.86 & 0.85 & -0.01 & no \\
Gemini-3-Flash & 0.55 & 0.23 & -0.33 & 0.92 & 0.91 & -0.01 & no \\
Gemini-3.8-Flash & 0.38 & 0.06 & -0.32 & 0.99 & 0.72 & -0.28 & no \\
Gemma-3-27B & 0.34 & 0.23 & -0.12 & 1.48 & 1.25 & -0.22 & no \\
Gemma-4-31B & 0.20 & 0.03 & -0.18 & 0.50 & 0.64 & +0.14 & yes \\
GLM-4.5V & 0.79 & 0.24 & -0.55 & 0.70 & 0.87 & +0.17 & yes \\
GPT-5-mini & 0.79 & 0.48 & -0.31 & 1.26 & 1.31 & +0.06 & yes \\
GPT-5.2 & 0.61 & 0.30 & -0.30 & 1.00 & 1.03 & +0.03 & yes \\
GPT-5.6-Luna-Pro & 0.42 & 0.43 & +0.01 & 0.99 & 1.05 & +0.06 & no \\
Grok-4.6 & 0.56 & 0.19 & -0.37 & 0.78 & 1.00 & +0.21 & yes \\
Llama-4-Scout & 0.57 & 0.33 & -0.24 & 0.23 & 0.09 & -0.14 & no \\
Qwen3-VL-235B & 0.41 & 0.24 & -0.18 & 0.73 & 0.76 & +0.03 & yes \\
Qwen3-VL-32B & 0.38 & 0.08 & -0.31 & 0.85 & 0.73 & -0.12 & no \\
Qwen3-VL-8B & 0.22 & 0.16 & -0.06 & 0.40 & 0.39 & -0.01 & no \\
Qwen3-VL-8B-Think & 0.63 & 0.35 & -0.28 & 0.88 & 1.02 & +0.14 & yes \\
Qwen3.8-27B & 0.69 & 0.51 & -0.18 & 0.90 & 0.96 & +0.07 & yes \\
\midrule
OASIS raters (pooled means) & 0.64 &  &  & 0.51 &  &  &  \\
\bottomrule
\end{tabular}
\\[2pt]\begin{minipage}{\linewidth}\footnotesize Models showing the Keil--Freund pattern: 8/17.\end{minipage}
\end{table}

\FloatBarrierAbove
\section{Shared Age Responses and Image Content}

\label{app:shared-responses}

We next ask whether models share an image-specific age response and which image properties are associated with that response. We report agreement across models before examining the content predictors.

\begin{table}[tbp]
\centering
\caption{The content gate resolved: the older-minus-young shift on the sensitive flag crossed with pooled human valence (unpleasant $<4$, pleasant $\ge 4$). Flagged-unpleasant images are mostly violence, injury and death ($n=113$); flagged-pleasant images are mostly nudity and erotic content (nude man, nude woman, nude couple, BDSM, massage) with a few cemetery and memorial scenes ($n=54$). Unflagged images: unpleasant $n=194$, pleasant $n=539$. The largest persona effect anywhere in the study is on the flagged-pleasant images, which the older persona rates close to a full point less pleasant; unpleasant images move far less whether flagged or not. The older human raters rate the flagged-pleasant images \emph{more} pleasant than the young raters do.}
\label{tab:gate}
\resizebox{\linewidth}{!}{
\begin{tabular}{lrrrrrrrr}
\toprule
\rowcolor{titlebar}
 & \multicolumn{4}{c}{Valence, older $-$ young} & \multicolumn{4}{c}{Arousal, older $-$ young} \\
\cmidrule(lr){2-5}\cmidrule(lr){6-9}
\rowcolor{titlebar}
Source & unfl.\,unpl. & unfl.\,pleas. & flag.\,unpl. & flag.\,pleas. & unfl.\,unpl. & unfl.\,pleas. & flag.\,unpl. & flag.\,pleas. \\
\midrule
\textbf{OASIS raters} & \textbf{-0.107} & \textbf{+0.227} & \textbf{-0.231} & \textbf{+0.443} & \textbf{+0.211} & \textbf{-0.075} & \textbf{+0.451} & \textbf{-0.266} \\
\quad 95\% CI & {\scriptsize [-0.36, 0.20]} & {\scriptsize [0.02, 0.43]} & {\scriptsize [-0.47, 0.00]} & {\scriptsize [0.10, 0.86]} & {\scriptsize [-0.27, 0.69]} & {\scriptsize [-0.46, 0.38]} & {\scriptsize [-0.15, 1.14]} & {\scriptsize [-0.78, 0.22]} \\
\midrule
Qwen3-VL-235B & -0.141 & -0.131 & -0.333 & -1.270 & \ns{-0.062} & -0.301 & -0.173 & -0.496 \\
Qwen3-VL-32B & \ns{-0.038} & -0.158 & -0.213 & -1.491 & -0.197 & -0.281 & -0.287 & -0.796 \\
Grok-4.6 & -0.181 & \ns{-0.044} & -0.354 & -1.006 & -0.155 & -0.415 & -0.319 & -1.113 \\
Gemma-3-27B & \ns{+0.026} & -0.124 & \ns{-0.078} & -1.309 & -0.325 & -0.299 & -0.463 & -1.574 \\
GLM-4.5V & -0.090 & -0.089 & -0.158 & -1.085 & -0.403 & -0.455 & -0.401 & -1.532 \\
GPT-5-mini & \ns{-0.039} & -0.086 & -0.163 & -0.910 & -0.285 & -0.548 & -0.515 & -0.988 \\
Claude-Sonnet-4.6 & -0.151 & \ns{-0.005} & -0.247 & -0.996 & -0.109 & -0.275 & -0.206 & -0.977 \\
Claude-Opus-5 & -0.101 & -0.035 & -0.188 & -0.828 & -0.063 & -0.210 & -0.149 & -0.837 \\
Llama-4-Scout & -0.126 & \ns{-0.020} & -0.232 & -0.393 & -0.161 & -0.134 & -0.234 & -0.326 \\
GPT-5.2 & \ns{+0.018} & \ns{-0.014} & -0.155 & -1.122 & -0.130 & -0.235 & -0.247 & -0.544 \\
GPT-5.6-Luna-Pro & \ns{+0.026} & -0.080 & \ns{-0.029} & -0.496 & -0.084 & -0.183 & -0.212 & -0.646 \\
Gemma-4-31B & \ns{-0.030} & \ns{+0.047} & -0.204 & -0.876 & -0.366 & -0.375 & -0.492 & -1.067 \\
Gemini-3.8-Flash & \ns{-0.036} & \ns{+0.015} & -0.134 & -0.494 & -0.171 & -0.289 & -0.258 & -0.544 \\
Gemini-3-Flash & -0.103 & +0.117 & -0.262 & -0.713 & -0.107 & -0.450 & -0.141 & -1.409 \\
Qwen3-VL-8B & \ns{+0.049} & \ns{-0.011} & \ns{-0.027} & -0.343 & -0.284 & -0.262 & -0.448 & -0.370 \\
Qwen3.8-27B & \ns{+0.044} & \ns{+0.012} & \ns{-0.031} & -0.298 & \ns{-0.065} & -0.178 & -0.111 & -0.357 \\
Qwen3-VL-8B-Think & +0.209 & \ns{+0.039} & \ns{-0.098} & -0.740 & -0.320 & -0.433 & -0.558 & -0.756 \\
\midrule
Models with the human sign & 11/17 & 5/17 & 17/17 & 0/17 & 0/17 & 17/17 & 0/17 & 17/17 \\
Models below the human 95\% CI & 0/17 & 14/17 & 0/17 & 17/17 & 6/17 & 1/17 & 15/17 & 9/17 \\
\bottomrule
\end{tabular}}
\\[2pt]\begin{minipage}{\linewidth}\footnotesize Human intervals are wide on the flagged subsets because only 25 (valence) and 16 (arousal) raters are aged 60 or over. Model entries in \textcolor{black!45}{grey} are not significant ($q\ge.05$), tested as in Table~\ref{tab:human_vs_model}.\end{minipage}
\end{table}

\subsection{Agreement across models}

\label{app:consensus}

\begin{figure}[tbp]
\centering
\includegraphics[width=\linewidth]{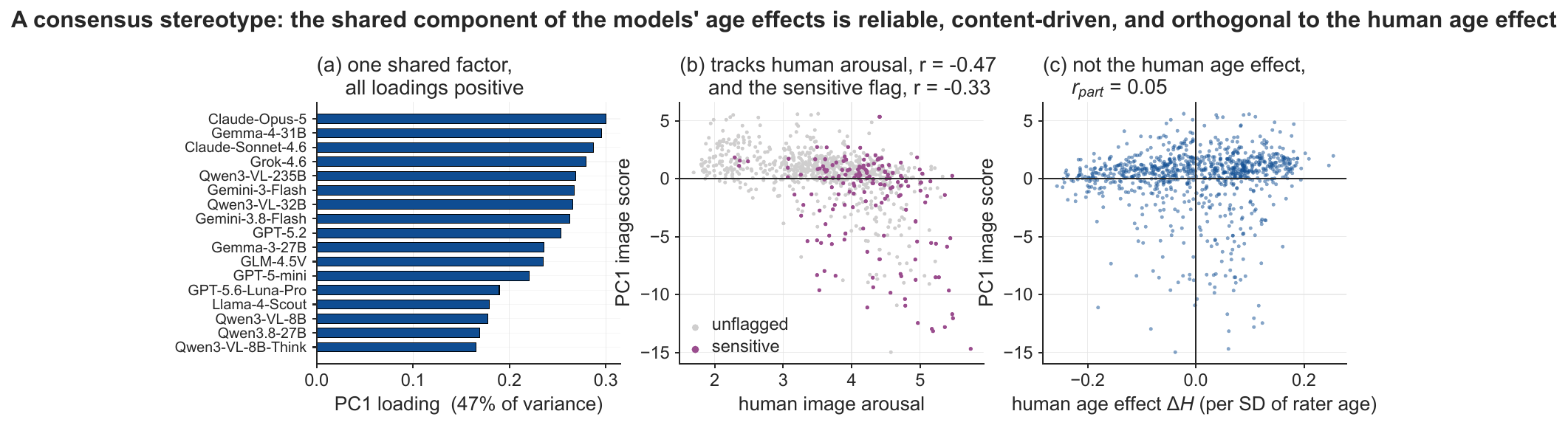}
\caption{The first principal component of the image$\times$model $\Delta M$ matrix: loadings (a), image scores against human arousal (b) and against the human age effect (c).}
\label{fig:consensus_factor}
\end{figure}

A principal component analysis of the image$\times$model matrix puts $\eConsPCone\%$ of the variance on a single component on which all $\eConsLoadPos$ models load positively, with the second at $\eConsPCtwo\%$ (Table~\ref{tab:consensus}). Clustering the pairwise correlations does not follow developers. The first component correlates $\eConsRpcArousal$ with human image arousal, $\eConsRpcSens$ with the sensitive flag and $\eConsRpcHv$ with human valence (Figure~\ref{fig:consensus_factor}); a t-SNE of the images in the space of all 34 model age effects shows the same organisation by flagged content. On arousal the pairwise agreement is weaker ($\eConsPairRAro$, first component $\eConsPConeAro\%$) and the consensus is negatively related to the human effect ($\eConsRpartAro$, $[\eConsRpartAroLo,\eConsRpartAroHi]$), subject to the caveat on the arousal reference in Appendix~\ref{app:alignment}.

\begin{table}[tbp]
\centering
\small
\caption{Between-model agreement on the image-specific age effect $\Delta M$ (list-centred). The pairwise correlation is taken over the 136 model pairs and split by developer with a permutation test on the developer labels. PC1 and PC2 are the first two components of the image$\times$model matrix. The split-half reliability is that of the 17-model consensus $\Delta M$ (Spearman--Brown, median of 500 random splits of the models). The consensus is reliable and tracks human image arousal and the sensitive flag, not human valence, and its partial correlation with the human age effect (image bootstrap CI) is as null as any single model's.}
\label{tab:consensus}
\begin{tabular}{lrr}
\toprule
\rowcolor{titlebar}
 & Valence & Arousal \\
\midrule
Images & 897 & 897 \\
Mean pairwise $r$ over 136 model pairs & 0.428 & 0.207 \\
\quad range & [0.16, 0.71] & [0.01, 0.52] \\
\quad same-developer pairs & 0.424 & 0.203 \\
\quad cross-developer pairs & 0.429 & 0.207 \\
\quad permutation $p$ (developer labels) & 0.54 & 0.53 \\
PC1 variance share & 47.0\% & 27.4\% \\
PC2 variance share & 6.1\% & 6.6\% \\
Models loading positively on PC1 & 17/17 & 17/17 \\
Split-half reliability of the consensus & 0.93 & 0.79 \\
$r$(PC1, human valence) & 0.01 & -0.27 \\
$r$(PC1, human arousal) & -0.47 & -0.37 \\
$r$(PC1, sensitive flag) & -0.33 & -0.27 \\
$\eta^2$(PC1, category) & 0.08 & 0.02 \\
$r$(PC1, $\Delta H$), valence partialled & 0.05 & -0.27 \\
Consensus $\Delta M$: $r_{\text{part}}$ with $\Delta H$ & 0.050 [-0.02, 0.13] & -0.267 [-0.33, -0.21] \\
\bottomrule
\end{tabular}
\end{table}

\begin{figure}[tbp]
\centering
\includegraphics[width=\linewidth]{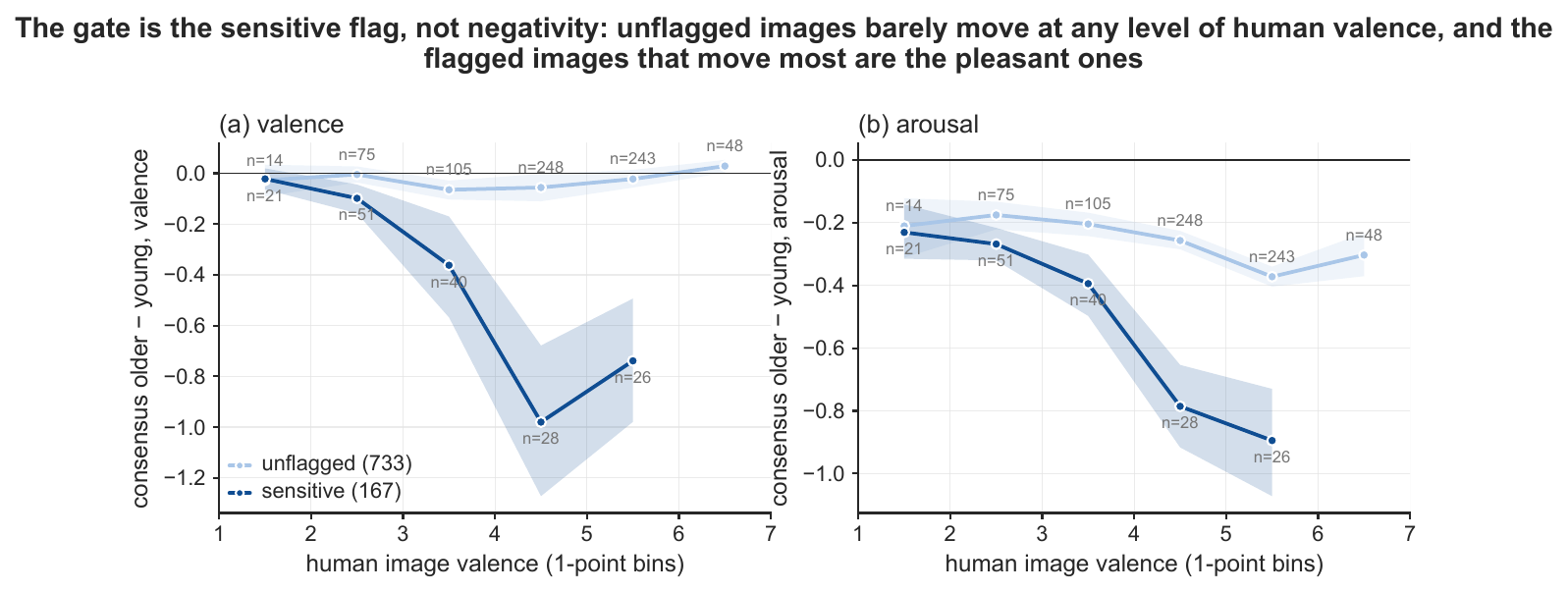}
\caption{Consensus older-minus-young valence shift by one-point bin of pooled human valence, for flagged and unflagged images separately, with 95\% image-bootstrap bands. Unflagged images barely move at any valence level; flagged images move most where humans rate them as pleasant.}
\label{fig:content_gate}
\end{figure}

\subsection{Content predictors and sensitive images}

\label{app:gate}

\begin{figure}[tbp]
\centering
\includegraphics[width=\linewidth]{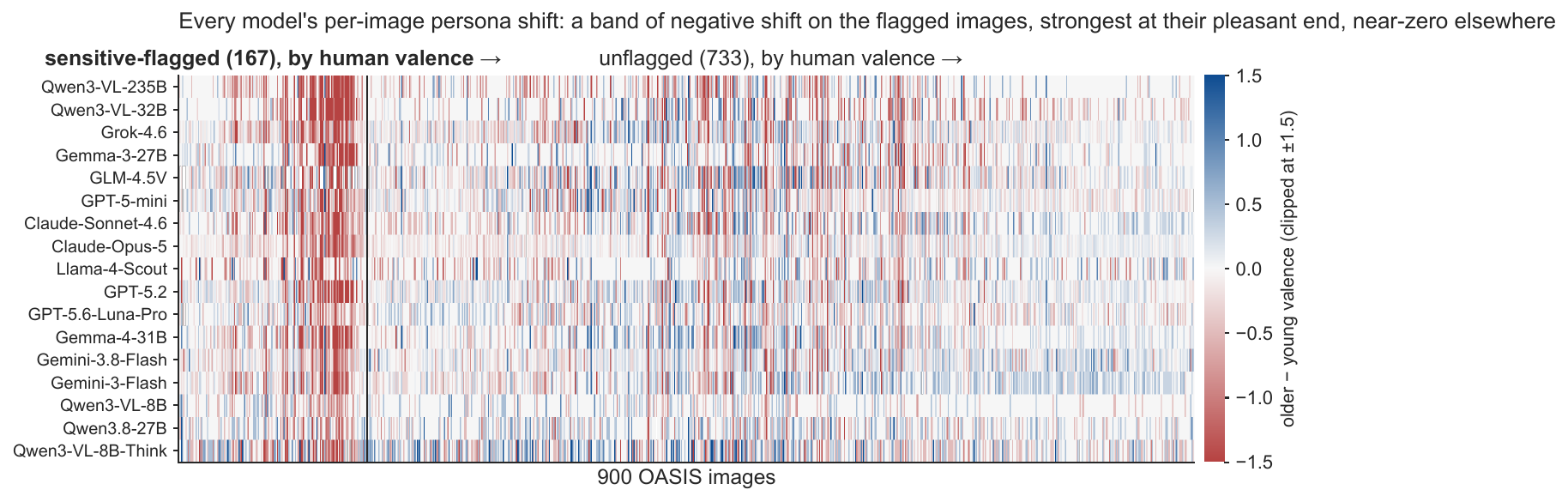}
\caption{Older-minus-young valence for every image (columns; flagged images first, each block ordered by human valence) and every model (rows, ordered by mean shift).}
\label{fig:image_heatmap}
\end{figure}

Over the 900 images, the sensitive flag alone explains $R^2=\ePredRtwoFlagValence$ of the consensus valence shift, a quadratic in human valence alone $\ePredRtwoValValence$, and the two together $\ePredRtwoValFlagValence$; adding human arousal and category raises the total to $\ePredRtwoFullValence$, with high-arousal images shifting more ($\beta=\ePredBetaAroValence$ per SD) and Animal images less than Person images (Table~\ref{tab:predictors}). The flag coefficient is $\ePredBetaFlagValence$ points ($[\ePredBetaFlagLoValence,\ePredBetaFlagHiValence]$) after all controls. Per model, the full set explains between $\ePredRtwoFullMin$ and $\ePredRtwoFullMax$ of the variance (median $\ePredRtwoFullMedian$; Table~\ref{tab:predictors_per_model}). Figure~\ref{fig:image_heatmap} shows the raw pattern: a band of negative shift across the flagged columns and near-zero elsewhere, repeated in every row.

On arousal, the human subset contrasts point in the direction \citet{keil2009changes} predict, higher for unpleasant images ($\eHumArousalFlagUnpl$ flagged, $\eHumArousalUnflUnpl$ unflagged) and lower for pleasant ones ($\eHumArousalFlagPleas$, $\eHumArousalUnflPleas$), although every such interval includes zero; every model lowers arousal in all four cells.

\begin{table}[tbp]
\centering
\small
\caption{Image-level predictors of the consensus persona shift (older $-$ young, averaged over the 17 models). Ordinary least squares on the 900 images with image-bootstrap intervals (2{,}000 replicates); continuous predictors are standardised, the sensitive flag and the category indicators (reference: Person) are 0/1. Coefficients are in rating points. The flag carries the effect: it explains more variance alone than the full valence polynomial does, and adding valence to the flag adds little.}
\label{tab:predictors}
\begin{tabular}{lrcrc}
\toprule
\rowcolor{titlebar}
Predictor & \multicolumn{2}{c}{Valence shift} & \multicolumn{2}{c}{Arousal shift} \\
\cmidrule(lr){2-3}\cmidrule(lr){4-5}
\rowcolor{titlebar}
 & $\beta$ & 95\% CI & $\beta$ & 95\% CI \\
\midrule
human valence (per SD) & -0.055 & [-0.078, -0.036] & -0.105 & [-0.123, -0.087] \\
human valence$^2$ (z) & +0.164 & [0.136, 0.194] & +0.040 & [0.019, 0.063] \\
human arousal (z) & -0.252 & [-0.293, -0.212] & -0.115 & [-0.141, -0.089] \\
sensitive flag & -0.185 & [-0.271, -0.099] & -0.180 & [-0.236, -0.124] \\
rater disagreement (z) & -0.019 & [-0.053, 0.018] & -0.019 & [-0.042, 0.005] \\
Scene & +0.062 & [0.005, 0.116] & +0.018 & [-0.027, 0.061] \\
Object & +0.010 & [-0.047, 0.060] & -0.060 & [-0.108, -0.014] \\
Animal & +0.255 & [0.196, 0.315] & +0.075 & [0.030, 0.121] \\
\midrule
$R^2$, sensitive flag only & 0.112 &  & 0.069 &  \\
$R^2$, human valence only & 0.019 &  & 0.055 &  \\
$R^2$, valence $+$ flag & 0.160 &  & 0.191 &  \\
$R^2$, full model & 0.432 &  & 0.302 &  \\
\bottomrule
\end{tabular}
\end{table}

\begin{table}[tbp]
\centering
\small
\caption{Per-model version of Table~\ref{tab:predictors}: variance in the older-minus-young shift explained by the sensitive flag alone, by a quadratic in human valence alone, and by the full predictor set, with the flag coefficient (rating points). Left block valence, right block arousal.}
\label{tab:predictors_per_model}
\begin{tabular}{lrrrrrrrr}
\toprule
\rowcolor{titlebar}
Model & $R^2_{\text{flag}}$ & $R^2_{\text{val}}$ & $R^2_{\text{full}}$ & $\beta_{\text{flag}}$ & $R^2_{\text{flag}}$ & $R^2_{\text{val}}$ & $R^2_{\text{full}}$ & $\beta_{\text{flag}}$ \\
\midrule
Claude-Opus-5 & 0.12 & 0.03 & 0.38 & -0.18 & 0.07 & 0.06 & 0.23 & -0.26 \\
Grok-4.6 & 0.07 & 0.02 & 0.35 & -0.14 & 0.03 & 0.06 & 0.17 & -0.29 \\
Claude-Sonnet-4.6 & 0.08 & 0.03 & 0.34 & -0.15 & 0.03 & 0.06 & 0.23 & -0.23 \\
Qwen3-VL-32B & 0.08 & 0.03 & 0.31 & -0.39 & 0.02 & 0.01 & 0.12 & -0.12 \\
Gemma-4-31B & 0.09 & 0.00 & 0.31 & -0.21 & 0.04 & 0.00 & 0.17 & -0.17 \\
Qwen3-VL-235B & 0.08 & 0.03 & 0.30 & -0.26 & 0.00 & 0.05 & 0.09 & -0.09 \\
Gemini-3-Flash & 0.09 & 0.05 & 0.29 & -0.18 & 0.01 & 0.13 & 0.22 & -0.29 \\
GPT-5.2 & 0.07 & 0.01 & 0.25 & -0.31 & 0.01 & 0.02 & 0.08 & -0.13 \\
Gemini-3.8-Flash & 0.04 & 0.02 & 0.24 & -0.03 & 0.01 & 0.02 & 0.07 & -0.08 \\
Gemma-3-27B & 0.04 & 0.02 & 0.24 & -0.25 & 0.10 & 0.01 & 0.21 & -0.41 \\
GLM-4.5V & 0.02 & 0.00 & 0.19 & -0.15 & 0.01 & 0.01 & 0.07 & -0.22 \\
GPT-5-mini & 0.04 & 0.01 & 0.18 & -0.17 & 0.02 & 0.03 & 0.11 & -0.20 \\
Qwen3-VL-8B-Think & 0.04 & 0.01 & 0.13 & -0.31 & 0.01 & 0.00 & 0.02 & -0.25 \\
Llama-4-Scout & 0.03 & 0.01 & 0.11 & -0.07 & 0.01 & 0.00 & 0.03 & -0.04 \\
GPT-5.6-Luna-Pro & 0.01 & 0.02 & 0.10 & -0.13 & 0.03 & 0.01 & 0.07 & -0.21 \\
Qwen3-VL-8B & 0.03 & 0.01 & 0.07 & -0.12 & 0.03 & 0.02 & 0.05 & -0.08 \\
Qwen3.8-27B & 0.01 & 0.00 & 0.07 & -0.06 & 0.00 & 0.01 & 0.03 & -0.03 \\
\bottomrule
\end{tabular}
\end{table}

\FloatBarrierAbove
\section{Emotion Labels, Descriptions, and Model Comparisons}

\label{app:language-results}

We examine how age conditioning changes emotion labels and descriptions, then relate these changes to numerical ratings and model properties. As described in Appendix~\ref{app:protocol}, the ratings precede the description and label in each response. We therefore interpret these outputs as parts of the same elicitation.

\subsection{Emotion labels}

\label{app:emotion}

\subsubsection{Transitions between emotion groups}

Restricting to images with a valid in-vocabulary label under both personas, between 105 and 208 images change label between young and older (median 163 of roughly 890 paired images). The aversive shift is significant after correction for 13 of 17 models and the appetitive loss for 6 of 17 (Table~\ref{tab:emotion}); Grok-4.6 gains a net 61 aversive images, Qwen3-VL-235B 53 and Claude-Opus-5 43.

\begin{table}[tbp]
\centering
\small
\caption{Categorical emotion shift between the young and older conditions on the same image. Y/O gives the count assigned to the grouping under young only and under older only; the test is a two-sided exact binomial on those discordant pairs.}
\label{tab:emotion}
\begin{tabular}{lrrrrrr}
\toprule
\rowcolor{titlebar}
Model & $n$ pairs & Changed & Appet. Y/O & $q$ & Avers. Y/O & $q$ \\
\midrule
Qwen3-VL-32B & 900 & 202 & 62/33 & 0.022 & 16/73 & $<$0.001 \\
Grok-4.6 & 900 & 193 & 33/44 & \ns{0.393} & 5/66 & $<$0.001 \\
Claude-Sonnet-4.6 & 898 & 206 & 65/44 & \ns{0.104} & 12/64 & $<$0.001 \\
Qwen3-VL-235B & 899 & 208 & 60/35 & 0.038 & 10/63 & $<$0.001 \\
Gemma-3-27B & 900 & 195 & 71/24 & $<$0.001 & 21/54 & $<$0.001 \\
GPT-5.2 & 900 & 175 & 55/29 & 0.026 & 10/53 & $<$0.001 \\
Gemma-4-31B & 900 & 197 & 34/46 & \ns{0.371} & 12/51 & $<$0.001 \\
GLM-4.5V & 818 & 163 & 48/40 & \ns{0.593} & 25/47 & 0.017 \\
Claude-Opus-5 & 899 & 140 & 51/27 & 0.030 & 3/46 & $<$0.001 \\
GPT-5-mini & 888 & 155 & 42/37 & \ns{0.740} & 11/39 & $<$0.001 \\
Gemini-3-Flash & 894 & 150 & 42/23 & \ns{0.060} & 7/38 & $<$0.001 \\
Gemini-3.8-Flash & 877 & 150 & 41/34 & \ns{0.593} & 6/36 & $<$0.001 \\
Qwen3-VL-8B-Think & 896 & 175 & 27/57 & 0.012 & 25/29 & \ns{0.683} \\
GPT-5.6-Luna-Pro & 895 & 122 & 28/26 & \ns{0.892} & 8/25 & 0.006 \\
Llama-4-Scout & 899 & 118 & 36/19 & \ns{0.064} & 12/24 & \ns{0.079} \\
Qwen3.8-27B & 898 & 156 & 42/32 & \ns{0.419} & 31/22 & \ns{0.289} \\
Qwen3-VL-8B & 900 & 105 & 20/17 & \ns{0.789} & 25/16 & \ns{0.239} \\
\bottomrule
\end{tabular}
\\[2pt]
\begin{minipage}{\linewidth}\footnotesize Only images with a valid in-vocabulary label under \emph{both} conditions contribute. $q$ is the Benjamini--Hochberg adjusted $p$-value across models; values of $q\ge.05$ are shown in \textcolor{black!45}{grey}.\end{minipage}
\end{table}

\begin{figure}[tbp]
\centering
\includegraphics[width=\linewidth]{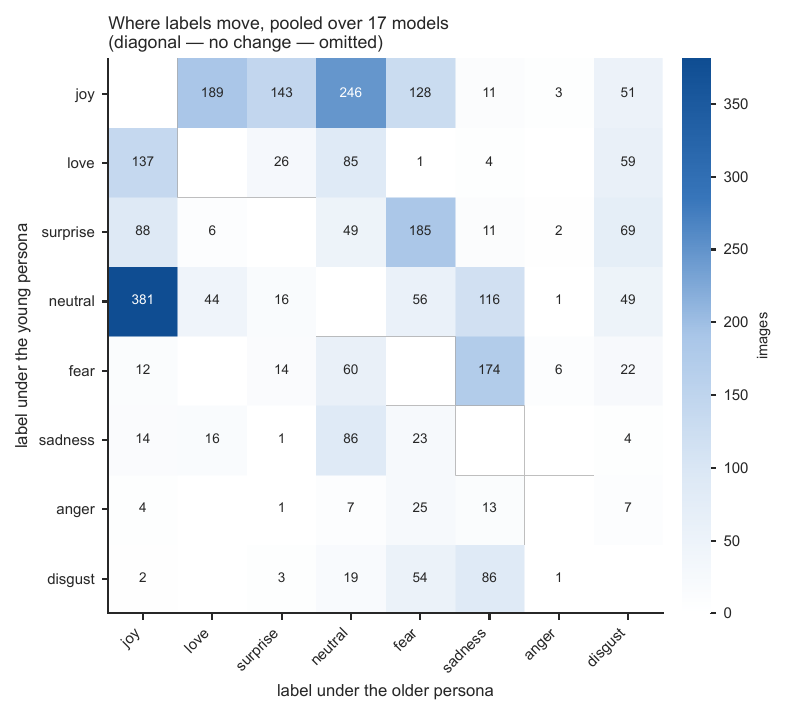}
\caption{Pooled label transitions from the young to the older persona. The diagonal, which carries most images, is omitted so that the off-diagonal structure is visible.}
\label{fig:transitions}
\end{figure}

\subsubsection{Changes in individual labels}

\begin{figure}[tbp]
\centering
\includegraphics[width=0.7\linewidth]{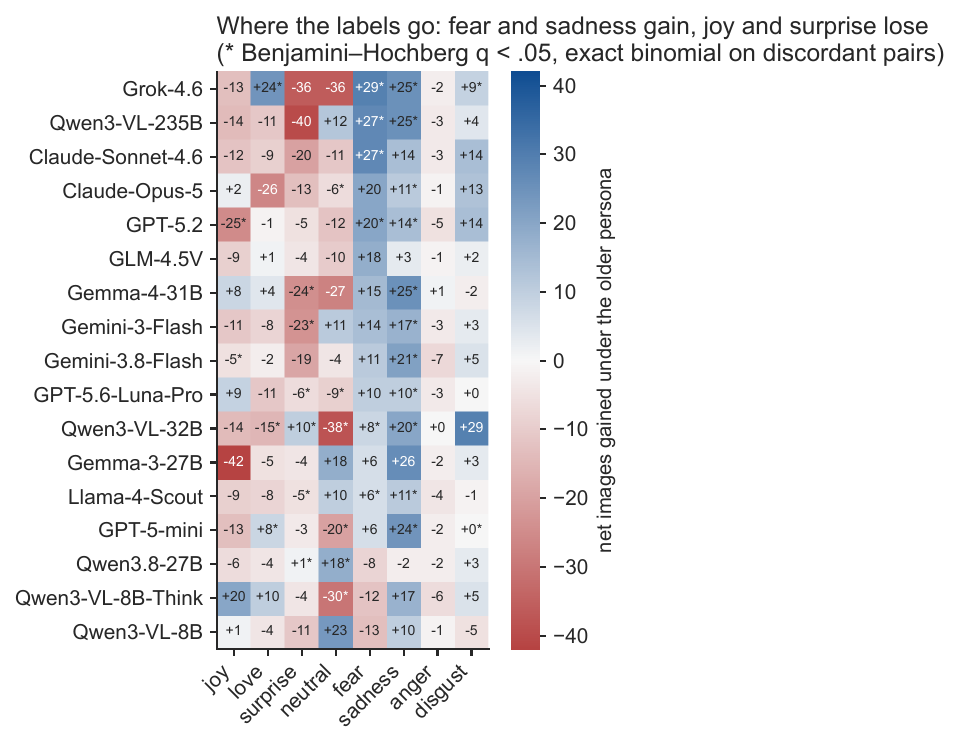}
\caption{Net images gained by each label under the older persona, per model.}
\label{fig:label_net}
\end{figure}

Per label (Table~\ref{tab:label_net}, Figure~\ref{fig:label_net}), the gain is carried by sadness ($\eSadNetPooled$) and fear ($\eFearNetPooled$; positive for $\eFearPos$ of 17 models), with a smaller gain for disgust ($\eDisgustNetPooled$) and a loss for anger ($\eAngerNetPooled$); surprise ($\eSurpriseNetPooled$), joy ($\eJoyNetPooled$) and neutral ($\eNeutralNetPooled$) lose most.

\begin{table}[tbp]
\centering
\small
\caption{Net images gained by each emotion label under the older persona (assigned under older only minus under young only, on images with a valid label in both conditions). Sadness and fear are the labels that gain, disgust gains modestly, anger does not; surprise, joy and neutral are shed. Significance: exact binomial on the discordant pairs, Benjamini--Hochberg over the 136 model$\times$label tests; entries in \textcolor{black!45}{grey} are not significant ($q\ge.05$), all others have $q<.05$.}
\label{tab:label_net}
\begin{tabular}{lrrrrrrrr}
\toprule
\rowcolor{titlebar}
Model & joy & love & surprise & neutral & fear & sadness & anger & disgust \\
\midrule
Grok-4.6 & \ns{-13} & +24 & -36 & -36 & +29 & +25 & -2 & \ns{+9} \\
Qwen3-VL-235B & \ns{-14} & \ns{-11} & -40 & \ns{+12} & +27 & +25 & -3 & \ns{+4} \\
Claude-Sonnet-4.6 & \ns{-12} & \ns{-9} & \ns{-20} & \ns{-11} & +27 & +14 & -3 & \ns{+14} \\
Claude-Opus-5 & \ns{+2} & -26 & \ns{-13} & \ns{-6} & +20 & +11 & -1 & +13 \\
GPT-5.2 & -25 & \ns{-1} & \ns{-5} & \ns{-12} & +20 & +14 & -5 & \ns{+14} \\
GLM-4.5V & \ns{-9} & \ns{+1} & \ns{-4} & \ns{-10} & +18 & \ns{+3} & -1 & \ns{+2} \\
Gemma-4-31B & \ns{+8} & \ns{+4} & -24 & -27 & \ns{+15} & +25 & +1 & \ns{-2} \\
Gemini-3-Flash & \ns{-11} & \ns{-8} & -23 & \ns{+11} & \ns{+14} & +17 & -3 & \ns{+3} \\
Gemini-3.8-Flash & \ns{-5} & \ns{-2} & -19 & \ns{-4} & \ns{+11} & +21 & -7 & \ns{+5} \\
GPT-5.6-Luna-Pro & \ns{+9} & \ns{-11} & \ns{-6} & \ns{-9} & \ns{+10} & \ns{+10} & -3 & \ns{+0} \\
Qwen3-VL-32B & \ns{-14} & -15 & \ns{+10} & -38 & \ns{+8} & +20 & +0 & +29 \\
GPT-5-mini & \ns{-13} & \ns{+8} & \ns{-3} & -20 & \ns{+6} & +24 & -2 & \ns{+0} \\
Gemma-3-27B & -42 & \ns{-5} & \ns{-4} & \ns{+18} & \ns{+6} & +26 & -2 & \ns{+3} \\
Llama-4-Scout & \ns{-9} & \ns{-8} & \ns{-5} & \ns{+10} & \ns{+6} & \ns{+11} & -4 & \ns{-1} \\
Qwen3.8-27B & \ns{-6} & \ns{-4} & \ns{+1} & \ns{+18} & \ns{-8} & \ns{-2} & -2 & \ns{+3} \\
Qwen3-VL-8B-Think & \ns{+20} & \ns{+10} & \ns{-4} & -30 & \ns{-12} & \ns{+17} & -6 & \ns{+5} \\
Qwen3-VL-8B & \ns{+1} & \ns{-4} & -11 & +23 & \ns{-13} & \ns{+10} & -1 & \ns{-5} \\
\textbf{Pooled} & \textbf{-133} & \textbf{-57} & \textbf{-206} & \textbf{-111} & \textbf{+184} & \textbf{+271} & \textbf{-44} & \textbf{+96} \\
\bottomrule
\end{tabular}
\end{table}

\subsubsection{Label diversity and composition}

The label distribution does not narrow: the median Jensen--Shannon divergence between young and older is $\eJSyoMedian$ bits, against $\eJSymMedian$ between young and middle (Table~\ref{tab:entropy}).

\begin{table}[tbp]
\centering
\small
\caption{Diversity of the emotion-label distribution by condition. $H$ is Shannon entropy in bits over the eight labels; JS is the Jensen--Shannon divergence (bits) between two conditions' label distributions. The older persona barely changes the diversity of labels but redistributes them: the share of \emph{fear} rises and that of \emph{neutral} falls for most models.}
\label{tab:entropy}
\begin{tabular}{lrrrrrrrr}
\toprule
\rowcolor{titlebar}
Model & $H_{\varnothing}$ & $H_Y$ & $H_M$ & $H_O$ & JS$(Y,O)$ & JS$(Y,M)$ & fear\,\% Y/O & neutral\,\% Y/O \\
\midrule
Grok-4.6 & 2.54 & 2.51 & 2.57 & 2.48 & 0.014 & 0.003 & 13.8/17.0 & 17.1/13.1 \\
Qwen3-VL-235B & 2.55 & 2.47 & 2.45 & 2.40 & 0.011 & 0.001 & 13.6/16.6 & 17.8/19.1 \\
Qwen3-VL-32B & 2.47 & 2.42 & 2.45 & 2.50 & 0.009 & 0.001 & 16.0/16.9 & 19.2/15.0 \\
Gemma-4-31B & 2.56 & 2.46 & 2.46 & 2.42 & 0.005 & 0.001 & 12.2/13.9 & 17.9/14.9 \\
Gemini-3.8-Flash & 2.59 & 2.56 & 2.52 & 2.51 & 0.005 & 0.002 & 14.2/15.6 & 18.4/18.0 \\
Claude-Sonnet-4.6 & 2.63 & 2.63 & 2.64 & 2.62 & 0.005 & 0.000 & 16.1/19.2 & 18.0/16.8 \\
Gemini-3-Flash & 2.59 & 2.55 & 2.52 & 2.50 & 0.004 & 0.003 & 13.8/15.3 & 16.9/18.1 \\
Qwen3-VL-8B-Think & 2.34 & 2.28 & 2.30 & 2.25 & 0.004 & 0.002 & 14.4/13.0 & 19.3/15.9 \\
Claude-Opus-5 & 2.66 & 2.64 & 2.64 & 2.63 & 0.004 & 0.000 & 14.8/17.0 & 15.8/15.1 \\
Qwen3-VL-8B & 2.23 & 2.15 & 2.14 & 2.05 & 0.004 & 0.003 & 11.4/10.0 & 23.7/26.2 \\
Gemma-3-27B & 2.54 & 2.51 & 2.55 & 2.50 & 0.003 & 0.002 & 20.0/20.7 & 22.0/24.0 \\
GPT-5.2 & 2.61 & 2.63 & 2.64 & 2.64 & 0.003 & 0.001 & 16.9/19.1 & 16.3/15.0 \\
GPT-5-mini & 2.52 & 2.48 & 2.47 & 2.47 & 0.002 & 0.001 & 14.6/15.4 & 9.7/7.6 \\
Llama-4-Scout & 2.45 & 2.42 & 2.43 & 2.38 & 0.001 & 0.001 & 11.0/11.7 & 29.2/30.4 \\
GLM-4.5V & 2.44 & 2.46 & 2.44 & 2.47 & 0.001 & 0.001 & 11.9/14.0 & 29.7/28.4 \\
GPT-5.6-Luna-Pro & 2.58 & 2.59 & 2.59 & 2.56 & 0.001 & 0.000 & 15.3/16.4 & 20.7/19.7 \\
Qwen3.8-27B & 2.50 & 2.56 & 2.51 & 2.54 & 0.001 & 0.001 & 15.2/14.3 & 29.0/31.0 \\
\bottomrule
\end{tabular}
\end{table}

\subsection{Emotion descriptions}

\label{app:text}

\subsubsection{Description length and word use}

\begin{figure}[tbp]
\centering
\includegraphics[width=\linewidth]{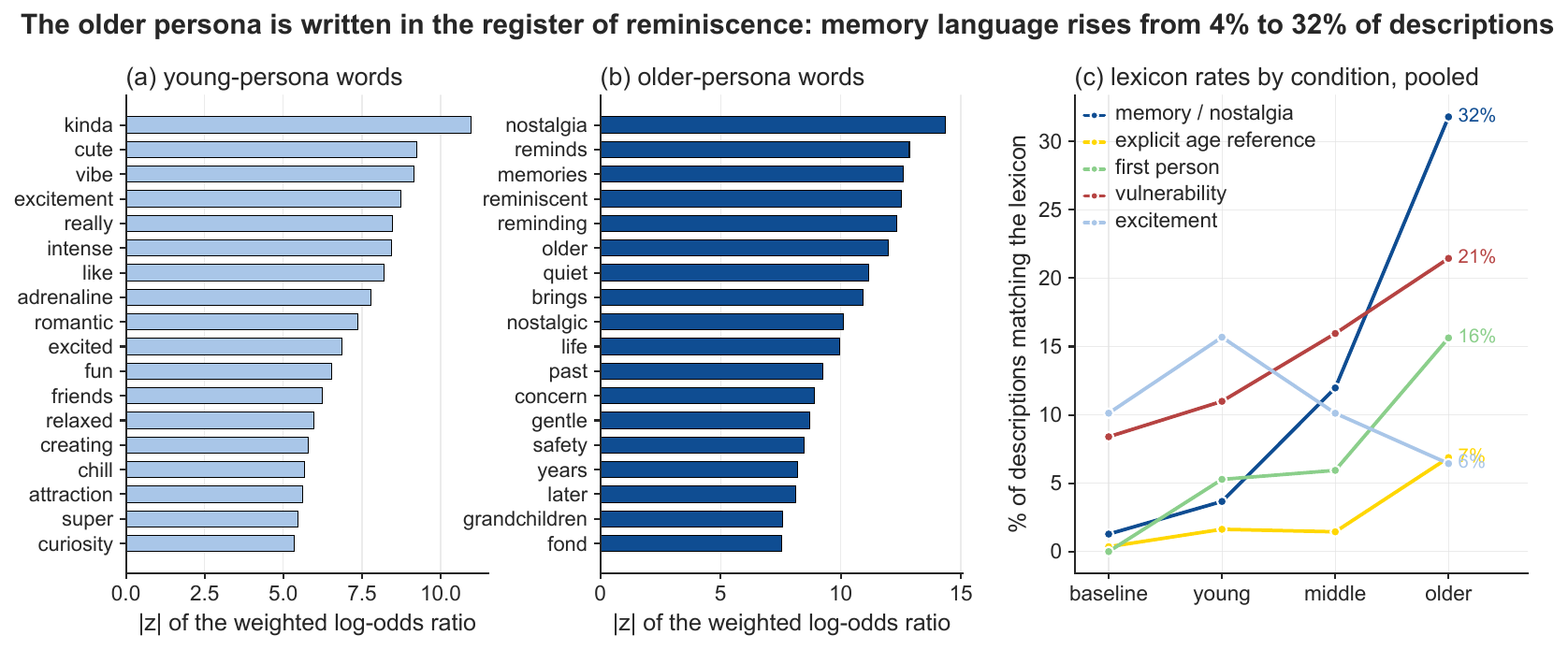}
\caption{Words most diagnostic of the young (a) and older (b) personas by weighted log-odds, and lexicon rates by condition pooled over models (c).}
\label{fig:lexicon}
\end{figure}

Pooled over models, the older persona lengthens the description (from $\eLexWordsYoung$ to $\eLexWordsOlder$ words, significant for $\eLexLengthSigModels$ of 17 models) and changes its register (Table~\ref{tab:lexicon}, Figure~\ref{fig:lexicon}). Memory language reaches $\eLexMemoryMaxRate\%$ of older-persona descriptions for \eLexMemoryMaxModel; explicit references to the rater's own age or generation rise from $\eLexExplicitAgeYoung\%$ to $\eLexExplicitAgeOlder\%$, and first-person framing from $\eLexFirstPersonYoung\%$ to $\eLexFirstPersonOlder\%$. The excitement lexicon falls for all $\eLexExciteDownModels$ models. A weighted log-odds analysis with an informative Dirichlet prior \citep{monroe2008fightin}, which does not depend on our lexica, returns the same picture (Table~\ref{tab:logodds}): the words most diagnostic of the young persona are \emph{\eLexTopYoung}.

\begin{table}[tbp]
\centering
\caption{What the older persona does to the free-text description. Each cell is young/older. \emph{words} is the mean description length; the remaining columns are the percentage of descriptions matching a lexicon: \emph{memory} (memories, reminds, nostalgic, reminiscent, the past, youth, decades), \emph{age ref.} (an explicit reference to the rater's age or generation), \emph{1st person}, \emph{excitement} (exciting, thrilling, adventure, fun), and \emph{vulnerability} (fragile, frail, mortality, loss, health, danger, worry). The lexica were fixed before the counts were run; the full regular expressions are in the analysis script.}
\label{tab:lexicon}
\resizebox{\linewidth}{!}{
\begin{tabular}{lrrrrrr}
\toprule
\rowcolor{titlebar}
Model & words Y/O & memory\,\% & age ref.\,\% & 1st person\,\% & excitement\,\% & vulnerab.\,\% \\
\midrule
Gemma-3-27B & 18.5/21.1 & 7.2/88.0 & 0.0/6.9 & 65.8/93.9 & 18.0/5.9 & 19.8/23.3 \\
Qwen3-VL-235B & 18.4/23.2 & 7.3/72.0 & 6.9/37.3 & 7.0/50.4 & 21.8/7.8 & 17.7/27.4 \\
Qwen3-VL-32B & 22.3/25.9 & 3.7/63.8 & 10.9/30.7 & 0.0/14.1 & 16.2/6.6 & 17.2/31.9 \\
Claude-Sonnet-4.6 & 20.8/25.3 & 5.9/53.9 & 2.6/10.6 & 1.2/18.3 & 22.9/6.8 & 14.7/28.3 \\
GPT-5-mini & 11.8/14.5 & 4.1/44.8 & 0.3/11.0 & 1.4/20.0 & 20.8/8.6 & 11.5/27.6 \\
Grok-4.6 & 10.1/11.4 & 4.2/33.3 & 0.7/1.9 & \ns{0.0/0.0} & 16.4/5.1 & 5.1/20.6 \\
GLM-4.5V & 16.0/19.3 & 2.7/33.0 & 0.2/2.9 & 0.6/25.9 & 13.7/4.0 & 7.2/23.2 \\
Gemma-4-31B & 12.1/13.7 & 2.9/22.7 & 0.4/2.2 & 0.1/4.7 & 12.2/3.8 & 5.8/17.3 \\
Claude-Opus-5 & 17.5/21.0 & 6.2/21.1 & \ns{0.9/1.3} & 0.4/4.2 & 22.9/11.1 & 12.6/20.4 \\
Qwen3-VL-8B-Think & 7.1/7.6 & 1.2/21.0 & \ns{0.7/0.9} & \ns{0.0/0.1} & 10.5/5.0 & 7.5/11.3 \\
Gemini-3-Flash & 15.3/16.3 & 5.1/20.8 & \ns{1.7/1.5} & 0.0/2.1 & 14.7/5.3 & 11.5/22.7 \\
GPT-5.2 & 13.2/15.6 & 1.6/17.4 & 0.3/2.2 & 10.6/27.2 & 14.3/7.0 & 15.0/30.4 \\
Qwen3.8-27B & 17.0/17.8 & 2.0/15.3 & 0.2/1.1 & \ns{0.1/0.0} & 11.7/5.5 & 8.5/17.0 \\
Gemini-3.8-Flash & \ns{10.0/10.1} & 3.9/12.4 & \ns{0.3/1.1} & \ns{0.0/0.0} & 12.2/3.1 & 6.8/19.3 \\
Llama-4-Scout & 5.8/6.5 & 1.0/9.7 & \ns{0.2/0.3} & 1.4/4.6 & 12.2/8.6 & 5.6/9.1 \\
Qwen3-VL-8B & 11.3/12.8 & 2.6/7.4 & 0.8/4.2 & \ns{0.7/0.2} & 12.6/7.7 & 10.1/14.0 \\
GPT-5.6-Luna-Pro & 11.6/12.2 & 0.8/4.7 & \ns{0.3/0.1} & \ns{0.1/0.0} & 13.2/7.5 & 10.2/20.7 \\
\textbf{Pooled} & 14.0/16.1 & 3.7/31.8 & 1.6/6.9 & 5.3/15.6 & 15.7/6.4 & 11.0/21.4 \\
\bottomrule
\end{tabular}}
\\[2pt]
\begin{minipage}{\linewidth}\footnotesize Significance: Wilcoxon (length) or exact binomial on discordant image pairs (flags), Benjamini--Hochberg across models within column; entries in \textcolor{black!45}{grey} are not significant ($q\ge.05$), all others have $q<.05$.\end{minipage}
\end{table}

\begin{table}[tbp]
\centering
\small
\caption{Words most over-represented in older-persona descriptions (left) and in young-persona descriptions (right), pooled over models: weighted log-odds ratio with an informative Dirichlet prior \citep{monroe2008fightin}, prior counts taken from the pooled corpus of all four conditions; $n_Y$, $n_O$ are the numbers of descriptions containing the word under each persona. The older persona is written in the register of reminiscence, the young one in the register of excitement.}
\label{tab:logodds}
\begin{tabular}{lrrrlrrr}
\toprule
\rowcolor{titlebar}
older word & $n_Y$ & $n_O$ & $z$ & young word & $n_Y$ & $n_O$ & $z$ \\
\midrule
nostalgia & 154 & 1280 & 14.4 & kinda & 347 & 0 & -11.0 \\
reminds & 35 & 880 & 12.9 & cute & 439 & 43 & -9.2 \\
memories & 25 & 702 & 12.6 & vibe & 250 & 1 & -9.1 \\
reminiscent & 44 & 763 & 12.5 & excitement & 1148 & 466 & -8.7 \\
reminding & 19 & 653 & 12.3 & really & 313 & 38 & -8.5 \\
older & 47 & 661 & 12.0 & intense & 893 & 317 & -8.4 \\
quiet & 1128 & 2833 & 11.2 & like & 1086 & 545 & -8.2 \\
brings & 8 & 516 & 10.9 & adrenaline & 261 & 14 & -7.8 \\
nostalgic & 267 & 1040 & 10.1 & romantic & 414 & 73 & -7.4 \\
life & 77 & 630 & 9.9 & excited & 245 & 33 & -6.9 \\
past & 23 & 379 & 9.3 & fun & 228 & 33 & -6.5 \\
concern & 343 & 1172 & 8.9 & friends & 289 & 76 & -6.3 \\
gentle & 327 & 1098 & 8.7 & relaxed & 413 & 136 & -6.0 \\
safety & 22 & 370 & 8.5 & creating & 178 & 9 & -5.8 \\
years & 0 & 243 & 8.2 & chill & 99 & 2 & -5.7 \\
later & 0 & 225 & 8.1 & attraction & 190 & 16 & -5.6 \\
grandchildren & 0 & 194 & 7.6 & super & 86 & 0 & -5.5 \\
fond & 6 & 218 & 7.5 & curiosity & 485 & 219 & -5.3 \\
times & 21 & 281 & 7.4 & rush & 116 & 9 & -5.2 \\
worry & 51 & 358 & 7.3 & cozy & 246 & 73 & -5.1 \\
\bottomrule
\end{tabular}
\end{table}

\subsubsection{Structure of the description embeddings}

A t-SNE of TF-IDF description embeddings is organised first by the emotion label that each model assigned; within it the older-persona texts form sub-clusters that coincide with the memory lexicon. The share of a description's ten nearest neighbours (in the 50-dimensional embedding) that come from its own condition is $\eFigPurityOlder\%$ for older, $\eFigPurityMiddle\%$ for middle, $\eFigPurityYoung\%$ for young and $\eFigPurityBaseline\%$ for the baseline, against 25\% chance. Placing the 68 model$\times$condition cells by their description centroids shows that model identity dominates the text channel as it does the numeric one: the nearest neighbour is the same model for $\eFigTextNNsame\%$ of cells.

\subsubsection{Associations between descriptions and rating changes}

Regressing each model's older-minus-young valence on the image covariates of Table~\ref{tab:predictors} plus the change in each text feature (Tables~\ref{tab:text_shift_valence} and~\ref{tab:text_shift_arousal}), the arrival of memory language raises the older rating (median $\beta=\eTextBetaMemoryMedian$, positive and significant for $\eTextBetaMemoryPos$ of 17 models), vulnerability language lowers it ($\eTextBetaVulnerabilityMedian$, negative for $\eTextBetaVulnerabilityNeg$), and retained excitement raises it ($\eTextBetaExcitementMedian$, positive for all $\eTextBetaExcitementPos$). Adding the text features raises $R^2$ by a median of $\eTextRtwoGainMedian$ and up to $\eTextRtwoGainMax$. These are associations within one generation, not causal effects.

\begin{table}[tbp]
\centering
\small
\caption{Does the description predict the numeric shift on the same image (valence)? Per model, the older-minus-young rating is regressed on the image covariates of Table~\ref{tab:predictors} (valence, valence$^2$, sensitive flag) plus the older-minus-young change in each text feature. Coefficients are rating points per unit change of the flag (or per SD for length). $R^2_{\text{img}}$ uses image covariates only; $R^2_{+\text{text}}$ adds the text features.}
\label{tab:text_shift_valence}
\begin{tabular}{lrrrrrrr}
\toprule
\rowcolor{titlebar}
Model & memory & age ref. & vulnerab. & excitement & $\Delta$length & $R^2_{\text{img}}$ & $R^2_{+\text{text}}$ \\
\midrule
Claude-Opus-5 & +0.14 & -0.18 & \ns{-0.06} & +0.22 & \ns{-0.01} & 0.16 & 0.21 \\
Claude-Sonnet-4.6 & +0.14 & -0.13 & -0.20 & +0.40 & -0.05 & 0.11 & 0.23 \\
Gemini-3-Flash & +0.10 & -0.12 & -0.26 & +0.55 & \ns{+0.02} & 0.12 & 0.24 \\
Gemini-3.8-Flash & \ns{+0.07} & +0.08 & -0.14 & +0.43 & \ns{+0.03} & 0.07 & 0.16 \\
Gemma-3-27B & +0.41 & -0.21 & -0.26 & +0.39 & -0.15 & 0.09 & 0.25 \\
Gemma-4-31B & +0.14 & +0.10 & \ns{-0.12} & +0.54 & -0.05 & 0.10 & 0.19 \\
GLM-4.5V & +0.36 & +0.11 & -0.45 & +1.00 & \ns{-0.04} & 0.03 & 0.27 \\
GPT-5-mini & +0.24 & -0.14 & -0.20 & +0.66 & \ns{-0.01} & 0.07 & 0.20 \\
GPT-5.2 & +0.33 & -0.31 & -0.15 & +0.68 & \ns{-0.03} & 0.10 & 0.23 \\
GPT-5.6-Luna-Pro & +0.15 & +0.19 & -0.17 & +0.40 & \ns{+0.01} & 0.05 & 0.12 \\
Grok-4.6 & +0.32 & +0.09 & -0.23 & +0.76 & -0.07 & 0.10 & 0.30 \\
Llama-4-Scout & \ns{+0.08} & -0.33 & \ns{+0.04} & +0.30 & \ns{+0.01} & 0.04 & 0.05 \\
Qwen3-VL-235B & +0.13 & -0.05 & -0.22 & +0.30 & \ns{-0.03} & 0.13 & 0.18 \\
Qwen3-VL-32B & +0.29 & -0.00 & -0.33 & +0.39 & -0.09 & 0.15 & 0.31 \\
Qwen3-VL-8B & \ns{-0.01} & -0.05 & +0.07 & +0.07 & \ns{-0.01} & 0.05 & 0.06 \\
Qwen3-VL-8B-Think & +0.23 & -0.37 & \ns{-0.02} & +0.44 & \ns{-0.03} & 0.07 & 0.11 \\
Qwen3.8-27B & \ns{+0.07} & +0.06 & -0.12 & +0.34 & \ns{+0.00} & 0.02 & 0.08 \\
\bottomrule
\end{tabular}
\\[2pt]
\begin{minipage}{\linewidth}\footnotesize HC1 standard errors. Entries in \textcolor{black!45}{grey} are not significant ($q\ge.05$, Benjamini--Hochberg across models within column); all others have $q<.05$.\end{minipage}
\end{table}

\begin{table}[tbp]
\centering
\small
\caption{Does the description predict the numeric shift on the same image (arousal)? Per model, the older-minus-young rating is regressed on the image covariates of Table~\ref{tab:predictors} (valence, valence$^2$, sensitive flag) plus the older-minus-young change in each text feature. Coefficients are rating points per unit change of the flag (or per SD for length). $R^2_{\text{img}}$ uses image covariates only; $R^2_{+\text{text}}$ adds the text features.}
\label{tab:text_shift_arousal}
\begin{tabular}{lrrrrrrr}
\toprule
\rowcolor{titlebar}
Model & memory & age ref. & vulnerab. & excitement & $\Delta$length & $R^2_{\text{img}}$ & $R^2_{+\text{text}}$ \\
\midrule
Claude-Opus-5 & +0.03 & -0.12 & \ns{+0.01} & +0.19 & -0.01 & 0.20 & 0.25 \\
Claude-Sonnet-4.6 & +0.01 & -0.08 & \ns{-0.05} & +0.21 & +0.02 & 0.14 & 0.18 \\
Gemini-3-Flash & -0.06 & -0.13 & +0.22 & +0.26 & -0.02 & 0.19 & 0.23 \\
Gemini-3.8-Flash & -0.03 & +0.09 & \ns{+0.04} & +0.20 & -0.04 & 0.06 & 0.09 \\
Gemma-3-27B & +0.06 & -0.23 & -0.17 & +0.24 & -0.06 & 0.13 & 0.19 \\
Gemma-4-31B & +0.00 & -0.07 & \ns{-0.03} & +0.49 & +0.01 & 0.06 & 0.12 \\
GLM-4.5V & +0.03 & -0.09 & \ns{+0.16} & +0.51 & +0.01 & 0.07 & 0.11 \\
GPT-5-mini & -0.06 & -0.04 & \ns{+0.02} & +0.33 & -0.02 & 0.07 & 0.11 \\
GPT-5.2 & -0.00 & -0.00 & \ns{-0.00} & +0.18 & +0.01 & 0.05 & 0.07 \\
GPT-5.6-Luna-Pro & +0.08 & +0.38 & \ns{+0.06} & +0.26 & -0.01 & 0.05 & 0.08 \\
Grok-4.6 & -0.02 & +0.12 & \ns{+0.10} & +0.30 & -0.00 & 0.13 & 0.17 \\
Llama-4-Scout & -0.15 & -0.20 & \ns{+0.04} & \ns{+0.16} & +0.04 & 0.01 & 0.02 \\
Qwen3-VL-235B & -0.06 & +0.00 & \ns{+0.04} & +0.22 & +0.01 & 0.06 & 0.09 \\
Qwen3-VL-32B & +0.03 & +0.01 & \ns{-0.06} & +0.31 & -0.01 & 0.05 & 0.09 \\
Qwen3-VL-8B & -0.10 & -0.02 & \ns{+0.02} & +0.17 & +0.01 & 0.04 & 0.05 \\
Qwen3-VL-8B-Think & -0.09 & -0.56 & \ns{+0.15} & +0.33 & -0.03 & 0.01 & 0.02 \\
Qwen3.8-27B & -0.07 & -0.02 & \ns{-0.01} & +0.39 & -0.01 & 0.02 & 0.06 \\
\bottomrule
\end{tabular}
\\[2pt]
\begin{minipage}{\linewidth}\footnotesize HC1 standard errors. Entries in \textcolor{black!45}{grey} are not significant ($q\ge.05$, Benjamini--Hochberg across models within column); all others have $q<.05$.\end{minipage}
\end{table}

\subsection{Model-level associations and within-developer comparisons}

\label{app:model-comparisons}

\subsubsection{Model properties and persona shifts}

We examine model-level correlates ($n=17$, exploratory; Table~\ref{tab:model_correlates}). The size of a model's valence shift is unrelated to baseline competence ($\rho=\eCorRbaseDelta$, $p=\eCorRbaseDeltaP$) and over-dispersion ($\rho=\eCorSdDelta$, $p=\eCorSdDeltaP$), but strongly related to its memory rate ($\rho=\eCorMemDelta$), description lengthening ($\rho=\eCorLenDelta$) and age references ($\rho=\eCorAgeRefDelta$).

\begin{table}[tbp]
\centering
\small
\caption{Across the 17 models, Spearman correlations between properties of a model and its response to the age persona ($n=17$ models; exploratory, Benjamini--Hochberg within column). The size of the persona shift is unrelated to baseline competence and to over-dispersion, but tracks how much the model writes about memory and about its own age under the older persona.}
\label{tab:model_correlates}
\begin{tabular}{lrrrr}
\toprule
\rowcolor{titlebar}
Model-level predictor & $\Delta_{O-Y}$ valence & $\Delta_{O-Y}$ arousal & aversive net & $r_{\text{part}}$ valence \\
\midrule
over-dispersion (SD ratio, valence) & \ns{-0.09} & -0.51 & \ns{-0.08} & -0.06 \\
baseline $r$ (valence) & \ns{-0.07} & 0.09 & \ns{0.41} & 0.48 \\
baseline MAE (valence) & \ns{-0.26} & -0.40 & \ns{-0.04} & -0.26 \\
description length (young) & \ns{-0.50} & 0.07 & \ns{0.50} & 0.35 \\
$\Delta$ description length (O$-$Y) & -0.73 & -0.12 & 0.61 & 0.30 \\
memory lexicon rate (older) & -0.76 & -0.44 & 0.67 & 0.33 \\
explicit age reference (older) & -0.65 & -0.39 & \ns{0.53} & 0.20 \\
first-person rate (older) & -0.60 & -0.27 & \ns{0.37} & 0.16 \\
run cost (USD, price proxy) & \ns{0.11} & 0.17 & \ns{0.19} & 0.22 \\
\bottomrule
\end{tabular}
\\[2pt]
\begin{minipage}{\linewidth}\footnotesize Two-sided tests; entries in \textcolor{black!45}{grey} are not significant ($q\ge.05$), all others have $q<.05$.\end{minipage}
\end{table}

\subsubsection{Scale, reasoning, generation, and model tier}

\begin{figure}[tbp]
  \centering
  \includegraphics[width=0.9 \linewidth]{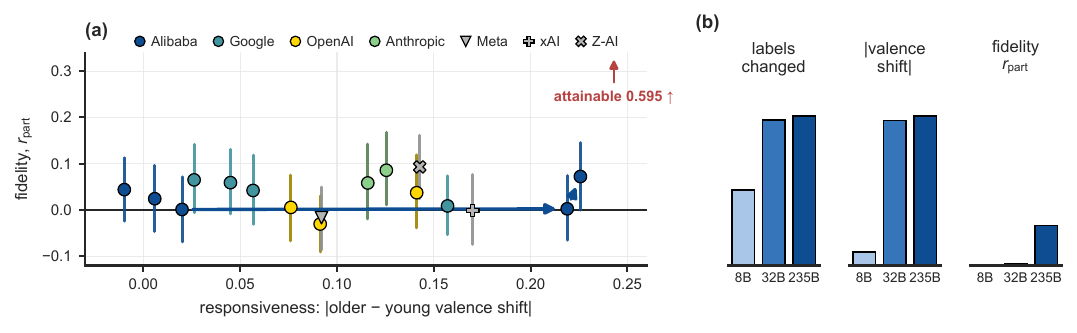}
\caption{\textbf{A stronger age response does not improve fidelity.}
(a)~Across 17 models, responsiveness and human fidelity are unrelated ($r=0.04$). The Qwen3 VL ladder (8B$\rightarrow$32B$\rightarrow$235B) moves mainly along the responsiveness axis, while the attainable fidelity ceiling is $0.595$.
(b)~Within Qwen3 VL, emotion label changes increase ($105\rightarrow202\rightarrow208$) and valence shifts grow ($0.02\rightarrow0.22\rightarrow0.23$), but fidelity remains low ($0.00\rightarrow0.00\rightarrow0.07$).}
\label{fig:scale}
\end{figure}

We see the same pattern in the controlled comparisons (Table~\ref{tab:pairs}): the Qwen3-VL step from 8B to 32B takes the memory rate from $\eLadderMemEight\%$ to $\eLadderMemThirtyTwo\%$, and the Gemma-3-27B to Gemma-4-31B step moves it from $\ePairGemmaOldMem\%$ to $\ePairGemmaNewMem\%$ and the shift from $\ePairGemmaOldDelta$ to $\ePairGemmaNewDelta$. Qwen3-VL-8B-Thinking has no valence shift ($\eThinkDeltaThink$ against $\eThinkDeltaBase$) but a larger arousal shift ($\eThinkAroThink$ against $\eThinkAroBase$), and is the one model that gains appetitive labels under the older persona.

\begin{table}[tbp]
\centering
\scriptsize
\setlength{\tabcolsep}{3pt}
\caption{Controlled comparisons within a developer. Each group holds the developer fixed and varies one thing: parameter count (the Qwen3-VL ladder), inference-time reasoning at a fixed 8B (Thinking), model generation, or product tier. Columns: baseline $r$ and SD ratio (valence); the older-minus-young shift on valence and arousal; the partial correlation with the human age effect; the share of images the persona does not move; labels changed and net aversive labels gained; and the share of older-persona descriptions containing memory language or an explicit age reference. The step that most increases the persona shift (Qwen3-VL 8B to 32B) is also the step that most increases memory and age-reference language; newer generations and higher tiers (Gemma-4, GPT-5.2 and 5.6, Claude-Opus-5) write less in that register and, in most cases, shift less. No step moves the human alignment out of the noise.}
\label{tab:pairs}
\resizebox{\linewidth}{!}{
\begin{tabular}{llrrrrrrrrrr}
\toprule
\rowcolor{titlebar}
Group & Model & $r_{\text{base}}$ & SD ratio & $\Delta^{V}_{O-Y}$ & $\Delta^{A}_{O-Y}$ & $r_{\text{part}}$ & flat & changed & avers. & memory\,\% & age ref.\,\% \\
\midrule
Qwen3-VL ladder & Qwen3-VL-8B & 0.885 & 1.34 & -0.020 & -0.296 & 0.001 & 73\% & 105 & -9 & 7.4 & 4.2 \\
 & Qwen3-VL-32B & 0.866 & 1.43 & -0.219 & -0.295 & 0.002 & 0\% & 202 & +57 & 63.8 & 30.7 \\
 & Qwen3-VL-235B & 0.902 & 1.31 & -0.226 & -0.245 & 0.072 & 0\% & 208 & +53 & 72.0 & 37.3 \\
\addlinespace
Reasoning at 8B & Qwen3-VL-8B & 0.885 & 1.34 & -0.020 & -0.296 & 0.001 & 73\% & 105 & -9 & 7.4 & 4.2 \\
 & Qwen3-VL-8B-Think & 0.837 & 1.66 & +0.010 & -0.441 & 0.044 & 14\% & 175 & +4 & 21.0 & 0.9 \\
\addlinespace
Qwen generation & Qwen3-VL-32B & 0.866 & 1.43 & -0.219 & -0.295 & 0.002 & 0\% & 202 & +57 & 63.8 & 30.7 \\
 & Qwen3.8-27B & 0.855 & 1.22 & -0.006 & -0.157 & 0.024 & 30\% & 156 & -9 & 15.3 & 1.1 \\
\addlinespace
Gemma generation & Gemma-3-27B & 0.863 & 1.40 & -0.157 & -0.402 & 0.009 & 1\% & 195 & +33 & 88.0 & 6.9 \\
 & Gemma-4-31B & 0.908 & 1.35 & -0.057 & -0.429 & 0.042 & 16\% & 197 & +39 & 22.7 & 2.2 \\
\addlinespace
Gemini generation & Gemini-3-Flash & 0.912 & 1.36 & -0.026 & -0.395 & 0.065 & 21\% & 150 & +31 & 20.8 & 1.5 \\
 & Gemini-3.8-Flash & 0.919 & 1.20 & -0.045 & -0.275 & 0.059 & 9\% & 150 & +30 & 12.4 & 1.1 \\
\addlinespace
OpenAI tier & GPT-5-mini & 0.877 & 1.50 & -0.141 & -0.518 & 0.037 & 4\% & 155 & +28 & 44.8 & 11.0 \\
 & GPT-5.2 & 0.891 & 1.38 & -0.091 & -0.231 & -0.030 & 7\% & 175 & +43 & 17.4 & 2.2 \\
 & GPT-5.6-Luna-Pro & 0.890 & 1.51 & -0.076 & -0.193 & 0.006 & 0\% & 122 & +17 & 4.7 & 0.1 \\
\addlinespace
Anthropic tier & Claude-Sonnet-4.6 & 0.900 & 1.36 & -0.126 & -0.272 & 0.086 & 0\% & 206 & +52 & 53.9 & 10.6 \\
 & Claude-Opus-5 & 0.923 & 1.28 & -0.116 & -0.208 & 0.058 & 11\% & 140 & +43 & 21.1 & 1.3 \\
\addlinespace
\bottomrule
\end{tabular}}
\end{table}

\begin{table}[tbp]
\centering
\small
\caption{\textbf{Qwen3-VL scale ladder, the only comparison in which developer, generation and instruction-tuning recipe are held fixed.} Cross-developer differences elsewhere confound parameter count with data and alignment procedure and are not read as scale effects.}
\label{tab:ladder}
\resizebox{\linewidth}{!}{
\begin{tabular}{llrrrrr}
\toprule
\rowcolor{titlebar!90}
Model & Scale (Model Parameters) & Dimension & $r_{\text{part}}$ & Flat & $\Delta_{O-Y}$ & Labels changed \\
\midrule
Qwen3-VL-8B & 8B & arousal & -0.108 & 0\% & -0.296 & 105 \\
Qwen3-VL-32B & 32B & arousal & -0.190 & 0\% & -0.295 & 202 \\
Qwen3-VL-235B & 235B & arousal & -0.067 & 2\% & -0.245 & 208 \\
Qwen3-VL-8B & 8B & valence & 0.001 & 73\% & -0.020 & 105 \\
Qwen3-VL-32B & 32B & valence & 0.002 & 0\% & -0.219 & 202 \\
Qwen3-VL-235B & 235B & valence & 0.072 & 0\% & -0.226 & 208 \\
\bottomrule
\end{tabular}}
\end{table}

\addtocontents{toc}{\protect\apxpart{Part III}{Mitigation and Robustness}}
\FloatBarrierAbove
\section{Agent-Based Mitigation Study}
\label{app:mitigation}

\subsection{Design and arms}
\label{app:design}
We test which ingredient, if any, helps an age persona track human age differences. All arms
use GPT-5.6-Luna-Pro, rate the same images under the young, middle and older personas, and return the
four fields of the main protocol (valence, arousal, description, emotion), so they are scored with the
estimators in Appendix~\ref{app:estimands}. The direct prompt (S0) reuses the main-study responses. Each
further arm adds one ingredient (Table~\ref{tab:arms}).

\begin{table}[tbp]
\centering
\small
\caption{Mitigation arms and the ingredient each isolates.}
\label{tab:arms}
\resizebox{\linewidth}{!}{%
\begin{tabular}{llccccc}
\toprule
\rowcolor{titlebar}
Arm & Type & Instruction & Critic loop & Human evidence & Panels + anchored calibration & Verified rule \\
\midrule
S0 direct prompt        & baseline  & & & & & \\
S1 debias instruction   & reasoning & \checkmark & & & & \\
S2 deliberation agent   & reasoning & & \checkmark & & & \\
S3 grounded agent       & grounded  & & \checkmark & \checkmark & & \\
S4 contrastive panel    & grounded  & & audit & \checkmark & \checkmark & \\
S5 VERA                 & grounded  & & & \checkmark & anchored only & \checkmark \\
VERA-Cal                & grounded  & & audit & \checkmark & \checkmark & \checkmark \\
\bottomrule
\end{tabular}}
\end{table}

\subsubsection{Balanced subset}

\label{app:subset}
The subset contains exactly 75 images per OASIS list, allocated within each list in proportion to image
category, valence class (unpleasant, neutral, pleasant) and the sensitive flag. Among 200 stratified
draws we kept the one whose human distributions best matched the remaining images, a criterion that
uses no model output and therefore cannot favour any arm (Table~\ref{tab:mitigation_subset}). On the subset, the
main-study results of the 17 VLMs are reproduced (correlation across models $0.94$ for the
older-minus-young shift and $0.76$ for $r_{\text{part}}$).

\begin{table}[tbp]
\centering
\small
\caption{The balanced subset against the full stimulus set.}
\label{tab:mitigation_subset}
\begin{tabular}{lccc}
\toprule
\rowcolor{titlebar}
 & Subset (300) & All (900) & KS $p$ \\
\midrule
Person / Scene / Object / Animal & 38 / 25 / 22 / 15\% & 38 / 24 / 22 / 15\% & -- \\
Pleasant / neutral / unpleasant  & 51 / 24 / 25\% & 51 / 24 / 25\% & -- \\
Sensitive-flagged                & 17\% & 19\% & -- \\
Human valence, mean (SD)         & 4.33 (1.21) & 4.33 (1.23) & 0.95 \\
Human arousal, mean (SD)         & 3.64 (0.84) & 3.67 (0.84) & 0.93 \\
Human age effect (valence), SD   & 0.107 & 0.106 & 0.97 \\
Human age effect (arousal), SD   & 0.057 & 0.056 & 0.97 \\
\bottomrule
\end{tabular}
\end{table}

\subsection{Agent components}

\label{app:agents}

\subsubsection{Perception agent}

Describes the image without any persona, returning its content and the elements a viewer is likely
to react to emotionally. Keeping this step persona-free separates perception from age conditioning.
The output is cached and shared by all agentic arms, so they reason over an identical description.

\subsubsection{Persona rater}

Receives the image, the perception output and the persona sentence of the main study, and returns a
rationale followed by the four rating fields. Placing the rationale first exposes the reasoning that
the critic evaluates.

\subsubsection{Stereotype critic and reviser (S2, S3)}

The critic flags drafts that rely on assumptions about the age group (frailty, fear of risk,
nostalgia, reduced enjoyment, disapproval) rather than on the image; flagged drafts are revised once,
and unparseable revisions fall back to the draft.

\subsubsection{Human-evidence retriever (S3--S5, VERA-Cal)}

Returns the images most similar to the target by cosine similarity of sentence embeddings
(all-mpnet-base-v2) of the model's own no-age descriptions, restricted to the three OASIS lists other
than the target's and excluding images of the same theme. For each neighbour it reports pooled human
valence and arousal and the estimated older-minus-young difference, obtained by scaling the human age
slope $\Delta H_i$ from $+1$\,SD of rater age to the distance between the band midpoints.

\subsubsection{Rater panel (S4, VERA-Cal)}

For each age band, simulates five deliberately different raters (gender, background, temperament),
each rating the image independently with the evidence in view. Panel means and spreads are passed on.

\subsubsection{Anchored calibrator (S4, VERA-Cal)}

Starting from the model's own no-age rating from the main study, outputs the three persona ratings,
sizing the age differences to the evidence rather than copying the panels, which tend to exaggerate
them.

\subsubsection{Auditor (S4, VERA-Cal)}

Flags caricature (age differences outside the range of the evidence) and stereotyped descriptions;
flagged outputs are recalibrated once.

\subsubsection{Hypothesiser and code verifier (S5, VERA-Cal)}

From three retrieved cases, the hypothesiser states one general, testable rule about how age changes
responses to this kind of content and predicts the target's older-minus-young difference. In a
text-only call the rule is applied to four further cases whose human answers are withheld from the
prompt, and the predictions are scored \emph{in code} against the human differences (same sign, or
both below $0.05$ and $0.1$ in magnitude). A rule agreeing on fewer than three of four cases is revised
once from the per-case errors and re-tested on four \emph{fresh} cases the agent has never seen; if it
fails again, the agent abstains and predicts no age difference. No case is ever used to grade a rule
that was fitted to it.

\subsection{Execution and validation}

\label{app:agent-execution}

\subsubsection{Leakage controls}

\label{app:leakage}
Evidence never includes the target image, any image from its OASIS list, or an image of the same theme;
we verified this for every image (0 violations). The verifier's test prompts contain no human answers,
refinement uses test cases only, and re-verification uses unseen cases. The anchor is the model's own
no-age rating, not a human value.

\subsubsection{Execution, caching, and cost}

\label{app:mitig_impl}
All calls use \texttt{openai/gpt-5.6-luna-pro} through OpenRouter with temperature $0$, one completion
and a 1{,}500-token budget, matching the main protocol. Every call is cached by model, prompt text and
image, and every intermediate output (perception, rationale, critique, neighbours, panels, rules and
verification scores) is stored with the final rating. The six arms produced 99.7--100\% valid ratings
on the 300 images, for a total cost of \$24.21.

\subsubsection{Pipeline diagrams}

\label{app:pipeline}

\begin{figure}[tbp]
\centering
\includegraphics[width=\linewidth]{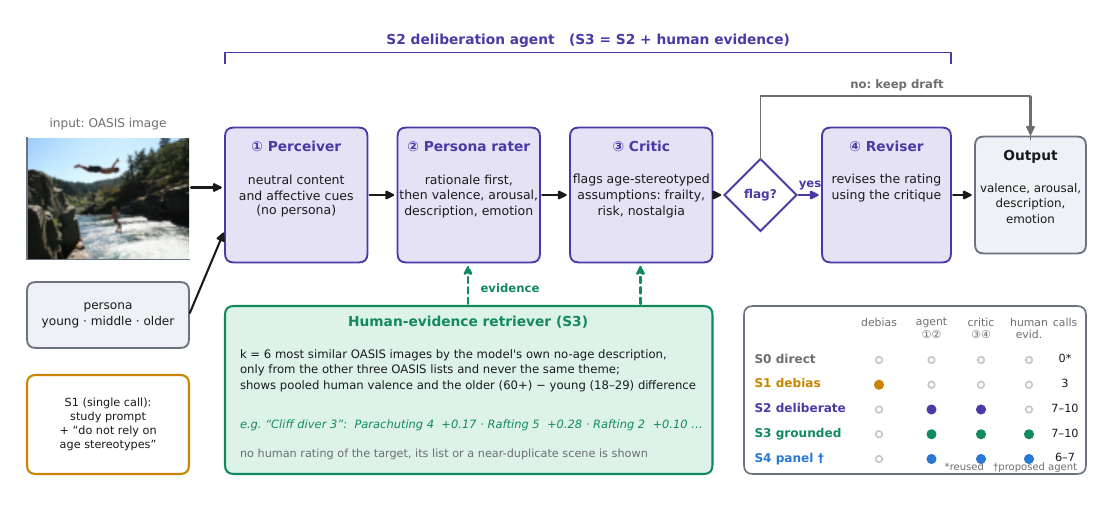}
\caption{Reasoning and grounded agents (S1--S3) and the components of every arm.}
\label{fig:pipeline}
\end{figure}

\begin{figure}[tbp]
\centering
\includegraphics[width=\linewidth]{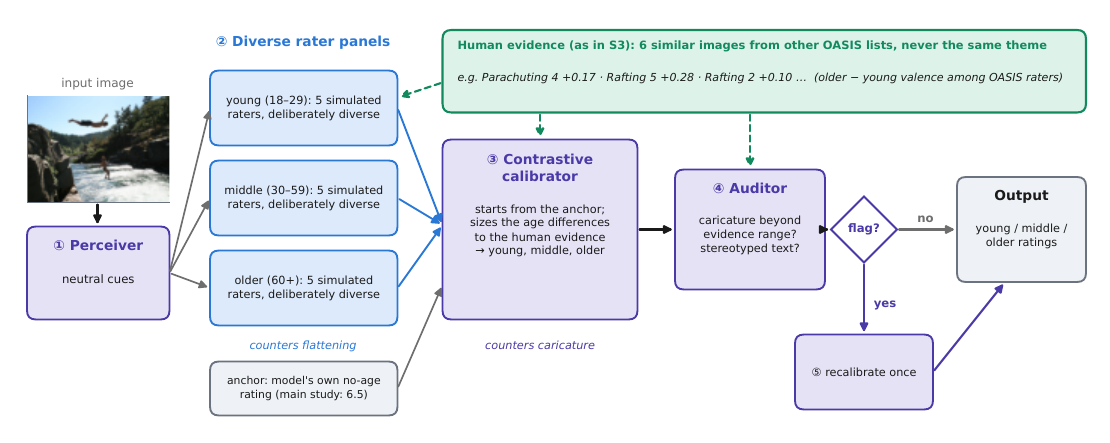}
\caption{The grounded contrastive panel (S4): diverse rater panels per age band, a calibrator anchored
on the model's own no-age rating, and an auditor for caricature and stereotyped text, all with human
evidence from the other OASIS lists.}
\label{fig:pipeline_s4}
\end{figure}

\begin{figure}[tbp]
\centering
\includegraphics[width=\linewidth]{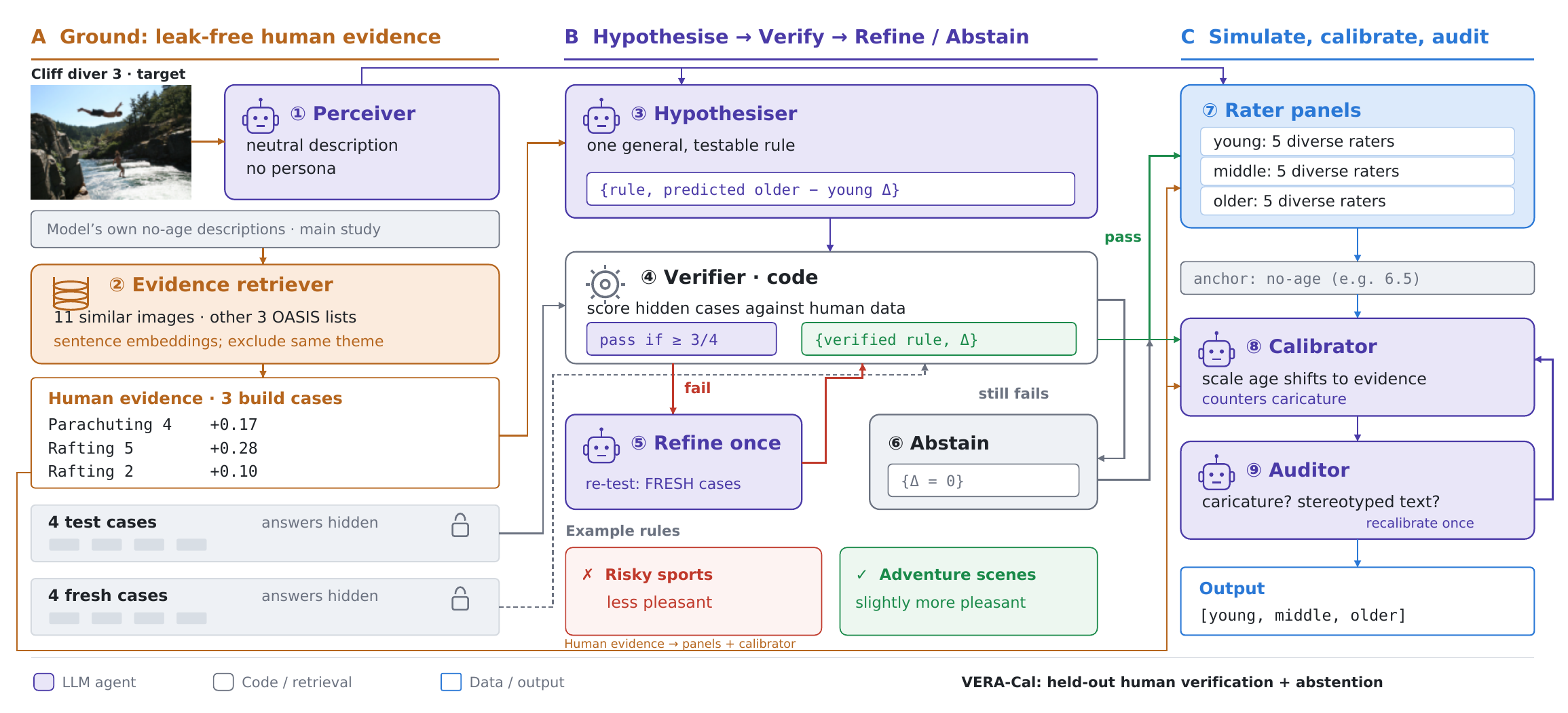}
\caption{\textbf{VERA-Cal.} (A)~Human evidence is retrieved from similar images in the other three
OASIS lists, never of the same theme, using sentence embeddings of the model's own no-age
descriptions, and split into build, test and fresh cases. (B)~A hypothesiser states one testable rule
about age differences from the build cases; a verifier implemented in code scores it on test cases
whose human answers are hidden (pass: $\geq 3/4$); a failing rule is refined once and re-tested on
fresh cases, otherwise the agent abstains ($\Delta = 0$). (C)~Diverse rater panels, a calibrator
anchored on the model's own no-age rating and an auditor produce the three persona ratings; the
human evidence and the verified rule are passed to both. Example rules and values are illustrative.}
\label{fig:pipeline_veracal}
\end{figure}

\begin{table}[tbp]
\centering
\small
\caption{Alignment with human age differences on the same 300 images for all systems. $r_{\text{part}}$ measures image level fidelity after controlling for pleasantness. Values for system groups are means with 95\% joint image bootstrap confidence intervals. The attainable ceiling is estimated from the reliability of the human age effects. Direction agreement can approach or exceed this ceiling because it is strongly driven by pleasantness and should not be read as human level image specific fidelity.}
\label{tab:mitig_alignment}
\resizebox{\linewidth}{!}{
\begin{tabular}{lccc}
\toprule
\rowcolor{titlebar}
Level & $r_{\text{part}}$ [95\% CI] & \% of ceiling & Direction agreement \\
\midrule
17 VLMs, direct prompt
& 0.06 [$-$0.04, 0.15]
& 9\%
& 51\% \\

Reasoning agents (S1, S2)
& $-$0.01 [$-$0.09, 0.07]
& $-$1\%
& 42\% \\

Human grounded agents (S3 to S5, VERA Cal)
& 0.16 [0.07, 0.25]
& 27\%
& 83\% \\

\quad Best: contrastive panel (S4)$^{\dagger}$
& 0.27 [0.16, 0.37]
& 44\%
& 88\% \\

Attainable ceiling
& 0.60
& 100\%
& 85\% \\
\midrule
\multicolumn{4}{l}{\emph{Bootstrap tests of group mean differences}} \\

Grounded $-$ VLMs
& $+$0.11 [$-$0.03, 0.25], $p=.13$
& & $+$33 pts, $p<.001$ \\

Grounded $-$ reasoning
& $+$0.17 [0.04, 0.30], $p=.011$
& & $+$41 pts, $p<.001$ \\

Reasoning $-$ VLMs
& $-$0.06 [$-$0.16, 0.03], $p=.20$
& & $-$8 pts, $p=.006$ \\

Ceiling $-$ every level
& all $p<.001$
& & \\
\bottomrule
\end{tabular}
}
\vspace{2pt}
\begin{minipage}{\linewidth}
\footnotesize
$^{\dagger}$S4 was selected post hoc. Compared with the best VLM, Qwen3 VL 235B,
$\Delta r_{\text{part}}=+0.13$ [$-$0.02, 0.29], $p=.09$.
\end{minipage}
\end{table}

\subsection{Evaluation and results}

\label{app:agent-results}

\subsubsection{Attainable ceiling}

\label{app:ceiling}
The human age effects $\Delta H_i$ are posterior means, so their reliability follows from the
shrinkage identity, $\rho=\Var(\Delta H)/(\Var(\Delta H)+\overline{\tau^2})$, where $\tau_i$ is the
posterior SD. For $r_{\text{part}}$ the relevant quantity is the age effect after the quadratic in
pooled valence is partialled out; with residual $e_i$,
$\rho_{\text{part}}=\Var(e)/(\Var(e)+\overline{\tau^2})$ and the attainable correlation is
$\sqrt{\rho_{\text{part}}}=0.60$ (0.86 for the raw age effect). The attainable direction agreement is
$\frac{1}{n}\sum_i\Phi(|\Delta H_i|/\tau_i)=85\%$, the rate at which the true effect matches the sign of
its estimate.

\subsubsection{Results and paired statistical tests}

\label{app:mitig_stats}
\begin{figure}[tbp]
\centering
\includegraphics[width=\linewidth]{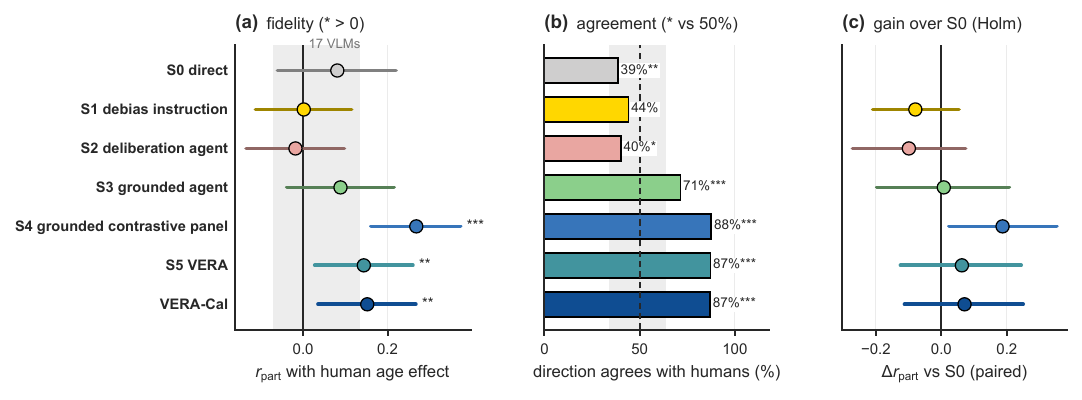}
\caption{Human alignment of each arm (300 images). (a)~$r_{\text{part}}$ with 95\% CI; stars mark
$r_{\text{part}}>0$; the shaded band is the range of the 17 VLMs. (b)~Direction agreement; stars mark a
binomial difference from 50\% (S0 and S2 are significantly \emph{below} chance). (c)~Paired gain in
$r_{\text{part}}$ over S0; none survives Holm correction.}
\label{fig:mitig_stats}
\end{figure}

We report every arm against the direct prompt below, with paired tests on the
same images: a paired bootstrap for $r_{\text{part}}$, an exact McNemar test on per-image direction
correctness and a Wilcoxon signed-rank test on per-image absolute error, Holm-corrected across the six
arms (Figure~\ref{fig:mitig_stats}). Grounded arms fix the direction of many more images than they
break (e.g.\ 211 vs.\ 15 for S4) and halve the error against the human change (0.48 to 0.20;
$p<.001$), whereas S1 and S2 change neither significantly. The advantage of S4 in $r_{\text{part}}$
(+0.19) is significant before but not after correction (Holm $p=.14$).

\begin{table}[t]
\centering
\scriptsize
\setlength{\tabcolsep}{3pt}
\caption{Human alignment of the mitigation arms (GPT-5.6-Luna-Pro, 300 images). $r_{\text{part}}$: fidelity to the human age effect with pleasantness partialled out (95\% image bootstrap CI). \emph{Agree}: direction agreement on images the model moves (binomial test vs.\ 50\%). $\Delta r_{\text{part}}$: paired bootstrap difference from the direct prompt. McNemar: images whose direction becomes correct / incorrect relative to the direct prompt. MAE: mean absolute error of the older-minus-young change. $p$-values Holm-corrected across the six arms; for \emph{Agree} and McNemar, entries in \textcolor{black!45}{grey} are not significant ($p\ge.05$), all others have $p<.05$.}
\label{tab:mitigation_stats}
\begin{tabular}{lcrrcrrr}
\toprule
\rowcolor{titlebar}
System & $r_{\text{part}}$ [95\% CI] & Pearson & Agree & $\Delta r_{\text{part}}$ [95\% CI] & McNemar & MAE & Wilcoxon $p$ \\
\midrule
S0 direct & 0.08 [-0.06, 0.22] & -0.11 & 39\% & -- & -- & 0.48 & -- \\
S1 debias instruction & 0.00 [-0.11, 0.11] & -0.03 & \ns{44\%} & -0.08 [-0.21, +0.05] & \ns{45/27} & 0.47 & .604 \\
S2 deliberation agent & -0.02 [-0.13, 0.10] & -0.22 & 40\% & -0.10 [-0.27, +0.07] & \ns{51/36} & 0.54 & .248 \\
S3 grounded agent & 0.09 [-0.04, 0.22] & +0.30 & 71\% & +0.01 [-0.20, +0.21] & 146/27 & 0.37 & $<.001$ \\
S4 grounded contrastive panel & 0.27 [0.16, 0.37] & +0.74 & 88\% & +0.19 [+0.02, +0.35] & 211/15 & 0.20 & $<.001$ \\
S5 VERA & 0.14 [0.03, 0.26] & +0.75 & 87\% & +0.06 [-0.12, +0.24] & 193/14 & 0.20 & $<.001$ \\
VERA-Cal & 0.15 [0.04, 0.27] & +0.75 & 87\% & +0.07 [-0.11, +0.25] & 200/14 & 0.20 & $<.001$ \\
\midrule
17 VLMs (range) & -0.07 to 0.14 & -0.13 to +0.19 & 34--64\% & -- & -- & 0.38--0.62 & -- \\
\bottomrule
\end{tabular}
\end{table}

\subsubsection{Comparison with the 17 VLMs}

\label{app:vsmodels}
\begin{figure}[tbp]
\centering
\includegraphics[width=\linewidth]{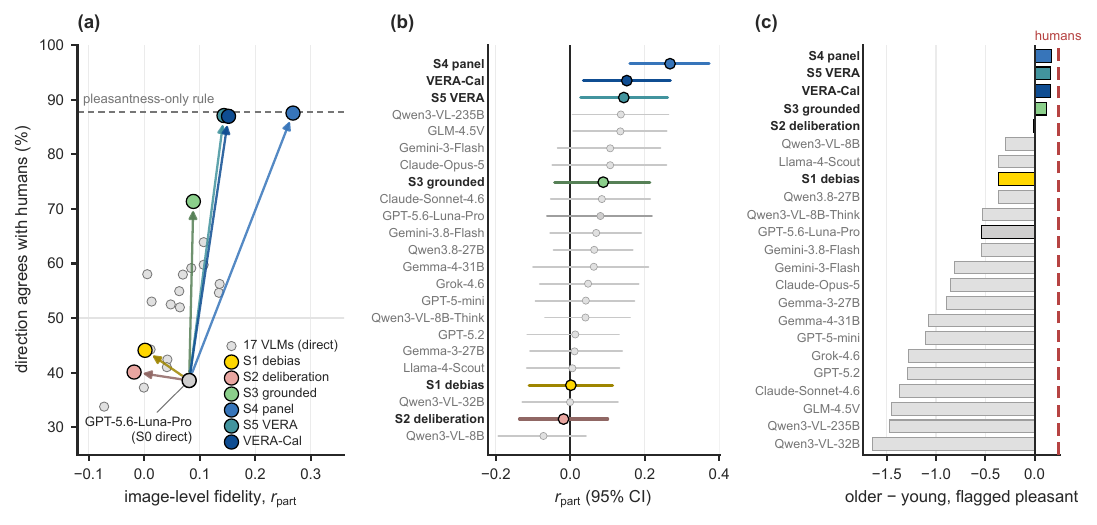}
\caption{Mitigation arms against the 17 VLMs on the same 300 images. (a)~Direction agreement against
$r_{\text{part}}$; arrows start from GPT-5.6-Luna-Pro's direct prompt; the dashed line is a
pleasantness-only rule fitted on the other lists. (b)~$r_{\text{part}}$ for all systems. (c)~Shift on
sensitive-flagged pleasant images.}
\label{fig:vsmodels}
\end{figure}

Figure~\ref{fig:vsmodels} places each arm among the 17 VLMs on the same images. Reasoning arms stay
within the range of existing models. On sensitive-flagged pleasant images, where older OASIS raters are
more positive than young raters ($+0.24$), all 17 VLMs move in the opposite direction ($-0.29$ to
$-1.65$); the debiasing instruction attenuates this ($-0.37$), deliberation neutralises it ($-0.01$), and
only the grounded arms move with humans ($+0.12$ to $+0.17$).

\subsubsection{Sources of the gains}

\label{app:mitig_detail}
\begin{figure}[tbp]
\centering
\includegraphics[width=\linewidth]{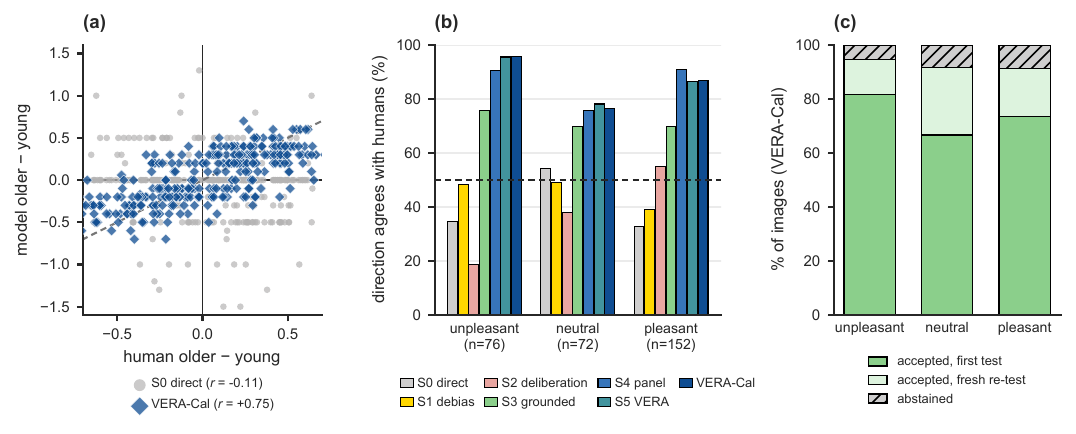}
\caption{(a)~Per image, the model's older-minus-young change against the human estimate for the
direct prompt and VERA-Cal. (b)~Direction agreement by valence class for every arm. (c)~Outcome of
VERA-Cal's rules: accepted on the first test, accepted after a fresh re-test, or abstained.}
\label{fig:mitig_detail}
\end{figure}

Direction agreement alone cannot separate genuine age knowledge from pleasantness: a rule that
predicts the human age effect from pooled valence alone, fitted on the other lists, agrees with humans
on 88\% of images (95.1\% under the moved-image definition of Table~\ref{tab:rb_rule}, which also reports its error). Grounded arms reach the same level, which is why we rely on $r_{\text{part}}$. Their
gains in agreement appear in all three valence classes, not only on pleasant images
(Figure~\ref{fig:mitig_detail}b). The panel's simulated raters are less varied than real raters (panel
SD $0.49$ of the human SD), and the grounded arms' age differences are smaller than the human
estimates on average (caricature ratio $0.75$--$0.97$), so the gain is not driven by exaggeration.

\subsubsection{Verified rules}

\label{app:rules}
Across the 300 images, VERA-Cal's first rule passed the hidden-case test for 222 images (74\%); 78 rules
were revised, of which 55 passed on fresh cases, and the agent abstained on 23 images (8\%).
We list verbatim rule examples below. Explicit rules make the agent's assumptions
inspectable, but compressing the evidence into one rule costs fidelity: VERA and VERA-Cal match the
panel in direction agreement (87\%) yet reach about half its $r_{\text{part}}$ (0.14--0.15 vs.\ 0.27).

\begin{table}[t]
\centering
\caption{Examples of rules stated by the agent, verbatim, with their agreement on hidden human cases (first test; fresh re-test after refinement) and the outcome.}
\label{tab:mitigation_rules}
\resizebox{\linewidth}{!}{
\begin{tabular}{p{1.6cm}p{6cm}rp{4cm}l}
\toprule
\rowcolor{titlebar}
Image & First rule & Pass & Refined rule (fresh pass) & Outcome \\
\midrule
Alcohol 7 & Compared with younger adults, older adults tend to rate alcohol-related scenes more pleasantly when they convey warm social conviviality, but less pleasantly when they are primarily impersonal product displays. & 75\% & -- & accepted \\
Camping 2 & For calm, positive scenes of children engaged together, adults aged 60 and older tend to rate pleasantness modestly higher than adults aged 18 to 29. & 75\% & -- & accepted \\
Camping 4 & For warm, affectionate, and playful interpersonal scenes, older adults tend to rate pleasantness modestly higher than younger adults, by roughly three-tenths of a scale point. & 100\% & -- & accepted \\
Ambulance 1 & For subdued or peaceful scenes, older adults' pleasantness ratings differ only modestly from younger adults' ratings and show no consistent directional shift across content. & 50\% & Older adults' pleasantness ratings shift upward relative to younger adults for benign or comforting scenes but downward for scenes suggesting vulnerability, threat, or wartime distress, with larger differences when those cues are salient. (75\%) & refined \\
Cat 13 & For calm, gentle, and affectionate animal images, older adults tend to give higher pleasantness ratings than younger adults. & 50\% & Older adults tend to rate emotionally positive and affiliative images, especially animals and babies, as more pleasant than younger adults, while neutral images show little or no age difference. (100\%) & refined \\
Cold 7 & The older-minus-younger pleasantness difference is context-dependent, tending to be positive for clearly peaceful, contentment-evoking landscapes but negative for stark snowy scenes emphasizing solitude or melancholy. & 50\% & Older adults generally rate scenes as more pleasant than younger adults, including quiet or solitary snowy scenes, although the size of this positive difference varies substantially by image. (100\%) & refined \\
Cat 6 & For calm, neutral-to-gently endearing faces or animals, older adults tend to give slightly higher pleasantness ratings than young adults. & 25\% & Older adults tend to rate clearly appealing and positive animals as more pleasant than young adults, but show little or lower pleasantness for neutral, ambiguous, sad, bored, or mildly intimidating stimuli. (25\%) & abstained \\
Cat 9 & Compared with younger adults, older adults tend to rate calm, solitary animals with gentle, attentive expressions as more pleasant, while this age advantage is smaller for interaction-focused scenes. & 50\% & Older adults tend to rate clearly warm, affiliative companion-animal scenes as more pleasant than younger adults, but show little or no advantage for merely attentive, ambiguous, or neutral expressions. (50\%) & abstained \\
Cliff diver 1 & For images depicting physically risky or exhilarating acts, older adults tend to report slightly lower pleasantness than younger adults on average, with the difference varying by scene. & 25\% & For physically risky or exhilarating scenes, older adults tend to report slightly higher pleasantness than younger adults, especially for aerial or open-water acts, while responses to climbing scenes may be near zero or slightly lower. (50\%) & abstained \\
\bottomrule
\end{tabular}}
\end{table}

\subsection{Limitations}

\label{app:mitig_limits}
We use one model, one persona wording and 300 images. At this size the $r_{\text{part}}$
intervals are about $\pm0.11$, so the advantage of the best arm over the direct prompt and over the
best VLM is suggestive but not significant after correction. The human evidence is itself noisy (only
41 raters are aged 60 or older), which bounds any grounded method, and the agents receive aggregate
human data rather than individual raters. A 60-image pilot run before the fixes described in
Appendix~\ref{app:agents} is not reported.

\begingroup
\FloatBarrierAbove
\section{Robustness Analyses and Additional Experiments}
\label{app:robust}

This appendix reports analyses added in revision. Sections~\ref{app:rb-rater}--\ref{app:rb-ceiling} test whether the fidelity null could arise from the small number of older raters or from the way the human age effect is estimated. Sections~\ref{app:rb-prompts}--\ref{app:rb-naps} test whether the findings depend on the prompt or the dataset. Sections~\ref{app:rb-geometry}--\ref{app:heldout} report a like-for-like human geometry analysis, a downstream prediction test, and a held-out replication of the best mitigation arm. The new model runs comprise 53{,}207 calls.

\subsection{Rater uncertainty}
\label{app:rb-rater}
The intervals in the main text resample images and treat the human age effects $\Delta H$ as fixed. Here each bootstrap replicate resamples raters within list, refits the hierarchical model of Eq.~\ref{eq:human}, and recomputes $r_{\text{part}}$; a second variant resamples raters and images jointly, and a leave-one-out analysis refits the model once per rater aged 60+ with that rater removed (Tables~\ref{tab:rb_rater_valence} and~\ref{tab:rb_rater_arousal}). With raters and images resampled, 0 of 17 valence intervals exclude zero and the largest upper bound is $0.187$; removing any single older rater moves the best model (GLM-4.5V) only within $[0.083,0.102]$. On arousal, the negative image-only intervals do not survive rater resampling, which supports not interpreting arousal fidelity. Resampling raters adds noise to the human estimate and therefore pulls intervals toward zero; Section~\ref{app:rb-power} shows the design retains power despite this.

\begin{table}[tbp]
\centering
\small
\caption{\textbf{Fidelity with rater uncertainty, valence.} In each of 1{,}000 bootstrap replicates, raters are resampled within list, the hierarchical human model (Eq.~\ref{eq:human}) is refitted, and $r_{\text{part}}$ is recomputed. ``Raters + images'' resamples both. The last column refits the model once per rater aged 60+ with that rater removed. The three interval columns report 95\% confidence intervals and identify the resampling unit.}
\label{tab:rb_rater_valence}
\resizebox{\linewidth}{!}{%
\begin{tabular}{lrrrrr}
\toprule
\rowcolor{titlebar!90}
Model & $r_{\text{part}}$ & Images & Raters & Raters + images & Leave one 60+ rater out \\
\midrule
Claude-Opus-5 & 0.058 & [$-$0.021, 0.134] & [$-$0.054, 0.149] & [$-$0.085, 0.183] & [0.041, 0.070] \\
Claude-Sonnet-4.6 & 0.086 & [0.006, 0.164] & [$-$0.028, 0.160] & [$-$0.069, 0.187] & [0.073, 0.096] \\
Gemini-3-Flash & 0.065 & [$-$0.004, 0.139] & [$-$0.033, 0.139] & [$-$0.063, 0.167] & [0.050, 0.077] \\
Gemini-3.8-Flash & 0.059 & [$-$0.011, 0.130] & [$-$0.033, 0.126] & [$-$0.069, 0.148] & [0.049, 0.067] \\
Gemma-3-27B & 0.009 & [$-$0.056, 0.076] & [$-$0.079, 0.098] & [$-$0.113, 0.124] & [$-$0.009, 0.021] \\
Gemma-4-31B & 0.042 & [$-$0.032, 0.118] & [$-$0.059, 0.125] & [$-$0.088, 0.155] & [0.032, 0.053] \\
GLM-4.5V & 0.093 & [0.027, 0.161] & [0.002, 0.141] & [$-$0.031, 0.173] & [0.083, 0.102] \\
GPT-5-mini & 0.037 & [$-$0.037, 0.111] & [$-$0.054, 0.109] & [$-$0.093, 0.134] & [0.018, 0.046] \\
GPT-5.2 & $-$0.030 & [$-$0.095, 0.035] & [$-$0.111, 0.064] & [$-$0.139, 0.094] & [$-$0.040, $-$0.020] \\
GPT-5.6-Luna-Pro & 0.006 & [$-$0.063, 0.072] & [$-$0.063, 0.066] & [$-$0.098, 0.100] & [$-$0.002, 0.012] \\
Grok-4.6 & $-$0.001 & [$-$0.077, 0.074] & [$-$0.085, 0.088] & [$-$0.119, 0.107] & [$-$0.012, 0.011] \\
Llama-4-Scout & $-$0.017 & [$-$0.083, 0.048] & [$-$0.074, 0.048] & [$-$0.111, 0.083] & [$-$0.025, $-$0.011] \\
Qwen3-VL-235B & 0.072 & [$-$0.002, 0.143] & [$-$0.036, 0.143] & [$-$0.071, 0.170] & [0.061, 0.083] \\
Qwen3-VL-32B & 0.002 & [$-$0.073, 0.079] & [$-$0.091, 0.102] & [$-$0.118, 0.117] & [$-$0.016, 0.014] \\
Qwen3-VL-8B & 0.001 & [$-$0.072, 0.070] & [$-$0.061, 0.063] & [$-$0.099, 0.094] & [$-$0.009, 0.011] \\
Qwen3-VL-8B-Think & 0.044 & [$-$0.027, 0.111] & [$-$0.030, 0.095] & [$-$0.060, 0.119] & [0.035, 0.053] \\
Qwen3.8-27B & 0.024 & [$-$0.047, 0.098] & [$-$0.039, 0.078] & [$-$0.073, 0.106] & [0.015, 0.035] \\
17-model consensus & 0.079 & [0.007, 0.145] & [$-$0.027, 0.146] & [$-$0.059, 0.170] & [0.061, 0.090] \\
\bottomrule
\end{tabular}}
\end{table}

\begin{table}[tbp]
\centering
\small
\caption{\textbf{Fidelity with rater uncertainty, arousal} (300 replicates; columns as in Table~\ref{tab:rb_rater_valence}). Image-only intervals are negative for most models, but none excludes zero once raters are resampled. The three interval columns report 95\% confidence intervals and identify the resampling unit.}
\label{tab:rb_rater_arousal}
\resizebox{\linewidth}{!}{%
\begin{tabular}{lrrrrr}
\toprule
\rowcolor{titlebar!90}
Model & $r_{\text{part}}$ & Images & Raters & Raters + images & Leave one 60+ rater out \\
\midrule
Claude-Opus-5 & $-$0.132 & [$-$0.189, $-$0.077] & [$-$0.039, 0.185] & [$-$0.060, 0.198] & [$-$0.167, $-$0.103] \\
Claude-Sonnet-4.6 & $-$0.236 & [$-$0.298, $-$0.173] & [$-$0.125, 0.112] & [$-$0.142, 0.145] & [$-$0.264, $-$0.211] \\
Gemini-3-Flash & $-$0.084 & [$-$0.153, $-$0.014] & [0.002, 0.197] & [$-$0.032, 0.208] & [$-$0.121, $-$0.057] \\
Gemini-3.8-Flash & $-$0.149 & [$-$0.223, $-$0.078] & [$-$0.103, 0.062] & [$-$0.132, 0.080] & [$-$0.164, $-$0.136] \\
Gemma-3-27B & $-$0.223 & [$-$0.282, $-$0.164] & [$-$0.118, 0.161] & [$-$0.145, 0.165] & [$-$0.252, $-$0.194] \\
Gemma-4-31B & $-$0.289 & [$-$0.349, $-$0.225] & [$-$0.182, 0.084] & [$-$0.193, 0.082] & [$-$0.314, $-$0.266] \\
GLM-4.5V & $-$0.164 & [$-$0.224, $-$0.104] & [$-$0.115, 0.059] & [$-$0.137, 0.068] & [$-$0.175, $-$0.151] \\
GPT-5-mini & $-$0.108 & [$-$0.170, $-$0.049] & [$-$0.046, 0.137] & [$-$0.063, 0.152] & [$-$0.132, $-$0.094] \\
GPT-5.2 & $-$0.118 & [$-$0.178, $-$0.057] & [$-$0.065, 0.095] & [$-$0.088, 0.113] & [$-$0.132, $-$0.103] \\
GPT-5.6-Luna-Pro & $-$0.082 & [$-$0.139, $-$0.023] & [$-$0.060, 0.090] & [$-$0.085, 0.110] & [$-$0.096, $-$0.072] \\
Grok-4.6 & $-$0.167 & [$-$0.229, $-$0.103] & [$-$0.097, 0.100] & [$-$0.120, 0.118] & [$-$0.185, $-$0.149] \\
Llama-4-Scout & $-$0.016 & [$-$0.076, 0.047] & [$-$0.027, 0.072] & [$-$0.048, 0.105] & [$-$0.022, $-$0.009] \\
Qwen3-VL-235B & $-$0.067 & [$-$0.128, $-$0.007] & [$-$0.043, 0.069] & [$-$0.076, 0.095] & [$-$0.074, $-$0.057] \\
Qwen3-VL-32B & $-$0.190 & [$-$0.250, $-$0.129] & [$-$0.101, 0.112] & [$-$0.126, 0.135] & [$-$0.214, $-$0.167] \\
Qwen3-VL-8B & $-$0.108 & [$-$0.173, $-$0.048] & [$-$0.079, 0.045] & [$-$0.111, 0.063] & [$-$0.115, $-$0.098] \\
Qwen3-VL-8B-Think & $-$0.083 & [$-$0.147, $-$0.024] & [$-$0.087, 0.020] & [$-$0.118, 0.060] & [$-$0.089, $-$0.078] \\
Qwen3.8-27B & $-$0.108 & [$-$0.168, $-$0.051] & [$-$0.084, 0.041] & [$-$0.115, 0.064] & [$-$0.115, $-$0.097] \\
17-model consensus & $-$0.241 & [$-$0.299, $-$0.183] & [$-$0.143, 0.126] & [$-$0.172, 0.140] & [$-$0.263, $-$0.217] \\
\bottomrule
\end{tabular}}
\end{table}

\subsection{Alternative estimators of the human age effect}
\label{app:rb-estimators}
The main analysis uses a linear age slope with partial pooling, whereas the model response is close to a step at 60 ($\S$\ref{app:shape}). Table~\ref{tab:rb_estimators} re-estimates the human effect ten ways: older-group contrasts at 60 and 50, two nonlinear age codings, raw group means, and an unpooled slope. Median $r_{\text{part}}$ is $0.05$ or below under every estimator. Contrasts built on the 25 valence raters aged 60+ have reliability near zero against under-60 raters, so they cannot serve as a target; the 50+ contrast (78 raters) has a usable ceiling ($0.52$) and gives median $r_{\text{part}}=0.003$. The weak positive association under the 60+ contrasts does not survive resampling raters and images (Table~\ref{tab:rb_contrast_boot}). The linear slope is the most reliable target: its split-half ceiling ($0.62$) agrees with the model-based ceiling ($0.595$).

\begin{table}[tbp]
\centering
\small
\caption{\textbf{Fidelity under ten estimators of the human age effect (valence).} ``Pooled'' uses the partially pooled model with a different age predictor; ``raw means'' is a plain difference of group means per image. Split-half reliability splits raters into two disjoint halves, estimates the pleasantness-adjusted effect in each, and applies Spearman--Brown; the ceiling is its square root.}
\label{tab:rb_estimators}
\resizebox{\linewidth}{!}{%
\begin{tabular}{llrrrrrr}
\toprule
\rowcolor{titlebar!90}
Human age effect & Older raters & Median $r_{\text{part}}$ & Range & CI $>0$ (images) & Consensus & Split-half reliability [95\%] & Ceiling \\
\midrule
linear slope, pooled (paper) & all 413 contribute & 0.037 & $-$0.030 to 0.093 & 2 & 0.079 & 0.39 [0.26, 0.50] & 0.62 \\
60+ vs under-60, pooled & 25 & 0.043 & $-$0.030 to 0.132 & 6 & 0.125 & 0.01 [$-$0.16, 0.11] & 0.09 \\
60+ vs 18-$-$29, pooled & 25 & 0.047 & $-$0.028 to 0.139 & 7 & 0.127 & $-$0.13 [$-$0.43, 0.02] & 0.00 \\
50+ vs under-50, pooled & 78 & 0.003 & $-$0.077 to 0.068 & 0 & 0.045 & 0.27 [0.12, 0.40] & 0.52 \\
hinge at 45 (nonlinear), pooled & all 413 contribute & 0.046 & $-$0.023 to 0.110 & 4 & 0.110 & 0.22 [0.07, 0.38] & 0.47 \\
log age (nonlinear), pooled & all 413 contribute & 0.025 & $-$0.037 to 0.086 & 2 & 0.070 & 0.38 [0.25, 0.48] & 0.62 \\
60+ vs under-60, raw means & 25 & 0.047 & $-$0.023 to 0.145 & 7 & 0.131 & $-$0.02 [$-$0.21, 0.13] & 0.00 \\
60+ vs 18-$-$29, raw means & 25 & 0.036 & $-$0.031 to 0.122 & 7 & 0.122 & 0.15 [$-$0.02, 0.28] & 0.39 \\
3-band trend, raw means & 25 in the older band & 0.036 & $-$0.031 to 0.122 & 7 & 0.123 & -- & -- \\
linear slope, no pooling (OLS) & all 413 contribute & 0.037 & $-$0.032 to 0.091 & 2 & 0.078 & -- & -- \\
\bottomrule
\end{tabular}}
\end{table}

\begin{table}[tbp]
\centering
\small
\caption{\textbf{Rater bootstrap for older-group contrasts} (2{,}000 replicates; raters resampled within list). Model side: older minus the mean of young and middle.}
\label{tab:rb_contrast_boot}
\resizebox{\linewidth}{!}{%
\begin{tabular}{lrrrrr}
\toprule
\rowcolor{titlebar!90}
Contrast & Median $r_{\text{part}}$ & Range & CI $>0$, raters & CI $>0$, raters + images & Consensus [95\% CI, raters + images] \\
\midrule
60+ vs under-60, raw means & 0.048 & $-$0.010 to 0.136 & 1 of 17 & 0 of 17 & 0.126 [$-$0.034, 0.209] \\
60+ vs 18-$-$29, raw means & 0.042 & $-$0.017 to 0.110 & 1 of 17 & 0 of 17 & 0.114 [$-$0.046, 0.205] \\
50+ vs under-50, raw means & $-$0.013 & $-$0.074 to 0.053 & 0 of 17 & 0 of 17 & 0.042 [$-$0.091, 0.162] \\
50+ vs 18-$-$29, raw means & 0.003 & $-$0.069 to 0.058 & 0 of 17 & 0 of 17 & 0.043 [$-$0.090, 0.150] \\
\bottomrule
\end{tabular}}
\end{table}

\subsection{Power}
\label{app:rb-power}
Table~\ref{tab:rb_power} plants a model age effect of known true fidelity into the real OASIS design and runs the full pipeline. Under graded age effects, the design has 92\% power at a true fidelity of $0.20$, and a true fidelity of $0.30$ would produce the best model's observed value ($0.093$) in only $1.7\%$ of runs. If human age effects were a pure step at 60, even a model with fidelity $1$ would show only $0.10$ to $0.19$, so that scenario could not be tested. The data do not resemble it: a pure step would give the linear age slope an SD of $0.016$ across images, whereas the observed SD is $0.106$. The simulation is slightly conservative (simulated SD $0.090$ under the graded model).

\begin{table}[tbp]
\centering
\small
\caption{\textbf{Power of the design.} A model age effect with known true fidelity is planted into the real OASIS design (same raters, ages, lists, assignments, fitted noise levels, integer 1--7 ratings) and the full pipeline is run 300 times per setting. Power is the share of runs above the 97.5th percentile of zero-fidelity runs; the last columns give the probability of observing a value at or below the best model (0.093) and the median model (0.037). (a) graded human age effects (the paper's model); (b) a pure step at 60 analysed with the linear slope; (c) a pure step at 60 analysed with a 60+ contrast.}
\label{tab:rb_power}
\resizebox{\linewidth}{!}{%
\begin{tabular}{lrrrrrrrr}
\toprule
\rowcolor{titlebar!90}
 & \multicolumn{4}{c}{(a) Graded effects} & \multicolumn{2}{c}{(b) Step, slope} & \multicolumn{2}{c}{(c) Step, contrast} \\
\cmidrule(lr){2-5}\cmidrule(lr){6-7}\cmidrule(lr){8-9}
\rowcolor{titlebar}
True fidelity & Mean observed & Power & $P(\le 0.093)$ & $P(\le 0.037)$ & Mean observed & Power & Mean observed & Power \\
\midrule
0.00 & $-$0.003 & 3\% & 0.993 & 0.890 & $-$0.003 & 3\% & $-$0.003 & 3\% \\
0.05 & 0.025 & 12\% & 0.977 & 0.623 & 0.002 & 4\% & 0.007 & 4\% \\
0.10 & 0.054 & 41\% & 0.883 & 0.310 & 0.007 & 5\% & 0.016 & 5\% \\
0.15 & 0.082 & 71\% & 0.627 & 0.093 & 0.013 & 9\% & 0.025 & 9\% \\
0.20 & 0.111 & 92\% & 0.283 & 0.020 & 0.018 & 11\% & 0.034 & 14\% \\
0.30 & 0.168 & 100\% & 0.017 & 0.000 & 0.028 & 17\% & 0.053 & 32\% \\
0.40 & 0.225 & 100\% & 0.000 & 0.000 & 0.039 & 25\% & 0.071 & 53\% \\
0.50 & 0.282 & 100\% & 0.000 & 0.000 & 0.049 & 37\% & 0.090 & 76\% \\
0.60 & 0.340 & 100\% & 0.000 & 0.000 & 0.060 & 51\% & 0.109 & 89\% \\
0.80 & 0.455 & 100\% & 0.000 & 0.000 & 0.081 & 76\% & 0.146 & 99\% \\
1.00 & 0.571 & 100\% & 0.000 & 0.000 & 0.104 & 92\% & 0.185 & 100\% \\
\bottomrule
\end{tabular}}
\end{table}

\subsection{Corrected ceiling and magnitude in matched units}
\label{app:rb-ceiling}
The ceiling for $r_{\text{part}}$ is $0.595$; the value $0.86$ applies to the raw age effect. Against $0.595$, the best model reaches $15.6\%$ of the attainable value and the median model $6.3\%$ (Table~\ref{tab:rb_ceiling_magnitude}). In matched units (rating points, older at 67 versus young at 23.5 years), the image-specific human effect has SD $0.441$, and model shifts are a median $1.4$ times larger ($2.2$ times after removing pleasantness). The magnitude mismatch is therefore real but smaller than a comparison in mixed units suggests.

\begin{table}[tbp]
\centering
\small
\caption{\textbf{Corrected ceiling and magnitude in matched units (valence).} Human and model age effects are expressed in rating points for older (67 years) versus young (23.5 years). The image-specific human effect has SD $0.441$ points ($0.281$ after removing pleasantness). Percentages use the ceiling for the pleasantness-adjusted effect ($0.595$) and, for reference, the raw-effect ceiling ($0.86$).}
\label{tab:rb_ceiling_magnitude}
\resizebox{\linewidth}{!}{%
\begin{tabular}{lrrrrrrr}
\toprule
\rowcolor{titlebar!90}
Model & Mean older $-$ young & SD of image-specific shift & Ratio to human SD & Ratio, pleasantness removed & $r_{\text{part}}$ & \% of 0.595 & \% of 0.86 \\
\midrule
GLM-4.5V & $-$0.143 & 0.852 & 1.93 & 3.03 & 0.093 & 15.6\% & 10.8\% \\
Claude-Sonnet-4.6 & $-$0.126 & 0.627 & 1.42 & 2.20 & 0.086 & 14.4\% & 10.0\% \\
Qwen3-VL-235B & $-$0.226 & 0.691 & 1.57 & 2.42 & 0.072 & 12.2\% & 8.4\% \\
Gemini-3-Flash & $-$0.028 & 0.592 & 1.34 & 2.05 & 0.065 & 10.9\% & 7.6\% \\
Gemini-3.8-Flash & $-$0.045 & 0.477 & 1.08 & 1.68 & 0.059 & 9.9\% & 6.9\% \\
Claude-Opus-5 & $-$0.116 & 0.385 & 0.87 & 1.35 & 0.058 & 9.8\% & 6.8\% \\
Qwen3-VL-8B-Think & +0.015 & 0.800 & 1.82 & 2.83 & 0.044 & 7.4\% & 5.1\% \\
Gemma-4-31B & $-$0.057 & 0.588 & 1.33 & 2.09 & 0.042 & 7.1\% & 4.9\% \\
GPT-5-mini & $-$0.132 & 0.736 & 1.67 & 2.61 & 0.037 & 6.3\% & 4.3\% \\
Qwen3.8-27B & $-$0.005 & 0.466 & 1.06 & 1.66 & 0.024 & 4.1\% & 2.8\% \\
Gemma-3-27B & $-$0.157 & 0.750 & 1.70 & 2.64 & 0.009 & 1.5\% & 1.0\% \\
GPT-5.6-Luna-Pro & $-$0.076 & 0.457 & 1.04 & 1.61 & 0.006 & 0.9\% & 0.6\% \\
Qwen3-VL-32B & $-$0.219 & 0.683 & 1.55 & 2.40 & 0.002 & 0.4\% & 0.3\% \\
Qwen3-VL-8B & $-$0.020 & 0.327 & 0.74 & 1.16 & 0.001 & 0.2\% & 0.1\% \\
Grok-4.6 & $-$0.170 & 0.704 & 1.60 & 2.48 & $-$0.001 & $-$0.2\% & $-$0.1\% \\
Llama-4-Scout & $-$0.092 & 0.527 & 1.20 & 1.87 & $-$0.017 & $-$2.8\% & $-$1.9\% \\
GPT-5.2 & $-$0.091 & 0.670 & 1.52 & 2.38 & $-$0.030 & $-$5.1\% & $-$3.5\% \\
\bottomrule
\end{tabular}}
\end{table}

\subsection{Prompt formats}
\label{app:rb-prompts}
Five models from five developers (GPT-5.6-Luna-Pro, Claude-Sonnet-4.6, Gemini-3-Flash, Qwen3-VL-235B, Llama-4-Scout) rated the 300-image subset under young, middle, and older personas in seven formats, each changing one element of the submission prompt:
\begin{itemize}[leftmargin=1.2em,itemsep=0pt,topsep=2pt]
\item \textbf{Chain of thought:} reason in two to four sentences, then give the JSON.
\item \textbf{Direct rating only:} valence and arousal, with no description and no emotion label.
\item \textbf{Paraphrase A:} ``Please answer as a person who is \{age\}, giving the response that someone of this age would give.''
\item \textbf{Paraphrase B:} ``Take the perspective of an adult who is \{age\} when you rate this image.''
\item \textbf{Numeric age:} 24, 45, or 67 in place of the age band.
\item \textbf{System prompt:} the unchanged age sentence moved to the system turn.
\item \textbf{Interview-style} \citep{lutz2025persona}: ``Interviewer: How old are you? Interviewee: I am \{n\} years old. Interviewer: Thank you. Please answer the next task as yourself.''
\end{itemize} The full prompts are shown in Figures~\ref{fig:prompts-rb1}--\ref{fig:prompts-rb3}.

\begin{figure}[p]
  \centering
  \promptpanel{5}{35 102.1 35 106.5}{(a)~Chain of thought}\vspace{4pt}
  \promptpanel{6}{35 137.2 35 141.6}{(b)~Direct rating only}\vspace{4pt}
  \promptpanel{7}{35 85.9 35 122.7}{(c)~Paraphrase A}\vspace{4pt}
  \caption{\textbf{Prompt formats for the robustness study (1/3).} Each format changes one element of the submission prompt (Figure~\ref{fig:prompts-main}); highlighted text marks the change. \{age\} is ``18 to 29 years old'', ``30 to 59 years old'' or ``60 years old or older''.}
  \label{fig:prompts-rb1}
\end{figure}

\begin{figure}[p]
  \centering
  \promptpanel{8}{35 85.9 35 122.7}{(d)~Paraphrase B}\vspace{4pt}
  \promptpanel{9}{35 102.1 35 106.5}{(e)~Numeric age}\vspace{4pt}
  \promptpanel{10}{35 59.6 35 64.0}{(f)~Age sentence in the system prompt}\vspace{4pt}
  \caption{\textbf{Prompt formats for the robustness study (2/3).} Highlighted text marks the change from the submission prompt. In (e), \{n\} is 24, 45 or 67 for the young, middle and older personas.}
  \label{fig:prompts-rb2}
\end{figure}

\begin{figure}[p]
  \centering
  \promptpanel{11}{35 77.1 35 81.5}{(g)~Interview-style persona}\vspace{4pt}
  \promptpanel{12}{35 142.6 35 147.0}{(h)~Third person, rating only: valence call}\vspace{4pt}
  \promptpanel{13}{35 133.9 35 138.3}{(i)~Third person, rating only: arousal call}\vspace{4pt}
  \caption{\textbf{Prompt formats for the robustness study (3/3).} (g)~Interview-style persona \citep{lutz2025persona}. (h--i)~Third-person, rating-only prompt used for Tables~\ref{tab:rb_qwen_format} and~\ref{tab:rb_thirdperson}: a system prompt, no description, and valence and arousal requested in separate calls.}
  \label{fig:prompts-rb3}
\end{figure}

Six of seven formats reproduce the valence findings: pleasant flagged images fall in all five models, and models move largely the same images as under the submission prompt (median per-image $r$ $0.67$ to $0.77$; Table~\ref{tab:rb_formats_val}). No format raises fidelity: 0 of 35 new $r_{\text{part}}$ intervals lie above zero and 0 of 35 paired differences from the submission prompt exclude zero (Table~\ref{tab:rb_formats_val_full}). The interview-style format removes most of the persona response and leaves fidelity near zero. With intervals of about $\pm0.13$ on 300 images, these runs exclude a clearly higher fidelity, not a small difference.

Two further prompts request ratings only, in the third person (``Rate how pleasant or unpleasant this image makes an adult aged 60 or older feel''), with valence and arousal in separate calls and a system prompt. On three Qwen3-VL models and all 900 images (Table~\ref{tab:rb_qwen_format}), and on six models from different developers (Table~\ref{tab:rb_thirdperson}), the older-persona valence downshift, the larger shift on flagged images, the reversal on pleasant flagged images, and near-zero $r_{\text{part}}$ all appear with no description requested; 6 of 6 models lower valence (largest $p=0.001$). The arousal downshift, by contrast, holds in all first-person formats (Table~\ref{tab:rb_formats_aro}) but in only 2 of 6 models under the third-person prompt, so we restrict the arousal claim to first-person persona prompts. Because the third-person prompt changes several elements at once, these runs do not identify which element removes the arousal shift.

\begin{table}[tbp]
\centering
\small
\caption{\textbf{Seven further prompt formats, valence: summary over five models} (GPT-5.6-Luna-Pro, Claude-Sonnet-4.6, Gemini-3-Flash, Qwen3-VL-235B, Llama-4-Scout; 300 images; 31{,}500 calls, 99.99\% valid). Each format changes one element of the submission prompt (see text). \emph{Pleasant flagged}: 17 images in this subset.}
\label{tab:rb_formats_val}
\resizebox{\linewidth}{!}{%
\begin{tabular}{lrrrrrrrr}
\toprule
\rowcolor{titlebar!90}
Format & Median older $-$ young & Negative shift & of which $p<.05$ & Median, pleasant flagged & Median $r_{\text{part}}$ & Range of $r_{\text{part}}$ & CI $>0$ & Median per-image $r$ with submission \\
\midrule
Submission prompt & $-$0.065 & 4 of 5 & 3 & $-$0.812 & 0.085 & 0.006 to 0.136 & 1 of 5 & -- \\
Chain of thought & $-$0.031 & 4 of 5 & 2 & $-$0.865 & 0.086 & 0.053 to 0.106 & 0 of 5 & 0.67 \\
Direct rating only & $-$0.096 & 4 of 5 & 4 & $-$0.600 & 0.088 & $-$0.008 to 0.128 & 0 of 5 & 0.67 \\
Paraphrase A & $-$0.099 & 4 of 5 & 4 & $-$0.753 & 0.093 & 0.045 to 0.121 & 0 of 5 & 0.70 \\
Paraphrase B & $-$0.069 & 5 of 5 & 3 & $-$0.759 & 0.120 & 0.036 to 0.135 & 0 of 5 & 0.77 \\
Numeric age & $-$0.047 & 5 of 5 & 2 & $-$0.712 & 0.076 & 0.052 to 0.116 & 0 of 5 & 0.73 \\
System prompt & $-$0.090 & 4 of 5 & 4 & $-$0.765 & 0.098 & 0.013 to 0.133 & 0 of 5 & 0.74 \\
Interview-style & $-$0.024 & 3 of 5 & 2 & $-$0.065 & 0.015 & $-$0.022 to 0.027 & 0 of 5 & 0.17 \\
\bottomrule
\end{tabular}}
\end{table}

\begin{table}[tbp]
\centering
\small
\caption{\textbf{Seven further prompt formats, arousal: summary over five models} (columns as in Table~\ref{tab:rb_formats_val}).}
\label{tab:rb_formats_aro}
\resizebox{\linewidth}{!}{%
\begin{tabular}{lrrrrrrrr}
\toprule
\rowcolor{titlebar!90}
Format & Median older $-$ young & Negative shift & of which $p<.05$ & Median, pleasant flagged & Median $r_{\text{part}}$ & Range of $r_{\text{part}}$ & CI $>0$ & Median per-image $r$ with submission \\
\midrule
Submission prompt & $-$0.184 & 5 of 5 & 5 & $-$0.647 & $-$0.133 & $-$0.237 to $-$0.057 & 0 of 5 & -- \\
Chain of thought & $-$0.452 & 5 of 5 & 5 & $-$0.994 & $-$0.190 & $-$0.216 to $-$0.139 & 0 of 5 & 0.26 \\
Direct rating only & $-$0.233 & 5 of 5 & 5 & $-$0.629 & $-$0.065 & $-$0.146 to 0.004 & 0 of 5 & 0.26 \\
Paraphrase A & $-$0.250 & 5 of 5 & 5 & $-$0.694 & $-$0.073 & $-$0.168 to $-$0.058 & 0 of 5 & 0.41 \\
Paraphrase B & $-$0.237 & 5 of 5 & 5 & $-$0.706 & $-$0.036 & $-$0.191 to 0.012 & 0 of 5 & 0.47 \\
Numeric age & $-$0.128 & 5 of 5 & 4 & $-$0.476 & $-$0.121 & $-$0.159 to $-$0.058 & 0 of 5 & 0.46 \\
System prompt & $-$0.185 & 5 of 5 & 5 & $-$0.765 & $-$0.104 & $-$0.195 to 0.044 & 0 of 5 & 0.46 \\
Interview-style & $-$0.047 & 3 of 5 & 3 & $-$0.135 & $-$0.035 & $-$0.082 to 0.038 & 0 of 5 & 0.11 \\
\bottomrule
\end{tabular}}
\end{table}

\begin{table}[tbp]
\centering
\scriptsize
\caption{\textbf{Every model and format, valence.} ``Change'' is the paired bootstrap difference in $r_{\text{part}}$ from the same model's submission-prompt value.}
\label{tab:rb_formats_val_full}
\resizebox{\linewidth}{!}{%
\begin{tabular}{llrrrrrrrr}
\toprule
\rowcolor{titlebar!90}
Model & Format & $n$ & Older $-$ young & $p$ & Flagged & Pleasant flagged & $r_{\text{part}}$ [95\% CI] & Change [95\% CI] & Per-image $r$ \\
\midrule
GPT-5.6-Luna-Pro & Submission prompt & 298 & $-$0.065 & 0.005 & $-$0.173 & $-$0.535 & 0.081 [$-$0.054, 0.216] & -- & -- \\
GPT-5.6-Luna-Pro & Chain of thought & 300 & $-$0.031 & 0.367 & $-$0.212 & $-$0.865 & 0.106 [$-$0.021, 0.231] & +0.025 [$-$0.106, 0.171] & 0.30 \\
GPT-5.6-Luna-Pro & Direct rating only & 300 & $-$0.059 & 0.005 & $-$0.183 & $-$0.365 & $-$0.008 [$-$0.117, 0.099] & $-$0.089 [$-$0.253, 0.081] & 0.17 \\
GPT-5.6-Luna-Pro & Paraphrase A & 300 & $-$0.126 & $<$\,0.001 & $-$0.354 & $-$0.676 & 0.078 [$-$0.039, 0.205] & $-$0.003 [$-$0.147, 0.139] & 0.32 \\
GPT-5.6-Luna-Pro & Paraphrase B & 300 & $-$0.032 & 0.212 & $-$0.050 & $-$0.488 & 0.036 [$-$0.086, 0.169] & $-$0.045 [$-$0.185, 0.098] & 0.28 \\
GPT-5.6-Luna-Pro & Numeric age & 299 & $-$0.047 & 0.056 & $-$0.073 & $-$0.371 & 0.076 [$-$0.049, 0.203] & $-$0.005 [$-$0.151, 0.146] & 0.26 \\
GPT-5.6-Luna-Pro & System prompt & 300 & $-$0.073 & 0.004 & $-$0.240 & $-$0.641 & 0.013 [$-$0.112, 0.144] & $-$0.068 [$-$0.202, 0.074] & 0.25 \\
GPT-5.6-Luna-Pro & Interview-style & 300 & $-$0.027 & 0.286 & +0.002 & $-$0.041 & 0.015 [$-$0.083, 0.114] & $-$0.066 [$-$0.230, 0.099] & 0.01 \\
\midrule
Claude-Sonnet-4.6 & Submission prompt & 300 & $-$0.086 & 0.017 & $-$0.544 & $-$1.371 & 0.085 [$-$0.052, 0.228] & -- & -- \\
Claude-Sonnet-4.6 & Chain of thought & 300 & $-$0.156 & 0.001 & $-$0.758 & $-$1.865 & 0.086 [$-$0.052, 0.220] & +0.001 [$-$0.078, 0.074] & 0.80 \\
Claude-Sonnet-4.6 & Direct rating only & 300 & $-$0.132 & $<$\,0.001 & $-$0.481 & $-$0.965 & 0.088 [$-$0.034, 0.226] & +0.003 [$-$0.100, 0.107] & 0.79 \\
Claude-Sonnet-4.6 & Paraphrase A & 300 & $-$0.099 & 0.012 & $-$0.588 & $-$1.371 & 0.121 [$-$0.013, 0.258] & +0.036 [$-$0.007, 0.079] & 0.93 \\
Claude-Sonnet-4.6 & Paraphrase B & 300 & $-$0.069 & 0.036 & $-$0.521 & $-$1.253 & 0.135 [$-$0.018, 0.279] & +0.050 [$-$0.004, 0.104] & 0.91 \\
Claude-Sonnet-4.6 & Numeric age & 300 & $-$0.058 & 0.046 & $-$0.394 & $-$0.971 & 0.116 [$-$0.020, 0.253] & +0.032 [$-$0.031, 0.093] & 0.88 \\
Claude-Sonnet-4.6 & System prompt & 300 & $-$0.090 & 0.006 & $-$0.538 & $-$1.206 & 0.108 [$-$0.029, 0.246] & +0.023 [$-$0.035, 0.076] & 0.91 \\
Claude-Sonnet-4.6 & Interview-style & 299 & $-$0.024 & 0.047 & $-$0.065 & $-$0.112 & 0.002 [$-$0.108, 0.114] & $-$0.083 [$-$0.254, 0.083] & 0.37 \\
\midrule
Gemini-3-Flash & Submission prompt & 297 & +0.020 & 0.530 & $-$0.324 & $-$0.812 & 0.107 [$-$0.023, 0.242] & -- & -- \\
Gemini-3-Flash & Chain of thought & 300 & +0.038 & 0.272 & $-$0.256 & $-$0.694 & 0.094 [$-$0.024, 0.217] & $-$0.013 [$-$0.102, 0.083] & 0.68 \\
Gemini-3-Flash & Direct rating only & 300 & +0.009 & 0.713 & $-$0.275 & $-$0.600 & 0.128 [$-$0.006, 0.264] & +0.021 [$-$0.070, 0.120] & 0.67 \\
Gemini-3-Flash & Paraphrase A & 300 & +0.015 & 0.654 & $-$0.302 & $-$0.753 & 0.106 [$-$0.019, 0.243] & $-$0.001 [$-$0.068, 0.069] & 0.80 \\
Gemini-3-Flash & Paraphrase B & 300 & $-$0.013 & 0.675 & $-$0.373 & $-$0.759 & 0.120 [$-$0.019, 0.265] & +0.013 [$-$0.061, 0.093] & 0.81 \\
Gemini-3-Flash & Numeric age & 300 & $-$0.018 & 0.469 & $-$0.300 & $-$0.712 & 0.105 [$-$0.033, 0.247] & $-$0.002 [$-$0.092, 0.088] & 0.73 \\
Gemini-3-Flash & System prompt & 300 & +0.022 & 0.500 & $-$0.321 & $-$0.606 & 0.133 [$-$0.013, 0.278] & +0.026 [$-$0.074, 0.128] & 0.74 \\
Gemini-3-Flash & Interview-style & 300 & +0.015 & 0.377 & $-$0.087 & $-$0.065 & $-$0.022 [$-$0.154, 0.115] & $-$0.129 [$-$0.279, 0.034] & 0.30 \\
\midrule
Qwen3-VL-235B & Submission prompt & 300 & $-$0.236 & $<$\,0.001 & $-$0.665 & $-$1.476 & 0.136 [0.003, 0.265] & -- & -- \\
Qwen3-VL-235B & Chain of thought & 300 & $-$0.248 & $<$\,0.001 & $-$0.773 & $-$1.806 & 0.053 [$-$0.056, 0.162] & $-$0.083 [$-$0.182, 0.016] & 0.67 \\
Qwen3-VL-235B & Direct rating only & 299 & $-$0.272 & $<$\,0.001 & $-$0.688 & $-$1.753 & 0.091 [$-$0.044, 0.221] & $-$0.045 [$-$0.148, 0.054] & 0.69 \\
Qwen3-VL-235B & Paraphrase A & 300 & $-$0.237 & $<$\,0.001 & $-$0.798 & $-$2.088 & 0.045 [$-$0.086, 0.175] & $-$0.091 [$-$0.190, 0.001] & 0.70 \\
Qwen3-VL-235B & Paraphrase B & 300 & $-$0.170 & $<$\,0.001 & $-$0.671 & $-$1.500 & 0.127 [$-$0.002, 0.260] & $-$0.009 [$-$0.075, 0.064] & 0.77 \\
Qwen3-VL-235B & Numeric age & 300 & $-$0.161 & $<$\,0.001 & $-$0.573 & $-$1.324 & 0.055 [$-$0.061, 0.171] & $-$0.081 [$-$0.159, 0.002] & 0.77 \\
Qwen3-VL-235B & System prompt & 300 & $-$0.169 & $<$\,0.001 & $-$0.654 & $-$1.629 & 0.090 [$-$0.041, 0.224] & $-$0.046 [$-$0.129, 0.037] & 0.77 \\
Qwen3-VL-235B & Interview-style & 300 & +0.020 & 0.204 & $-$0.050 & $-$0.053 & 0.027 [$-$0.102, 0.146] & $-$0.109 [$-$0.260, 0.044] & 0.17 \\
\midrule
Llama-4-Scout & Submission prompt & 300 & $-$0.033 & 0.267 & $-$0.225 & $-$0.365 & 0.006 [$-$0.115, 0.136] & -- & -- \\
Llama-4-Scout & Chain of thought & 300 & $-$0.021 & 0.595 & $-$0.187 & $-$0.706 & 0.070 [$-$0.048, 0.190] & +0.064 [$-$0.075, 0.202] & 0.35 \\
Llama-4-Scout & Direct rating only & 300 & $-$0.096 & 0.002 & $-$0.265 & $-$0.412 & 0.025 [$-$0.087, 0.132] & +0.019 [$-$0.122, 0.158] & 0.30 \\
Llama-4-Scout & Paraphrase A & 300 & $-$0.095 & 0.004 & $-$0.237 & $-$0.635 & 0.093 [$-$0.021, 0.222] & +0.088 [$-$0.039, 0.219] & 0.44 \\
Llama-4-Scout & Paraphrase B & 300 & $-$0.101 & $<$\,0.001 & $-$0.312 & $-$0.559 & 0.065 [$-$0.070, 0.200] & +0.060 [$-$0.108, 0.221] & 0.29 \\
Llama-4-Scout & Numeric age & 300 & $-$0.022 & 0.452 & +0.006 & $-$0.071 & 0.052 [$-$0.056, 0.154] & +0.046 [$-$0.112, 0.203] & 0.17 \\
Llama-4-Scout & System prompt & 300 & $-$0.248 & $<$\,0.001 & $-$0.433 & $-$0.765 & 0.098 [$-$0.019, 0.222] & +0.092 [$-$0.049, 0.231] & 0.37 \\
Llama-4-Scout & Interview-style & 300 & $-$0.105 & $<$\,0.001 & $-$0.123 & $-$0.218 & 0.024 [$-$0.093, 0.145] & +0.018 [$-$0.144, 0.174] & 0.06 \\
\bottomrule
\end{tabular}}
\end{table}

\begin{table}[tbp]
\centering
\scriptsize
\caption{\textbf{Every model and format, arousal} (columns as in Table~\ref{tab:rb_formats_val_full}).}
\label{tab:rb_formats_aro_full}
\resizebox{\linewidth}{!}{%
\begin{tabular}{llrrrrrrrr}
\toprule
\rowcolor{titlebar!90}
Model & Format & $n$ & Older $-$ young & $p$ & Flagged & Pleasant flagged & $r_{\text{part}}$ [95\% CI] & Change [95\% CI] & Per-image $r$ \\
\midrule
GPT-5.6-Luna-Pro & Submission prompt & 298 & $-$0.179 & $<$\,0.001 & $-$0.343 & $-$0.647 & $-$0.133 [$-$0.226, $-$0.031] & -- & -- \\
GPT-5.6-Luna-Pro & Chain of thought & 300 & $-$0.326 & $<$\,0.001 & $-$0.533 & $-$0.994 & $-$0.145 [$-$0.248, $-$0.037] & $-$0.012 [$-$0.123, 0.098] & 0.26 \\
GPT-5.6-Luna-Pro & Direct rating only & 300 & $-$0.165 & $<$\,0.001 & $-$0.321 & $-$0.629 & $-$0.049 [$-$0.156, 0.057] & +0.084 [$-$0.069, 0.222] & $-$0.04 \\
GPT-5.6-Luna-Pro & Paraphrase A & 300 & $-$0.190 & $<$\,0.001 & $-$0.335 & $-$0.694 & $-$0.059 [$-$0.165, 0.046] & +0.074 [$-$0.058, 0.203] & 0.18 \\
GPT-5.6-Luna-Pro & Paraphrase B & 300 & $-$0.253 & $<$\,0.001 & $-$0.202 & $-$0.300 & $-$0.016 [$-$0.123, 0.086] & +0.118 [$-$0.045, 0.266] & 0.02 \\
GPT-5.6-Luna-Pro & Numeric age & 299 & $-$0.126 & $<$\,0.001 & $-$0.275 & $-$0.476 & $-$0.107 [$-$0.219, 0.009] & +0.026 [$-$0.119, 0.169] & 0.09 \\
GPT-5.6-Luna-Pro & System prompt & 300 & $-$0.185 & $<$\,0.001 & $-$0.308 & $-$0.729 & 0.044 [$-$0.060, 0.144] & +0.177 [0.033, 0.314] & 0.12 \\
GPT-5.6-Luna-Pro & Interview-style & 300 & $-$0.047 & 0.048 & $-$0.050 & $-$0.135 & $-$0.082 [$-$0.185, 0.021] & +0.051 [$-$0.093, 0.191] & 0.11 \\
\midrule
Claude-Sonnet-4.6 & Submission prompt & 300 & $-$0.244 & $<$\,0.001 & $-$0.448 & $-$1.076 & $-$0.237 [$-$0.344, $-$0.129] & -- & -- \\
Claude-Sonnet-4.6 & Chain of thought & 300 & $-$0.414 & $<$\,0.001 & $-$0.658 & $-$1.618 & $-$0.190 [$-$0.295, $-$0.091] & +0.047 [$-$0.053, 0.148] & 0.58 \\
Claude-Sonnet-4.6 & Direct rating only & 300 & $-$0.186 & $<$\,0.001 & $-$0.306 & $-$0.706 & $-$0.146 [$-$0.245, $-$0.043] & +0.091 [0.006, 0.182] & 0.59 \\
Claude-Sonnet-4.6 & Paraphrase A & 300 & $-$0.250 & $<$\,0.001 & $-$0.419 & $-$1.094 & $-$0.168 [$-$0.266, $-$0.061] & +0.068 [0.007, 0.135] & 0.80 \\
Claude-Sonnet-4.6 & Paraphrase B & 300 & $-$0.237 & $<$\,0.001 & $-$0.471 & $-$1.071 & $-$0.191 [$-$0.283, $-$0.102] & +0.045 [$-$0.052, 0.130] & 0.75 \\
Claude-Sonnet-4.6 & Numeric age & 300 & $-$0.256 & $<$\,0.001 & $-$0.379 & $-$0.853 & $-$0.123 [$-$0.228, $-$0.025] & +0.113 [0.036, 0.192] & 0.71 \\
Claude-Sonnet-4.6 & System prompt & 300 & $-$0.148 & $<$\,0.001 & $-$0.273 & $-$0.929 & $-$0.131 [$-$0.239, $-$0.018] & +0.106 [0.028, 0.193] & 0.75 \\
Claude-Sonnet-4.6 & Interview-style & 299 & $-$0.067 & $<$\,0.001 & $-$0.077 & $-$0.224 & $-$0.035 [$-$0.130, 0.054] & +0.202 [0.078, 0.332] & 0.27 \\
\midrule
Gemini-3-Flash & Submission prompt & 297 & $-$0.352 & $<$\,0.001 & $-$0.471 & $-$1.341 & $-$0.140 [$-$0.242, $-$0.035] & -- & -- \\
Gemini-3-Flash & Chain of thought & 300 & $-$0.553 & $<$\,0.001 & $-$0.669 & $-$1.194 & $-$0.196 [$-$0.309, $-$0.085] & $-$0.056 [$-$0.161, 0.045] & 0.59 \\
Gemini-3-Flash & Direct rating only & 300 & $-$0.281 & $<$\,0.001 & $-$0.440 & $-$1.188 & $-$0.065 [$-$0.178, 0.038] & +0.075 [$-$0.021, 0.164] & 0.59 \\
Gemini-3-Flash & Paraphrase A & 300 & $-$0.374 & $<$\,0.001 & $-$0.473 & $-$1.212 & $-$0.135 [$-$0.237, $-$0.021] & +0.005 [$-$0.080, 0.091] & 0.68 \\
Gemini-3-Flash & Paraphrase B & 300 & $-$0.312 & $<$\,0.001 & $-$0.544 & $-$1.494 & $-$0.099 [$-$0.208, 0.011] & +0.042 [$-$0.049, 0.127] & 0.70 \\
Gemini-3-Flash & Numeric age & 300 & $-$0.430 & $<$\,0.001 & $-$0.592 & $-$1.694 & $-$0.159 [$-$0.260, $-$0.050] & $-$0.018 [$-$0.107, 0.065] & 0.69 \\
Gemini-3-Flash & System prompt & 300 & $-$0.296 & $<$\,0.001 & $-$0.488 & $-$1.353 & $-$0.067 [$-$0.166, 0.046] & +0.073 [$-$0.016, 0.157] & 0.65 \\
Gemini-3-Flash & Interview-style & 300 & $-$0.096 & $<$\,0.001 & $-$0.077 & $-$0.312 & 0.038 [$-$0.055, 0.134] & +0.178 [0.052, 0.304] & 0.29 \\
\midrule
Qwen3-VL-235B & Submission prompt & 300 & $-$0.184 & $<$\,0.001 & $-$0.215 & $-$0.500 & $-$0.076 [$-$0.179, 0.036] & -- & -- \\
Qwen3-VL-235B & Chain of thought & 300 & $-$0.672 & $<$\,0.001 & $-$0.433 & $-$0.824 & $-$0.216 [$-$0.322, $-$0.107] & $-$0.141 [$-$0.285, 0.006] & 0.22 \\
Qwen3-VL-235B & Direct rating only & 299 & $-$0.233 & $<$\,0.001 & $-$0.202 & $-$0.129 & 0.004 [$-$0.109, 0.123] & +0.079 [$-$0.065, 0.222] & 0.21 \\
Qwen3-VL-235B & Paraphrase A & 300 & $-$0.263 & $<$\,0.001 & $-$0.292 & $-$0.388 & $-$0.058 [$-$0.166, 0.051] & +0.017 [$-$0.100, 0.139] & 0.41 \\
Qwen3-VL-235B & Paraphrase B & 300 & $-$0.218 & $<$\,0.001 & $-$0.281 & $-$0.418 & 0.012 [$-$0.112, 0.130] & +0.087 [$-$0.029, 0.198] & 0.47 \\
Qwen3-VL-235B & Numeric age & 300 & $-$0.128 & $<$\,0.001 & $-$0.233 & $-$0.282 & $-$0.058 [$-$0.161, 0.053] & +0.018 [$-$0.085, 0.127] & 0.46 \\
Qwen3-VL-235B & System prompt & 300 & $-$0.168 & $<$\,0.001 & $-$0.267 & $-$0.388 & $-$0.104 [$-$0.217, 0.012] & $-$0.029 [$-$0.142, 0.086] & 0.46 \\
Qwen3-VL-235B & Interview-style & 300 & +0.045 & 0.016 & +0.019 & +0.112 & 0.027 [$-$0.086, 0.139] & +0.102 [$-$0.067, 0.271] & $-$0.10 \\
\midrule
Llama-4-Scout & Submission prompt & 300 & $-$0.164 & $<$\,0.001 & $-$0.340 & $-$0.359 & $-$0.057 [$-$0.167, 0.048] & -- & -- \\
Llama-4-Scout & Chain of thought & 300 & $-$0.452 & $<$\,0.001 & $-$0.613 & $-$0.635 & $-$0.139 [$-$0.245, $-$0.027] & $-$0.082 [$-$0.233, 0.077] & 0.00 \\
Llama-4-Scout & Direct rating only & 300 & $-$0.275 & $<$\,0.001 & $-$0.450 & $-$0.588 & $-$0.102 [$-$0.222, 0.017] & $-$0.045 [$-$0.173, 0.086] & 0.26 \\
Llama-4-Scout & Paraphrase A & 300 & $-$0.151 & $<$\,0.001 & $-$0.192 & $-$0.324 & $-$0.073 [$-$0.175, 0.036] & $-$0.016 [$-$0.164, 0.130] & $-$0.01 \\
Llama-4-Scout & Paraphrase B & 300 & $-$0.109 & 0.001 & $-$0.342 & $-$0.706 & $-$0.036 [$-$0.146, 0.073] & +0.021 [$-$0.115, 0.157] & 0.21 \\
Llama-4-Scout & Numeric age & 300 & $-$0.074 & 0.050 & $-$0.265 & $-$0.400 & $-$0.121 [$-$0.219, $-$0.018] & $-$0.064 [$-$0.198, 0.068] & 0.09 \\
Llama-4-Scout & System prompt & 300 & $-$0.243 & $<$\,0.001 & $-$0.352 & $-$0.765 & $-$0.195 [$-$0.290, $-$0.096] & $-$0.138 [$-$0.271, $-$0.011] & 0.17 \\
Llama-4-Scout & Interview-style & 300 & +0.002 & 0.960 & +0.031 & +0.065 & $-$0.036 [$-$0.137, 0.067] & +0.021 [$-$0.129, 0.168] & 0.06 \\
\bottomrule
\end{tabular}}
\end{table}

\begin{table}[tbp]
\centering
\small
\caption{\textbf{Second prompt format, Qwen3-VL models, all 900 images.} The second format is third person (``how does this make an adult aged 60 or older feel''), requests the rating only, places valence and arousal in separate calls, and uses a system prompt. \emph{Pleasant flagged}: the 54 sensitive images with pooled human valence $\ge 4$.}
\label{tab:rb_qwen_format}
\resizebox{\linewidth}{!}{%
\begin{tabular}{lllrrrrrr}
\toprule
\rowcolor{titlebar!90}
Dimension & Model & Format & Older $-$ young & Flagged & Unflagged & Pleasant flagged & $r_{\text{part}}$ [95\% CI] & Per-image $r$ between formats \\
\midrule
valence & Qwen3-VL-8B & paper prompt & $-$0.020 & $-$0.129 & +0.005 & $-$0.343 & 0.001 [$-$0.074, 0.071] & -- \\
valence & Qwen3-VL-8B & second format & $-$0.087 & $-$0.254 & $-$0.049 & $-$0.926 & $-$0.051 [$-$0.111, 0.016] & 0.35 \\
valence & Qwen3-VL-32B & paper prompt & $-$0.219 & $-$0.626 & $-$0.126 & $-$1.491 & 0.002 [$-$0.071, 0.075] & -- \\
valence & Qwen3-VL-32B & second format & $-$0.247 & $-$0.628 & $-$0.160 & $-$1.587 & $-$0.012 [$-$0.077, 0.052] & 0.65 \\
valence & Qwen3-VL-235B & paper prompt & $-$0.226 & $-$0.632 & $-$0.134 & $-$1.270 & 0.072 [0.002, 0.147] & -- \\
valence & Qwen3-VL-235B & second format & $-$0.348 & $-$0.720 & $-$0.263 & $-$1.493 & 0.058 [$-$0.011, 0.124] & 0.64 \\
\midrule
arousal & Qwen3-VL-8B & paper prompt & $-$0.296 & $-$0.423 & $-$0.268 & $-$0.370 & $-$0.108 [$-$0.170, $-$0.048] & -- \\
arousal & Qwen3-VL-8B & second format & $-$0.007 & $-$0.156 & +0.027 & $-$0.185 & $-$0.155 [$-$0.210, $-$0.104] & 0.14 \\
arousal & Qwen3-VL-32B & paper prompt & $-$0.295 & $-$0.451 & $-$0.259 & $-$0.796 & $-$0.190 [$-$0.250, $-$0.127] & -- \\
arousal & Qwen3-VL-32B & second format & +0.130 & $-$0.038 & +0.168 & $-$0.398 & $-$0.055 [$-$0.110, 0.001] & 0.22 \\
arousal & Qwen3-VL-235B & paper prompt & $-$0.245 & $-$0.276 & $-$0.238 & $-$0.496 & $-$0.067 [$-$0.127, $-$0.004] & -- \\
arousal & Qwen3-VL-235B & second format & +0.087 & $-$0.020 & +0.111 & $-$0.324 & $-$0.096 [$-$0.149, $-$0.042] & 0.13 \\
\bottomrule
\end{tabular}}
\end{table}

\begin{table}[tbp]
\centering
\scriptsize
\caption{\textbf{Rating-only, third-person prompt} (no description requested; valence and arousal in separate calls; system prompt). Five models on the 300-image subset (9{,}000 calls, 99.4\% valid); Qwen3-VL-235B is the existing 900-image run restricted to the same images. GLM-4.5V has 10 or 11 valid pleasant flagged images.}
\label{tab:rb_thirdperson}
\resizebox{\linewidth}{!}{%
\begin{tabular}{lllrrrrrrrr}
\toprule
\rowcolor{titlebar!90}
Dim. & Model & Format & $n$ & Older $-$ young & $p$ & Flagged & Pleasant flagged & $r_{\text{part}}$ [95\% CI] & Change [95\% CI] & Per-image $r$ \\
\midrule
valence & Claude-Sonnet-4.6 & submission & 300 & $-$0.086 & 0.017 & $-$0.544 & $-$1.371 & 0.085 [$-$0.052, 0.228] & -- & -- \\
valence & Claude-Sonnet-4.6 & rating-only, third person & 300 & $-$0.206 & $<$\,0.001 & $-$0.621 & $-$1.588 & 0.105 [$-$0.031, 0.249] & +0.020 [$-$0.068, 0.111] & 0.78 \\
valence & GLM-4.5V & submission & 292 & $-$0.090 & 0.069 & $-$0.473 & $-$1.455 & 0.135 [0.008, 0.265] & -- & -- \\
valence & GLM-4.5V & rating-only, third person & 291 & $-$0.188 & $<$\,0.001 & $-$0.474 & $-$1.360 & 0.037 [$-$0.087, 0.167] & $-$0.097 [$-$0.210, 0.008] & 0.63 \\
valence & GPT-5.6-Luna-Pro & submission & 298 & $-$0.065 & 0.005 & $-$0.173 & $-$0.535 & 0.081 [$-$0.054, 0.216] & -- & -- \\
valence & GPT-5.6-Luna-Pro & rating-only, third person & 300 & $-$0.104 & $<$\,0.001 & $-$0.106 & $-$0.135 & 0.141 [0.015, 0.264] & +0.060 [$-$0.105, 0.217] & 0.17 \\
valence & Gemini-3-Flash & submission & 297 & +0.020 & 0.530 & $-$0.324 & $-$0.812 & 0.107 [$-$0.023, 0.242] & -- & -- \\
valence & Gemini-3-Flash & rating-only, third person & 300 & $-$0.077 & 0.001 & $-$0.440 & $-$0.894 & 0.114 [$-$0.022, 0.243] & +0.006 [$-$0.092, 0.103] & 0.60 \\
valence & Llama-4-Scout & submission & 300 & $-$0.033 & 0.267 & $-$0.225 & $-$0.365 & 0.006 [$-$0.115, 0.136] & -- & -- \\
valence & Llama-4-Scout & rating-only, third person & 300 & $-$0.228 & $<$\,0.001 & $-$0.702 & $-$1.424 & 0.062 [$-$0.087, 0.210] & +0.056 [$-$0.087, 0.190] & 0.38 \\
valence & Qwen3-VL-235B & submission & 300 & $-$0.236 & $<$\,0.001 & $-$0.665 & $-$1.476 & 0.136 [0.003, 0.265] & -- & -- \\
valence & Qwen3-VL-235B & rating-only, third person & 300 & $-$0.341 & $<$\,0.001 & $-$0.675 & $-$1.588 & 0.046 [$-$0.082, 0.172] & $-$0.090 [$-$0.207, 0.035] & 0.61 \\
arousal & Claude-Sonnet-4.6 & submission & 300 & $-$0.244 & $<$\,0.001 & $-$0.448 & $-$1.076 & $-$0.237 [$-$0.344, $-$0.129] & -- & -- \\
arousal & Claude-Sonnet-4.6 & rating-only, third person & 300 & +0.050 & 0.031 & $-$0.015 & $-$0.435 & $-$0.109 [$-$0.207, $-$0.010] & +0.128 [0.030, 0.229] & 0.46 \\
arousal & GLM-4.5V & submission & 292 & $-$0.416 & $<$\,0.001 & $-$0.391 & $-$1.382 & $-$0.181 [$-$0.298, $-$0.068] & -- & -- \\
arousal & GLM-4.5V & rating-only, third person & 291 & $-$0.130 & $<$\,0.001 & $-$0.160 & $-$0.300 & $-$0.157 [$-$0.250, $-$0.066] & +0.023 [$-$0.117, 0.164] & 0.24 \\
arousal & GPT-5.6-Luna-Pro & submission & 298 & $-$0.179 & $<$\,0.001 & $-$0.343 & $-$0.647 & $-$0.133 [$-$0.226, $-$0.031] & -- & -- \\
arousal & GPT-5.6-Luna-Pro & rating-only, third person & 300 & +0.009 & 0.748 & $-$0.140 & $-$0.394 & $-$0.101 [$-$0.191, $-$0.013] & +0.032 [$-$0.092, 0.146] & 0.16 \\
arousal & Gemini-3-Flash & submission & 297 & $-$0.352 & $<$\,0.001 & $-$0.471 & $-$1.341 & $-$0.140 [$-$0.242, $-$0.035] & -- & -- \\
arousal & Gemini-3-Flash & rating-only, third person & 300 & $-$0.057 & 0.009 & $-$0.140 & $-$0.624 & 0.017 [$-$0.084, 0.118] & +0.157 [0.044, 0.272] & 0.41 \\
arousal & Llama-4-Scout & submission & 300 & $-$0.164 & $<$\,0.001 & $-$0.340 & $-$0.359 & $-$0.057 [$-$0.167, 0.048] & -- & -- \\
arousal & Llama-4-Scout & rating-only, third person & 300 & +0.010 & 0.773 & $-$0.181 & $-$0.306 & $-$0.009 [$-$0.108, 0.081] & +0.048 [$-$0.099, 0.193] & 0.07 \\
arousal & Qwen3-VL-235B & submission & 300 & $-$0.184 & $<$\,0.001 & $-$0.215 & $-$0.500 & $-$0.076 [$-$0.179, 0.036] & -- & -- \\
arousal & Qwen3-VL-235B & rating-only, third person & 300 & +0.090 & 0.001 & +0.017 & $-$0.488 & $-$0.058 [$-$0.151, 0.033] & +0.018 [$-$0.112, 0.144] & 0.14 \\
\bottomrule
\end{tabular}}
\end{table}

\subsection{Control personas with no age}
\label{app:rb-controls}
To test whether the older-persona response is an age stereotype or a cautious, safety-shaped style, we replaced the age sentence of the submission prompt with ``imagine that you are a retired person'' (a later-life role with no age) or ``imagine that you are a cautious person'' (Tables~\ref{tab:rb_controls_shift}--\ref{tab:rb_controls_lexicon}). ``Retired'' reproduces the older response: the same images move (median $r=0.75$), by a similar amount on pleasant flagged images, with the same rise in memory language. ``Cautious'' lowers valence on all images, correlates $0.46$ with the older shift, leaves $64\%$ to $99\%$ of its variance unexplained, and produces vulnerability rather than memory words. The older response is therefore tied to a later-life role and is not only a cautious style, although the two overlap on flagged images. The two prompts are shown in Figure~\ref{fig:prompts-control}.

\begin{figure}[p]
  \centering
  \promptpanel{14}{35 116.2 35 120.6}{(a)~Control persona: retired person (later-life role, no age)}\vspace{4pt}
  \promptpanel{15}{35 116.2 35 120.6}{(b)~Control persona: cautious person (safety style, no age)}\vspace{4pt}
  \caption{\textbf{Control-persona prompts.} The age sentence of the submission prompt is replaced by a persona that carries no age; highlighted text marks the change.}
  \label{fig:prompts-control}
\end{figure}

\begin{table}[tbp]
\centering
\small
\caption{\textbf{Control personas with no age: valence shift against the no-age prompt} (all 300 images / flagged / pleasant flagged). The age sentence is replaced by ``imagine that you are a retired person'' or ``\dots a cautious person''; six models, 3{,}600 calls, 99.5\% valid.}
\label{tab:rb_controls_shift}
\resizebox{\linewidth}{!}{%
\begin{tabular}{lcccc}
\toprule
\rowcolor{titlebar!90}
Model & Older & Retired & Cautious & Young \\
\midrule
Claude-Sonnet-4.6 & $-$0.02 / $-$0.38 / $-$1.02 & +0.08 / $-$0.25 / $-$0.77 & $-$0.48 / $-$0.60 / $-$1.45 & +0.07 / +0.17 / +0.35 \\
Gemini-3-Flash & +0.12 / $-$0.22 / $-$0.50 & +0.13 / $-$0.37 / $-$0.61 & $-$1.04 / $-$0.66 / $-$1.46 & +0.10 / +0.11 / +0.31 \\
GLM-4.5V & $-$0.02 / $-$0.36 / $-$1.31 & +0.16 / $-$0.01 / $-$0.64 & $-$1.35 / $-$0.82 / $-$2.58 & +0.07 / +0.11 / +0.15 \\
GPT-5.6-Luna-Pro & $-$0.03 / $-$0.24 / $-$0.48 & +0.00 / $-$0.31 / $-$0.73 & $-$0.55 / $-$0.70 / $-$1.61 & +0.03 / $-$0.07 / +0.05 \\
Llama-4-Scout & $-$0.08 / $-$0.28 / $-$0.43 & $-$0.03 / $-$0.20 / $-$0.31 & $-$0.36 / $-$0.34 / $-$0.66 & $-$0.04 / $-$0.05 / $-$0.06 \\
Qwen3-VL-235B & $-$0.15 / $-$0.62 / $-$1.42 & $-$0.12 / $-$0.71 / $-$1.60 & $-$0.35 / $-$0.53 / $-$1.32 & +0.08 / +0.05 / +0.05 \\
\textbf{Median} & $-$0.03 / $-$0.32 / $-$0.76 & +0.04 / $-$0.28 / $-$0.68 & $-$0.52 / $-$0.63 / $-$1.46 & +0.07 / +0.08 / +0.10 \\
\bottomrule
\end{tabular}}
\end{table}

\begin{table}[tbp]
\centering
\small
\caption{\textbf{How control-persona shifts relate to the older-persona shift} (valence, per image, 300 images).}
\label{tab:rb_controls_corr}
\resizebox{\linewidth}{!}{%
\begin{tabular}{lrrrrrr}
\toprule
\rowcolor{titlebar!90}
Model & $r$: retired vs older & $r$: cautious vs older & $r$: young vs older & Older $-$ cautious [95\% CI] & $p$ & Older-shift variance not explained by cautious \\
\midrule
Claude-Sonnet-4.6 & 0.89 & 0.56 & 0.02 & +0.462 [0.397, 0.527] & $<$\,0.001 & 69\% \\
Gemini-3-Flash & 0.80 & 0.08 & 0.30 & +1.165 [1.042, 1.288] & $<$\,0.001 & 99\% \\
GLM-4.5V & 0.70 & 0.43 & 0.13 & +1.334 [1.201, 1.467] & $<$\,0.001 & 82\% \\
GPT-5.6-Luna-Pro & 0.58 & 0.34 & 0.58 & +0.520 [0.431, 0.609] & $<$\,0.001 & 88\% \\
Llama-4-Scout & 0.70 & 0.50 & 0.49 & +0.285 [0.220, 0.350] & $<$\,0.001 & 75\% \\
Qwen3-VL-235B & 0.79 & 0.60 & 0.12 & +0.202 [0.137, 0.268] & $<$\,0.001 & 64\% \\
\textbf{Median} & 0.75 & 0.46 & 0.22 & +0.491 &  & 78\% \\
\bottomrule
\end{tabular}}
\end{table}

\begin{table}[tbp]
\centering
\small
\caption{\textbf{Control personas: arousal shift against the no-age prompt} (300 images).}
\label{tab:rb_controls_arousal}
\resizebox{\linewidth}{!}{%
\begin{tabular}{lrrrrrr}
\toprule
\rowcolor{titlebar!90}
Model & Older & Retired & Cautious & Young & $r$: retired vs older & $r$: cautious vs older \\
\midrule
Claude-Sonnet-4.6 & $-$0.139 & $-$0.141 & $-$0.204 & +0.105 & 0.85 & 0.52 \\
Gemini-3-Flash & $-$0.147 & $-$0.206 & +0.482 & +0.204 & 0.74 & 0.40 \\
GLM-4.5V & $-$0.340 & $-$0.450 & +0.261 & +0.076 & 0.63 & 0.44 \\
GPT-5.6-Luna-Pro & $-$0.144 & $-$0.298 & $-$0.170 & +0.035 & 0.61 & 0.45 \\
Llama-4-Scout & $-$0.115 & $-$0.196 & $-$0.227 & +0.049 & 0.58 & 0.60 \\
Qwen3-VL-235B & $-$0.125 & $-$0.190 & $-$0.100 & +0.058 & 0.67 & 0.54 \\
\bottomrule
\end{tabular}}
\end{table}

\begin{table}[tbp]
\centering
\scriptsize
\caption{\textbf{Share of descriptions containing memory or vulnerability words} (lexica of Table~\ref{tab:lexicon}).}
\label{tab:rb_controls_lexicon}
\resizebox{\linewidth}{!}{%
\begin{tabular}{lrrrrrrrrrr}
\toprule
\rowcolor{titlebar!90}
Model & Mem.: no age & Mem.: young & Mem.: older & Mem.: retired & Mem.: cautious & Vul.: no age & Vul.: young & Vul.: older & Vul.: retired & Vul.: cautious \\
\midrule
Claude-Sonnet-4.6 & 2\% & 5\% & 57\% & 70\% & 1\% & 13\% & 18\% & 30\% & 28\% & 62\% \\
Gemini-3-Flash & 0\% & 4\% & 24\% & 46\% & 1\% & 16\% & 16\% & 22\% & 19\% & 76\% \\
GLM-4.5V & 0\% & 1\% & 32\% & 52\% & 1\% & 16\% & 15\% & 24\% & 18\% & 69\% \\
GPT-5.6-Luna-Pro & 1\% & 1\% & 4\% & 8\% & 0\% & 18\% & 21\% & 26\% & 26\% & 48\% \\
Llama-4-Scout & 0\% & 1\% & 11\% & 7\% & 0\% & 11\% & 13\% & 18\% & 15\% & 30\% \\
Qwen3-VL-235B & 1\% & 6\% & 72\% & 78\% & 1\% & 17\% & 18\% & 28\% & 26\% & 56\% \\
\bottomrule
\end{tabular}}
\end{table}

\subsection{A second dataset: NAPS}
\label{app:rb-naps}
The NAPS release \citep{marchewka2014naps} provides pooled image norms without rater ages linked to individual ratings. Its norming sample comprised 204 volunteers, mainly college students and young employees (mean age 23.9 years, SD 3.4), so a young-adult prompt approximates the rater profile and the no-age prompt is the baseline on the same images. An older prompt would show how much a model changes, but NAPS provides no age-linked reference to score that change. We sampled 450 images across the five NAPS content categories and the valence range, mapped the 1--9 norms linearly to 1--7, and ran six models with each prompt (5{,}400 calls). Baseline competence replicates (median valence $r$ $0.89$ on NAPS and $0.90$ on OASIS), every calibration slope exceeds 1 on both datasets, and the young prompt improves agreement with the young-adult norms for 0 of 6 models and worsens it for 1 (Table~\ref{tab:rb_naps}). The older versus young fidelity analysis requires the participant-level ages that OASIS provides.

\begin{table}[tbp]
\centering
\scriptsize
\caption{\textbf{Second dataset: NAPS.} Six models rated 450 NAPS images under no-age and young prompts (5{,}400 calls, 99.85\% valid); NAPS norms (1--9) are mapped linearly to 1--7. OASIS rows are the same models on all 900 OASIS images, scored against raters aged 18--29. ``Change'' is the RMSE difference (young persona minus no age) against young-adult norms.}
\label{tab:rb_naps}
\resizebox{\linewidth}{!}{%
\begin{tabular}{lrrrrrrrrrrr}
\toprule
\rowcolor{titlebar!90}
Model & Dataset & $n$ & No-age valence $r$ & No-age arousal $r$ & Calibration slope & Young $-$ no age, valence ($p$) & Young $-$ no age, arousal ($p$) & RMSE, no age & RMSE, young & Change [95\% CI] & Label changes \\
\midrule
Claude-Sonnet-4.6 & OASIS & 899 & 0.900 & 0.709 & 1.23 & +0.087 ($<$\,0.001) & +0.088 ($<$\,0.001) & 0.862 & 0.854 & $-$0.008 [$-$0.028, 0.013] & 13\% \\
Claude-Sonnet-4.6 & NAPS & 449 & 0.896 & 0.656 & 1.18 & +0.022 (0.099) & +0.106 ($<$\,0.001) & 0.832 & 0.827 & $-$0.006 [$-$0.029, 0.017] & 11\% \\
\midrule
GLM-4.5V & OASIS & 873 & 0.895 & 0.668 & 1.28 & +0.102 ($<$\,0.001) & +0.073 (0.013) & 0.918 & 0.947 & +0.029 [$-$0.002, 0.062] & 13\% \\
GLM-4.5V & NAPS & 447 & 0.860 & 0.607 & 1.17 & $-$0.017 (0.470) & +0.223 ($<$\,0.001) & 0.921 & 0.907 & $-$0.014 [$-$0.066, 0.035] & 17\% \\
\midrule
GPT-5.6-Luna-Pro & OASIS & 895 & 0.890 & 0.675 & 1.35 & +0.050 ($<$\,0.001) & +0.045 (0.004) & 1.055 & 1.085 & +0.031 [0.008, 0.052] & 12\% \\
GPT-5.6-Luna-Pro & NAPS & 450 & 0.882 & 0.701 & 1.33 & +0.018 (0.323) & +0.010 (0.673) & 0.968 & 0.978 & +0.011 [$-$0.026, 0.049] & 9\% \\
\midrule
Gemini-3-Flash & OASIS & 899 & 0.912 & 0.676 & 1.24 & +0.102 ($<$\,0.001) & +0.174 ($<$\,0.001) & 0.834 & 0.902 & +0.068 [0.046, 0.091] & 14\% \\
Gemini-3-Flash & NAPS & 449 & 0.923 & 0.765 & 1.25 & +0.081 ($<$\,0.001) & +0.194 ($<$\,0.001) & 0.755 & 0.776 & +0.022 [$-$0.011, 0.057] & 14\% \\
\midrule
Llama-4-Scout & OASIS & 900 & 0.878 & 0.665 & 1.16 & $-$0.010 (0.547) & +0.032 (0.110) & 0.897 & 0.820 & $-$0.077 [$-$0.106, $-$0.046] & 13\% \\
Llama-4-Scout & NAPS & 450 & 0.855 & 0.407 & 1.08 & $-$0.046 (0.057) & $-$0.002 (0.949) & 0.835 & 0.800 & $-$0.035 [$-$0.089, 0.020] & 16\% \\
\midrule
Qwen3-VL-235B & OASIS & 900 & 0.902 & 0.638 & 1.18 & +0.088 ($<$\,0.001) & +0.076 ($<$\,0.001) & 0.948 & 1.016 & +0.068 [0.046, 0.091] & 16\% \\
Qwen3-VL-235B & NAPS & 450 & 0.907 & 0.669 & 1.16 & +0.038 (0.037) & +0.138 ($<$\,0.001) & 0.710 & 0.783 & +0.073 [0.035, 0.111] & 18\% \\
\bottomrule
\end{tabular}}
\end{table}

\begin{table}[tbp]
\centering
\small
\caption{\textbf{Agreement between models on the young $-$ no-age valence shift.}}
\label{tab:rb_naps_agree}
\resizebox{\linewidth}{!}{%
\begin{tabular}{lrrrrrr}
\toprule
\rowcolor{titlebar!90}
Dataset & Models & Images & Mean pairwise $r$ & Range & Positive mean shift & Median shift \\
\midrule
NAPS & 6 & 445 & 0.069 & $-$0.050 to 0.226 & 4 of 6 & +0.020 \\
OASIS & 6 & 869 & 0.203 & 0.085 to 0.387 & 5 of 6 & +0.088 \\
\bottomrule
\end{tabular}}
\end{table}

\subsection{Valence--arousal geometry in OASIS humans}
\label{app:rb-geometry}
\citet{keil2009changes} measure the coupling between pleasure and arousal within each person, which OASIS cannot measure because each rater provides only one dimension. Table~\ref{tab:rb_human_geometry} therefore computes, for OASIS raters by age group, the same across-image statistics used for the models. OASIS older raters show no reliable change in either coupling. The one reliable difference is that they rate pleasant images higher ($+0.309$ $[0.029,0.585]$ for 60+, $+0.373$ $[0.221,0.525]$ for 50+), whereas 16 of 17 models rate them lower under the older persona. The model counts here use the correlation-based coupling (17 of 17 rise on the aversive side); Appendix~\ref{app:geometry} reports the regression slope, which rises for 9 of 17 models.

\begin{table}[tbp]
\centering
\small
\caption{\textbf{Valence--arousal geometry in OASIS humans by age, computed with the same across-image statistics used for the models.} Young estimates use random subsamples of young raters matched to the older group's size (1{,}000 replicates). The last column counts models (of 17) that move upward under the older persona.}
\label{tab:rb_human_geometry}
\resizebox{\linewidth}{!}{%
\begin{tabular}{llrrrcr}
\toprule
\rowcolor{titlebar!90}
Contrast & Measure & Older & Young (matched $n$) & Older $-$ young [95\% CI] & Keil \& Freund predict & Models moving up \\
\midrule
60+ vs 18-$-$29 & Appetitive coupling (Fisher $z$) & 0.194 & 0.193 & +0.001 [$-$0.188, 0.197] & down & 0 \\
60+ vs 18-$-$29 & Aversive coupling (Fisher $z$) & 0.419 & 0.362 & +0.057 [$-$0.166, 0.336] & up & 17 \\
60+ vs 18-$-$29 & $R^2$, linear & 0.020 & 0.014 & +0.006 [$-$0.037, 0.040] & up & 16 \\
60+ vs 18-$-$29 & $R^2$, quadratic & 0.085 & 0.059 & +0.026 [$-$0.052, 0.097] & down & 0 \\
60+ vs 18-$-$29 & Mean valence, neutral images & 4.090 & 4.068 & +0.021 [$-$0.190, 0.241] & up & 2 \\
60+ vs 18-$-$29 & Mean valence, pleasant images & 5.487 & 5.178 & +0.309 [0.029, 0.585] & down & 1 \\
\midrule
50+ vs 18-$-$29 & Appetitive coupling (Fisher $z$) & 0.380 & 0.278 & +0.101 [$-$0.022, 0.223] & down & 0 \\
50+ vs 18-$-$29 & Aversive coupling (Fisher $z$) & 0.499 & 0.556 & $-$0.058 [$-$0.201, 0.075] & up & 17 \\
50+ vs 18-$-$29 & $R^2$, linear & 0.006 & 0.010 & $-$0.004 [$-$0.028, 0.018] & up & 16 \\
50+ vs 18-$-$29 & $R^2$, quadratic & 0.153 & 0.112 & +0.041 [$-$0.023, 0.103] & down & 0 \\
50+ vs 18-$-$29 & Mean valence, neutral images & 4.062 & 4.066 & $-$0.004 [$-$0.105, 0.092] & up & 2 \\
50+ vs 18-$-$29 & Mean valence, pleasant images & 5.548 & 5.175 & +0.373 [0.221, 0.525] & down & 1 \\
\bottomrule
\end{tabular}}
\end{table}

\subsection{Downstream test: predicting real older raters}
\label{app:rb-downstream}
A developer considering age conditioning can keep a model's baseline estimate of viewer pleasantness or revise it when told the viewer is older. Both choices produce a prediction before any human rating is consulted, so their difference in error measures the value of the age cue for that decision. Table~\ref{tab:rb_downstream} scores both against the mean ratings of real OASIS raters aged 60+ and 50+. Over 900 images, the older persona raises median RMSE against 60+ raters from $0.756$ to $0.789$, harming 7 of 17 models and helping none; on the 54 pleasant flagged images it raises median RMSE from $0.754$ to $1.214$ and harms 16 of 17. Starting instead from the observed mean of raters aged 18--29 (RMSE $0.583$), adding each model's predicted age adjustment raises median RMSE to $0.696$ and harms 17 of 17. Field audits of personalized assistants measure how answers change with user context \citep{citp2026chatgpt}; this test adds the comparison needed when the goal is prediction, namely whether the context-conditioned answer outperforms the same system without that context on independent outcomes.

\begin{table}[tbp]
\centering
\small
\caption{\textbf{Downstream test: predicting real older raters (valence).} Targets are mean ratings of OASIS raters aged 60+ (and 50+). Model ratings are calibrated to the human scale using the no-age condition. RMSE values are medians over 17 models; ``worse / better'' counts models whose change in squared error has a 95\% image-bootstrap interval excluding zero. Per-image older means rest on about 6 raters, which inflates all RMSEs equally but does not affect the paired comparisons.}
\label{tab:rb_downstream}
\resizebox{\linewidth}{!}{%
\begin{tabular}{llrrrrrr}
\toprule
\rowcolor{titlebar!90}
Target & Images & No age & Older persona & Young humans & Young humans + persona shift & Persona vs no age (worse / better) & Adding shift (worse / better) \\
\midrule
60+ raters & all 900 & 0.756 & 0.789 & 0.583 & 0.696 & 7 / 0 & 17 / 0 \\
60+ raters & flagged pleasant (54) & 0.754 & 1.214 & 0.622 & 1.314 & 16 / 0 & 17 / 0 \\
60+ raters & flagged unpleasant (113) & 0.923 & 0.914 & 0.594 & 0.682 & 2 / 4 & 6 / 0 \\
60+ raters & unflagged (733) & 0.730 & 0.733 & 0.578 & 0.641 & 1 / 2 & 14 / 0 \\
\midrule
50+ raters & all 900 & 0.683 & 0.731 & 0.495 & 0.636 & 10 / 0 & 17 / 0 \\
50+ raters & flagged pleasant (54) & 0.611 & 1.160 & 0.559 & 1.288 & 16 / 0 & 17 / 0 \\
50+ raters & flagged unpleasant (113) & 0.850 & 0.841 & 0.513 & 0.662 & 4 / 2 & 15 / 0 \\
50+ raters & unflagged (733) & 0.670 & 0.681 & 0.487 & 0.575 & 1 / 3 & 15 / 0 \\
\bottomrule
\end{tabular}}
\end{table}

\subsection{Mitigation: pleasantness-only baseline and held-out replication}
\label{app:heldout}
Table~\ref{tab:rb_rule} rescored every mitigation arm next to a rule that uses no model. The rule exceeds every arm on direction agreement and error, so those two measures show only that grounded arms recover what pleasantness predicts; $r_{\text{part}}$, which removes pleasantness, is the informative target. Because S4 was selected on the 300 development images, we ran it unchanged on the 600 remaining images (Tables~\ref{tab:rb_heldout} and~\ref{tab:rb_heldout_paired}). Its $r_{\text{part}}$ falls from $0.268$ to $0.105$ $[0.014,0.196]$, so the development value was optimistic. A gain remains: the held-out value exceeds zero ($p=.011$) and direct prompting ($+0.130$ $[0.006,0.254]$, $p=.041$; one pre-specified comparison), and S4 fixes 400 image directions while breaking 35. The pleasantness-only rule still exceeds S4 on direction agreement ($94.3\%$ vs.\ $85.3\%$) and error ($0.133$ vs.\ $0.210$).

\begin{table}[tbp]
\centering
\small
\caption{\textbf{Mitigation arms against a pleasantness-only rule (300 development images).} Direction agreement is the share of moved images whose older $-$ young direction matches humans; error is the mean absolute difference from the human older $-$ young effect. The rule uses no model: a quadratic in pooled pleasantness fitted on the other three lists.}
\label{tab:rb_rule}
\resizebox{\linewidth}{!}{%
\begin{tabular}{lp{6.2cm}rrr}
\toprule
\rowcolor{titlebar!90}
Arm & What it does & $r_{\text{part}}$ [95\% CI] & Direction agreement & Mean absolute error \\
\midrule
S0 direct prompt & Submission prompt, GPT-5.6-Luna-Pro & 0.081 [$-$0.061, 0.220] & 38.6\% & 0.475 \\
S1 debias instruction & Submission prompt plus one instruction not to rely on age stereotypes & 0.002 [$-$0.113, 0.115] & 44.1\% & 0.465 \\
S2 deliberation agent & Perceiver → persona rater → stereotype critic → reviser; no human data & $-$0.018 [$-$0.135, 0.098] & 40.1\% & 0.535 \\
S3 grounded agent & S2 plus retrieved human age differences for similar images from the other three lists & 0.089 [$-$0.039, 0.215] & 71.4\% & 0.367 \\
S4 grounded contrastive panel & S3 evidence, five simulated raters per age band, a calibrator anchored on the no-age rating, and an auditor & 0.268 [0.161, 0.373] & 87.5\% & 0.205 \\
S5 VERA & A hypothesized age-difference rule, kept only if it passes a re-test on held-out human cases; rating anchored on the no-age rating & 0.144 [0.028, 0.260] & 87.1\% & 0.198 \\
VERA-Cal & The verified rule from S5 passed to the rater panels and calibrator of S4 & 0.152 [0.035, 0.267] & 87.0\% & 0.200 \\
Pleasantness-only rule (no model) & No model: a quadratic in pooled pleasantness fitted on the other three lists & n/a & 95.1\% & 0.140 \\
\bottomrule
\end{tabular}}
\end{table}

\begin{table}[tbp]
\centering
\small
\caption{\textbf{The fixed S4 arm on 600 held-out OASIS images.} S4 was run unchanged (same prompts, model, and parameters) on the images outside the 300-image development subset (3{,}707 calls, 100\% valid) and scored as in $\S$\ref{app:agent-results}. Direction agreement is computed over moved images (count in parentheses).}
\label{tab:rb_heldout}
\resizebox{\linewidth}{!}{%
\begin{tabular}{llrrrrrrrr}
\toprule
\rowcolor{titlebar!90}
Images & System & $n$ & Older $-$ young & Pleasant flagged & Direction agreement & Mean absolute error & Pearson $r$ & $r_{\text{part}}$ [95\% CI] & $p$ ($r_{\text{part}}$$>0$) \\
\midrule
held-out 600 & S0 direct prompt & 596 & $-$0.082 & $-$0.478 & 38.1\% (310) & 0.511 & $-$0.103 & $-$0.025 [$-$0.101, 0.053] & 0.728 \\
held-out 600 & S4 grounded contrastive panel & 600 & +0.018 & +0.170 & 85.3\% (518) & 0.210 & 0.696 & 0.105 [0.014, 0.196] & 0.011 \\
held-out 600 & Pleasantness-only rule (no model) & 600 & +0.000 & +0.167 & 94.3\% (472) & 0.133 & 0.893 & -- & -- \\
\midrule
original 300 & S0 direct prompt & 299 & $-$0.065 & $-$0.535 & 38.6\% (140) & 0.475 & $-$0.110 & 0.081 [$-$0.061, 0.220] & 0.127 \\
original 300 & S4 grounded contrastive panel & 300 & +0.019 & +0.171 & 87.5\% (257) & 0.205 & 0.744 & 0.268 [0.161, 0.373] & $<$\,0.001 \\
original 300 & Pleasantness-only rule (no model) & 300 & $-$0.001 & +0.162 & 95.1\% (224) & 0.140 & 0.887 & -- & -- \\
\bottomrule
\end{tabular}}
\end{table}

\begin{table}[tbp]
\centering
\small
\caption{\textbf{S4 against direct prompting on the same images (paired tests).}}
\label{tab:rb_heldout_paired}
\begin{tabular}{lrrrrr}
\toprule
\rowcolor{titlebar!90}
Images & Gain in $r_{\text{part}}$ [95\% CI] & $p$ & Directions fixed / broken & $p$ (McNemar) & $p$ (Wilcoxon, error) \\
\midrule
held-out 600 & +0.130 [0.006, 0.254] & 0.041 & 400 / 35 & $<$\,0.001 & $<$\,0.001 \\
original 300 & +0.187 [0.023, 0.353] & 0.024 & 211 / 15 & $<$\,0.001 & $<$\,0.001 \\
\bottomrule
\end{tabular}
\end{table}
\endgroup

\begingroup
\addtocontents{toc}{\protect\apxpart{Part IV}{Qualitative Analysis}}
\FloatBarrierAbove
\section{Qualitative Examples}
\label{app:examples}

\endgroup

Figure~\ref{fig:qual} in $\S$\ref{sec:persona-patterns} shows three of these examples; Figures~\ref{fig:examples1}--\ref{fig:examples13} give all 15. We use them to show, image by image, the patterns summarised in Appendix~\ref{app:numerical-results}.
The examples illustrate the main-study findings; they are not additional evidence. We base the
main-study quantitative claims on the full 900-image analyses. The separate mitigation study uses
the 300-image subset described in Appendix~\ref{app:subset}.

\subsection{Image and model selection}

We draw 15 OASIS images that are not sensitive-flagged and that carry a supported human valence
age effect (Appendix~\ref{app:human}). To cover the stimulus set, we stratify by OASIS category
(Person, Scene, Animal, Object) and by pooled human valence, take in each stratum the image with
the largest 17-model older-minus-young shift, and allow one image per theme. Each page shows four
models. GPT-5.6-Luna-Pro, the strongest GPT model we evaluate, appears on every page as a fixed
reference. The other three come from three other developers, chosen as those with the largest
absolute shift on that image, and none of these additional models appears on more than four pages. All values and quotes
are copied verbatim from the released model responses.

\subsection{Panel definitions and colour coding}

The \emph{ground truth} panel reports pooled OASIS valence and arousal (1--7), the number of
raters, and the image-specific human age effect $\Delta H_i$ (per SD of rater age), estimated
from all 822 participants. The age effect is labelled green when older raters rate the image
more pleasant and red when they rate it less pleasant. Each \emph{model} panel reports:
\begin{itemize}
  \item baseline (no-age) valence and arousal, green when within one scale point of the human mean;
  \item valence and emotion label under the young, middle and older personas;
  \item the older-minus-young shift, green when it points in the direction of $\Delta H_i$,
        red when it points the opposite way, and grey when it is smaller than $0.1$;
  \item the older persona's description, with vulnerability, memory and excitement words
        coloured by the lexica in Table~\ref{tab:lexicon}.
\end{itemize}
The comparison is directional only: $\Delta H_i$ is a per-SD slope and the model shift is a
difference between two personas, so their magnitudes are not on a common scale.

\subsection{Patterns illustrated by the examples}

Four patterns recur across developers.
\textbf{(i) Direction is close to chance.} The 17-model mean shift agrees in sign with the human
age effect on 7 of the 15 images, points the opposite way on 7, and does not move on one
(Picnic~2). This is consistent with the near-zero image-level alignment in
Table~\ref{tab:alignment_valence}. Agreement, where it occurs, also appears on images where human and
model shifts share a sign for unrelated reasons (Fire~6, Boxing~2).

\textbf{(ii) Leisure and risk scenes are scripted as danger.} On Skydiving~3, Rollercoaster~3 and
Cliff diver~3, older OASIS raters rate the image slightly \emph{more} pleasant, yet models switch
from \emph{joy} to \emph{fear}. On Cliff diver~3, GPT-5.6-Luna-Pro moves from $6.0$ to $3.0$ and
describes the scene in terms of risk, worry and safety.

\textbf{(iii) The older persona reminisces.} Descriptions add nostalgia, memories and invented
grandchildren (Picnic~2, Pigeon~6, Car race~4), sometimes while the rating is unchanged, so the
text ages independently of the judgment.

\textbf{(iv) The middle persona usually copies one end.} Across the 60 model panels, the middle
rating tends to equal either the young or the older rating rather than lie between them, matching
the step pattern in Figure~\ref{fig:monotonicity}.

\subsection{Examples}

\newcommand{\exrow}[1]{%
  \includegraphics[page=#1,height=0.305\textheight,keepaspectratio]{figure/appendix_examples_v3.pdf}}
\newcommand{\appexamples}[4]{%
  \begin{figure}[p]
    \centering
    \exrow{#1}\\[4pt]
    \exrow{#2}\\[4pt]
    \exrow{#3}
    \caption{Qualitative examples #4. Panel definitions and colour coding are given in
    Appendix~\ref{app:examples}.}
    \label{fig:examples#1}
  \end{figure}}
\appexamples{1}{2}{3}{1--3}
\appexamples{4}{5}{6}{4--6}
\appexamples{7}{8}{9}{7--9}
\appexamples{10}{11}{12}{10--12}
\appexamples{13}{14}{15}{13--15}
\clearpage

\end{document}